\documentclass[11pt,letterpaper,oneside]{article}
\usepackage[margin=1in]{geometry}
\usepackage[T1]{fontenc}
\usepackage{lmodern}
\usepackage{amsmath,amssymb,amsthm}
\usepackage{graphicx,booktabs}
\usepackage{float}
\floatstyle{ruled}
\newfloat{algorithm}{tbp}{loa}
\floatname{algorithm}{Algorithm}
\usepackage{tikz}
\usetikzlibrary{shapes.geometric,calc}
\usepackage[round,authoryear]{natbib}
\usepackage{microtype,xurl}
\usepackage[hidelinks]{hyperref}
\makeatletter
\AtBeginDocument{\renewcommand\@biblabel[1]{[#1]}\let\@bibsetup\NAT@bibsetnum}
\makeatother
\newcommand{\R}{\mathbb R}
\newcommand{\E}{\mathbb E}
\newcommand{\Prob}{\mathbb P}
\newcommand{\bbox}{\operatorname{bbox}}

\newcommand{\TV}{\operatorname{TV}}

\newcommand{\one}{\mathbf 1}

\newcommand{\lca}{\operatorname{lca}}
\newcommand{\Out}{\operatorname{Out}}
\newtheorem{definition}{Definition}
\newtheorem{proposition}{Proposition}
\newtheorem{theorem}[proposition]{Theorem}
\newtheorem{corollary}{Corollary}
\newtheorem{lemma}{Lemma}
\theoremstyle{remark}

\title{Random Cut Forests, Games and Diffusion over Approximately Nested Anisotropic Local Geometry}
\author{Sudipto Guha\thanks{Email:{\tt contactsudiptoguha@gmail.com}}}
\date{}
\begin{document}
\maketitle

\begin{abstract}
Just as the data in a dynamic data stream changes, questions asked of the stream also change in their intent; which pieces of information to be held constant, counterfactuals and so forth. Energy based learning has been celebrated for its uniform approach towards losses and estimations, but interpretability remains challenging. 
Tree-based methods, on the other hand, offer high interpretability but often rely on bespoke, disjointed estimators. 
Neither paradigm readily supports changing questions on streaming data.
In this paper we take a step towards combining combining energy, interpretability, and streaming.

Random cut forests retain an evolving geometry of the data. We study this geometry as the basis for diffusion dynamics, prediction, explanation, and games. We show that several of the scores the forest already reports are escape probabilities of random walks on its trees. The affinity those walks induce is realized exactly by a sparse electrical network on each tree, which gives a single energy calculus, a Hamiltonian, that the stream keeps current along one path per update. 
The common energy formulation connects an estimate, its supporting evidence, and the response to intervention through the same boxes, gaps, masses, and paths.
The different questions then correspond to different clamp decisions on the energy. Fixing observed values gives harmonic prediction with Gaussian uncertainty. Fixing a partition gives normalized cuts whose cost is attributed exactly to oriented faces. Releasing clamps gives coalition games, in which collusive displacement measures the response of the field. With noise, the same energy is a diffusion dynamics whose relaxation rates are read from the boxes. Axis-parallel cuts localize at a rate set by the affine dimension of the data rather than the ambient one, and this control passes to the averaged energy. Across a forest the boxes are only approximately nested; common reference observations align the trees. A rotating star with added noise illustrates the measurements.  The resulting framework interprets the forest as a maintained geometric representation whose related estimators can be computed and examined through a common energy calculus.
\end{abstract}

\section{Introduction}
\label{sec:intro}
A stream changes, and so do the questions asked of it; often without foreknowledge.
An inference is often the beginning of an analysis. A partition of the
data, for example one obtained by normalized cuts, prompts questions
about which directions support a group, where it is weakly connected
to its surroundings, and how a new observation relates to it. An
outlier score prompts related questions: which coordinates contributed,
what values would be expected in their place, and what would change
if a value were required to be larger than the current one?
The answer to one question generates a need for further information
and verification.

These questions differ in intent. Reading a prediction can hold the
observations and reference values fixed. Testing a prescribed answer
allows the other inferred values to respond. Scoring a possible
insertion asks how the query would participate in the model. The
distinctions are expressed through conditioning and normalization:
what is held fixed, what may change, and against which population
an answer is compared. Out-of-sample extensions of spectral methods
make a related distinction between extending an answer and refitting
its evidence \citep{bengio2004outofsample,alzate2010multiway,delalleau2005efficient}.
Chaining inference routines can obscure these distinctions when each
routine supplies its own neighborhood, scale, or reference population.

Decision forests have long served regression and classification
\citep{Breiman1984,breiman1996bagging,freund1996experiments,friedman2001greedy,breiman2001random}. Streaming tree learners also maintain fitted
predictors as observations arrive
\citep{domingos2000mining,gomes2018regression}.
These constructions reflect
choices about which distinctions are relevant to a particular prediction task.
However a directional explanation or a counterfactual likely asks for
different distinctions which are discarded by a predictor. 
 \citet{geurts2006extremely} analyze how random splits on irrelevant attributes reduce the sample
available for prediction, introducing bias that is not removed by ensemble averaging. 
Tree-specific explanation methods take a different approach, 
\citet{lundberg2020local} develop efficient Shapley attributions. Using those attributions for another decision requires
specifying the reference distribution and the intended use.
\citet{bhattacharjee2025discard}, for example, show that small aggregate
SHAP values over the data support need not justify discarding a
feature, and derive guarantees using an extended distribution.
These results reinforce the need to connect an explanation, or a question that is not known ex-ante, to original inference without constructing families of small estimators that fragment the sample or repeat work.

We believe that algorithms should explain their own work, without the need for depending on other algorithms for explanations. The observations, local geometry, and conditions used to obtain an answer
should remain available to the questions that follow it.
Relevance depends on the question: a distinction that contributes little to a
prediction may still matter to a directional explanation or a
counterfactual; and follow-up questions often arise on instances where prediction is ambiguous.
Retaining that evidence also has an efficiency
benefit. When an estimate and its directional contributions are
accumulated in the same traversal, separate calls for the answer and
its decomposition become one joint query and can be efficiently pipelined in a modern streaming system. For that pair, two tree
traversals become one; richer interventions can reuse the geometry
while specifying the additional computation they require.

Energy-based learning provides a uniform formulation of related
inference tasks: observed variables are fixed and the remaining ones
are inferred by minimizing a common energy
\citep{lecun2006tutorial}. For explanations, its terms must also be
connected to the observations and directions retained by the algorithm,
and the boundary conditions must specify what a query changes.
Here the energy is derived from the local geometry represented by
the trees. Its terms
remain connected to the boxes, gaps, masses, and paths from which
they were computed. Changing the question then changes the boundary
conditions or the measurement of this same geometry. This gives
prediction, explanation, and intervention a common calculation,
with explicit evidence for each reading.
We call this scalar energy the Hamiltonian; 
its noisy relaxation is described by Langevin dynamics with random thermal forces (i.e., Brownian motion) \citep{langevin1908,lemons1997,pavliotis2014stochastic}.
For the quadratic fields used here, conditioning the associated
Gaussian law gives Kriging predictions and uncertainty
\citep{stein1999interpolation,rasmussen2006gaussian}.

A random cut forest (RCF) provides one way to maintain this evidence
as observations change over a dynamic data stream \citep{guha2016rrcf}. Its trees recursively cut
data boxes, and insertions and deletions preserve their distribution;
the estimators are kept apart from the maintained tree structures
\citep{guha2019rcfblog}. Scores, attributions, imputed values, and
forecasts are already read from the same trees in OpenSearch
\citep{opensearchad,opensearchforecast,opensearchrcf}.
RCF also provides anomaly detection in Amazon Managed Service for
Prometheus \citep{ampanomaly}. 

This paper proposes the development of uniform, energy based analysis techniques that can function on dynamic data streams. It uses RCF both as an example that separates stream maintenance, as well as a class of trees that provides a wealth of properties that make the energy analysis over streams feasible. In particular this paper shows that many of the known uses of RCF can be reinterpreted as hitting functionals of random walks on the trees. We can then 
derive their energy, and use that energy to treat predictions,
partitions, and coalitions together. The geometric identities and energy calculations are conditional on
the chosen observations and reference sets; they require no particular
sampling scheme. Maintaining the trees as their observations change
is distinct from jointly adapting the reference and query sets.
Axis-parallel cuts can adapt to affine structure: for width-proportional
cuts we prove localization and stretch guarantees controlled by affine
dimension, without estimating or rotating the support. This gives a
coordinate-cut counterpart, under an affine-support assumption, to the
intrinsic-dimension questions studied for random projection and other
partition trees \citep{dasgupta2008random,verma2009spatial}.
Theorem~\ref{thm:geometryenergy} transfers this geometric control to
the averaged Hamiltonian.

\paragraph{Main Results}
\label{sec:mainresults}
We state the results informally, in the order of the sections that prove them.

\begin{description}
\item{Local cuts and affine dimension (Section~\ref{sec:model}).}
We prove localization and stretch guarantees controlled by affine
dimension for axis-parallel RCF cuts, without identifying the support's
orientation or rotating the observations
(Theorem~\ref{prop:affinegeometry}). The guarantees persist above a
bounded-noise scale (Corollary~\ref{cor:noisygeometry}). Exact
insertion/deletion consistency keeps this $\ell_1$ geometry maintainable
as the sample changes (Proposition~\ref{prop:maintainable}).

\item{Hitting times and scores (Section~\ref{sec:existing}).}
We give inverse-depth, range, and normalized displacement a common
escape interpretation, identifying what each measurement reads from
the same tree geometry (Theorem~\ref{thm:escape}); collusive
displacement averages the maximum mass-escape odds along a path
(Corollary~\ref{cor:displacement}).
Requiring a declared reference mass gives near-duplicate resilience
without coalition search (Theorem~\ref{prop:massescape}). The same
calculation retains $2d$ signed contributions that sum to the
measurement \eqref{eq:recurrence}.

\item{Sparse energy and diffusion (Section~\ref{sec:exact}).}
We realize a dense pairwise agreement energy exactly by a sparse
network on each tree (Theorem~\ref{prop:laplacerealization}). Affine
geometric control passes to the averaged Hamiltonian
(Theorem~\ref{thm:geometryenergy}) and bounds the statistical
distinguishability of Gaussian mean shifts \eqref{eq:affinefisher}.
The free observation field's merge contrasts relax at rates read
directly from the hierarchy \eqref{eq:mergemodes}; grounded Langevin
relaxation has stationary covariance equal to temperature times the
Green's function \eqref{eq:langevin}.

\item{Landmarks and Normalized Cuts  (Section~\ref{sec:extension}).}
We align different tree geometries on common references and compute
forest affinity products and normalized-cut objectives by linear tree
passes, without constructing a dense affinity matrix
(Proposition~\ref{prop:nested}). The pooled Kriging variance separates
missing support, weak connection after first contact, and the cost of
requiring common prediction weights across trees
(Theorem~\ref{prop:krigingsupport}); range escape is the initial rate
of its support term.

\item{Coalition games and strategic responses (Section~\ref{sec:recursiveclamping}).}
We construct a quadratic Hamiltonian and an exact potential game for
arbitrary worths assigned to descendant coalitions. Its Green's
function is independent of the worths, and two tree passes find a
pure equilibrium under participation clamps
(Theorem~\ref{prop:laminarparticipation}). On the mass-resistance
path, collusive displacement is the square of the largest relative
change of the minimizing field under a one-clamp release, measured
in the energy norm \eqref{eq:codisprelaxation}.

\item{Collective cut improvement and attribution (Section~\ref{sec:recursiveclamping}).}
We prove that a partition can resist every smaller move while a
descendant-block move improves its normalized cut by an arbitrarily
large factor (Theorem~\ref{prop:ncutcoalitiongap}). The collusive cut
proposes blocks by tree passes and tests a common move against the
pooled objective (Theorem~\ref{prop:collusivecut}). Boundary prices
attribute the fixed cut cost exactly to tree merges, to oriented
faces together with unresolved leaf contributions, and to members
(Theorem~\ref{prop:boundaryprices}; \eqref{eq:contourmeasure} and
\eqref{eq:cutshares}).

\item{Directional audits and perturbations (Section~\ref{sec:audits}).}
We compute how directional geometry changes the minimized energy
(Proposition~\ref{prop:directionalaudit}) and sharply bound changes
in harmonic predictions under conductance perturbations, with edge
support and boundary values fixed
(Theorem~\ref{prop:energystability}). Audit stiffness quantifies the
energy cost of requiring a different answer at the probe
\eqref{eq:probeaudit}.

\item{Measurements and an illustration (Section~\ref{sec:measurements}).}
Directional ratios \eqref{eq:aloft} compare a query's expansion
with a neighborhood of declared mass. 
A rotating star with noise
coordinates compares these readings, the cuts, and their
Nystr\"om extension.
\end{description}

\paragraph{Approximate Nesting and Organization}
\label{sec:nesting}
Each tree supplies an exact laminar hierarchy of descendant sets,
with boxes nested along every ancestral chain. Inspecting both
children or comparing siblings does not change this hierarchy.
Query intent can nevertheless require a branching family of
neighborhoods: masking coordinates or augmenting boxes by the query
can produce overlapping boxes from different branches
(Sections~\ref{sec:completion} and~\ref{sec:forecast}). Across trees,
the resulting neighborhoods need not belong to one common nested
hierarchy. We use \emph{approximately nested} geometry for these
combined readings; the per-tree escape identities and energy
realizations remain exact. Common references align the trees for
pooled predictions and cuts (Section~\ref{sec:extension}).

In closing, the sections take multiple steps toward one goal. Section~\ref{sec:model} discusses the affine geometry.
Section~\ref{sec:existing} explains what the forest already computes, as hitting times on its trees. Sections~\ref{sec:exact} and~\ref{sec:extension} hold the boundary fixed and read the field it induces: predictions, their uncertainty, and the energy of a partition on common references. Section~\ref{sec:recursiveclamping} moves the boundary and lets the players respond. Sections~\ref{sec:audits} and~\ref{sec:measurements} keep the directional evidence, bound its response to perturbation, and compare the measurements on a rotating star. Diffusion ties these together: the multiscale structure of the trees becomes the multiscale structure of a dynamics, whose rates are read from the boxes 
and change only on an insertion or a deletion that affects the approximately nested geometry. 
Affine structure controls tree geometry; tree geometry controls energy; energy controls distinguishability of field responses.
Hitting times, Laplacians, and Nystr\"om extensions thus formalize the notion of diffusion dynamics in streaming data; estimates and objects of interest being formed and dissociated by clamping the energy of the operators, aided by the information geometry \citep{amari2000methods} of approximately nested anisotropic local neighborhoods.

\section{Local and Global Geometry}
\label{sec:model}
This section identifies the evidence a tree retains and the
computations a query can reuse. We begin with the two sides of each
coordinate, the boxes around the observations, and the observations
visited by a query. We then define the cut family and explain why
length allocation permits insertions and deletions while preserving
the tree distribution. The same probabilities combine an answer and
its directional contributions on the way back up the tree.

The global question is whether these local cuts preserve useful
distances. We recall the original RCF stretch guarantee, then show
how affine dimension controls localization and stretch without
rotating the coordinates, including above a bounded-noise scale.
The final two examples use these ingredients to explain a score,
propose a replacement, and complete a time-series history. Section~\ref{sec:existing}
then reads the query probabilities as escape functionals.

\subsection{Local Geometry and Query Traversals}
\label{sec:halfdimensions}\label{sec:fold}\label{sec:law}
Let $S$ be a finite empirical sample in $\R^d$, allowing multiplicities or nonnegative weights. A rooted binary tree assigns each nonempty node $v$ a sample $S_v$, mass $m_v$, and data box $B_v=\bbox(S_v)=\prod_j[\ell_{vj},u_{vj}]$. Write $\psi=m_{\rm root}$ for the total sample mass. These boxes are nested along ancestry. The routing cell, defined by the cuts, can contain a query outside its data box.

\begin{definition}[Query-relative half dimensions]
\label{def:halfdimensions}
Fix a complete query $x\in\R^d$ and write $t_+=\max\{t,0\}$. The two \emph{half dimensions} of coordinate $j$ retain the nonnegative reaches
\[
 z^-_{x,j}(y)=(x_j-y_j)_+,\qquad z^+_{x,j}(y)=(y_j-x_j)_+.
\]
Their extents on $S_v$ are the reaches from $x$ to the faces of $M_v(x)=\bbox(B_v\cup\{x\})$:
\[
 a^-_{vj}=\max_{y\in S_v}z^-_{x,j}(y)=(x_j-\ell_{vj})_+,\qquad
 a^+_{vj}=\max_{y\in S_v}z^+_{x,j}(y)=(u_{vj}-x_j)_+.
\]
\end{definition}
The $2d$ reaches retain both sides of the original $d$ coordinates.
For fixed $x$, the augmented boxes remain nested along ancestry.

Fix coordinate units. At a box $B=\prod_j[\ell_j,u_j]$, put $r_j=u_j-\ell_j$, $R(B)=\sum_jr_j$, and define the exterior face gaps
\[
 g_j^-=(\ell_j-x_j)_+,\qquad g_j^+=(x_j-u_j)_+,\qquad G=\sum_j(g_j^-+g_j^+).
\]

\begin{definition}[Query transcript]
A \emph{transcript} records the visited nodes, their ancestry,
masses and boxes, including the origin of query-adjusted and sibling
boxes. At a split on an observed coordinate, the query follows its
child. A missing split coordinate admits both children; their
projected boxes can overlap.
\end{definition}
On the way down, the traversal identifies the relevant observations
and geometry. On the way back up, it combines the child answers
under the query's declared branch probabilities. We call this upward
combination a \emph{fold}. It can carry a scalar estimate and its
directional contributions together. The next subsection specifies
the cut law and the probabilities used in this computation.

\subsection{Cut Laws and Dynamic Maintenance}
\label{sec:cutgeometry}\label{app:cutlaws}\label{app:hypergeometry}
A cut law specifies which coordinate is cut and where its threshold
falls. Repeating the law on the two children defines the distribution
of a tree. We first give the family of cut laws, then ask which one
can be maintained by inserting and deleting observations.

\begin{definition}[The $p$-cut law and its tree distance]
\label{def:pcut}
At a box with coordinate widths $w_j$, choose a positive-width axis
with probability
\begin{equation}
 \pi_{p,j}(w)=\frac{w_j^p}{\sum_{l:w_l>0}w_l^p},
 \qquad w_j>0,\quad p\geq0,
 \label{eq:powerallocation}
\end{equation}
then choose a uniform threshold on that axis and recurse.
The exponent $p=0$ gives uniform coordinate allocation, as in
isolation forests \citep{liu2008isolation}; $p=1$ gives the
length allocation of RCF; $p=2$ gives squared-width allocation.
For $p>0$ put $R_p(B)=(\sum_jw_j^p)^{1/p}$, with $R_1(B)=R(B)$.
For $p\geq1$ this is the $\ell_p$ diameter of the box.
For stored points and $p>0$ define $d_{T,p}(a,b)=R_p(B_{\lca(a,b)})$;
for $p\geq1$ write $\Delta_p(A)=\operatorname{diam}_p(A)$ for their data
diameter. Uniform allocation is an exponent-zero cut law, with
no $\ell_0$ ruler intended.
\end{definition}

\paragraph{Insertion and restriction.}
For $p=1$, an auxiliary cut of $M=\bbox(B\cup\{x\})$ separates
the query on the exterior face intervals with probabilities
\begin{equation}
 p_{j,\pm}(x,B)=\frac{g_j^\pm}{R(B)+G},\qquad
 q(x,B)=\frac{G}{R(B)+G}.
 \label{eq:hazard}
\end{equation}
The denominator is $R(M)$, since
$a_j^-+a_j^+=r_j+g_j^-+g_j^+$. At zero expanded range set
these probabilities to zero. At a node $v$ write
$G_v=R(M_v(x))-R(B_v)$ for its total exterior gap.
If the cut does not isolate the query, its restriction to the old
box has the original cut law. An existing cut of $B$ therefore
supplies the correct continuation of an insertion.

The restriction criterion also explains deletion. For any
width-based coordinate probabilities $\pi_j(w)$ and uniform
thresholds, let $a_j(B)=\pi_j(w(B))/w_j(B)$ be the cut-position
density on a positive-width axis. For $B\subseteq M$,
\begin{equation}
 W_j(B,M)=\frac{a_j(M)}{a_j(B)},\qquad
 \E_B W=\sum_jw_j(B)a_j(M).
 \label{eq:restrictedcut}
\end{equation}
The expectation is the probability that a cut of $M$ lies in
the old cut intervals. Conditional on those intervals, the old
cut law is recovered exactly when $W_j$ is constant there.
Length allocation gives $W_j=R(B)/R(M)=1-q(x,B)$.

\begin{proposition}[Length allocation and deletion]
\label{prop:maintainable}
Suppose a recursive tree law chooses a positive-width coordinate $j$ with probability $\pi_j(w)>0$ as a function of the widths $w$, then a uniform threshold on that coordinate. Deleting any stored location and contracting its parent preserves the construction law on the remaining sample if and only if
\[
 \pi_j(w)=\frac{c_jw_j}{\sum_l c_lw_l}
 \qquad(c_j>0)
\]
on the positive-width coordinates. The probability of separating two points of a box $B$ is then their weighted $\ell_1$ distance divided by $\sum_jc_jR_j(B)$.
\end{proposition}
\begin{proof}
For these probabilities, rescaling the coordinates by $c_j$ gives the ordinary random-cut construction, which is preserved by deletion \citep{guha2016rrcf}. For the converse, enlarge a box of widths $w'$ by adding one point beyond just one face, producing widths $w$. A root cut outside the old box isolates the added point, leaving a fresh subtree of the old sample after contraction. A root cut inside the old box remains the root cut after deletion. The contracted root law is the mixture of these two laws. Whenever the old box admits a cut, the second has positive weight and must equal the original root law. By \eqref{eq:restrictedcut}, the densities $\pi_j(w)/w_j$ and $\pi_j(w')/w'_j$ are proportional on the positive coordinates of $w'$. By enlarging one coordinate at a time to the coordinatewise maximum of any two width vectors, this fixes every density ratio on their shared positive coordinates. Comparing to an all-positive box gives the constants $c_j$, and normalization gives the stated probabilities. Integrating the cut density over the intervals between the two points gives the separation probability.
\end{proof}

\paragraph{Combining query responses.}
On a path $v_0,\ldots,v_h$ from the root, put $p_{i,j,\pm}=p_{j,\pm}(x,B_{v_i})$, $q_i=q(x,B_{v_i})$, $s_0=1$, and $s_{i+1}=s_i(1-q_i)$. The separation law, including a terminal state $\partial$, is
\begin{equation}
 \mu_{T,x}\{(i,j,\pm)\}=s_i p_{j,\pm}(x,B_{v_i}),\quad
 \mu_{T,x}\{\partial\}=s_{h+1}.
 \label{eq:measure}
\end{equation}
Here $s_i$ is survival to level $i$, whereas $q_i$ is the local separation probability. For an unseen query taken to a singleton leaf the last hazard is one; a matching query or a truncated path retains a terminal atom with a declared payoff.

\begin{proposition}[Transcript expectations]
For a finite-valued scalar or vector payoff $g$, the separation law has total mass one and its expectation is computed on the way back up the path
\begin{equation}
 H_i=\sum_{j,\pm}p_{i,j,\pm}g(i,j,\pm)+(1-q_i)H_{i+1}.
 \label{eq:recurrence}
\end{equation}
with $H_{h+1}=g(\partial)$ and $H_0=\int g\,d\mu_{T,x}$.
\end{proposition}
\begin{proof}
The level mass is $s_iq_i=s_i-s_{i+1}$. Summation telescopes, including the terminal atom. Expanding the recurrence gives the stated integral.
\end{proof}

\paragraph{Reading another cut law.}
The maintained tree uses length allocation. A query can nevertheless
estimate a reading under another cut law by likelihood weighting
\citep{rcfcode}. For a target coordinate law $\pi_j(M)$ on the
augmented box, its separating-face probabilities are
\begin{equation}
 p^{\rm target}_{j,\pm}(x,B)
   =\pi_j(M)\frac{g_j^\pm}{w_j(M)},\qquad
 q_{\rm target}=\sum_{j,\pm}p^{\rm target}_{j,\pm}.
 \label{eq:hyperhazard}
\end{equation}
Flat coordinates contribute zero to \eqref{eq:hyperhazard}.
For a nonseparating cut on axis $j$, reweight continuation by
\begin{equation}
 W_j(B,M)=\frac{w_j(B)}{w_j(M)}
               \frac{\pi_j(M)}{\pi_{1,j}(w(B))}.
 \label{eq:cutlikelihood}
\end{equation}
This is the density ratio \eqref{eq:restrictedcut}, taking its
numerator from the target law on $M$ and its denominator from
the maintained law on $B$. It requires positive base density
where the target has nonseparating cuts.
For a stored pair, let $D_T(i,j)$ be its box distance at separation
under a declared ruler, and $\mathcal P_{ij}$ its common path,
including the separating cut. The corresponding change of law is
\begin{equation}
 \E_p e^{-D_T(i,j)/h}
 =\E_1\left[e^{-D_T(i,j)/h}
       \prod_{v\in\mathcal P_{ij}}
       \frac{\pi_{p,j(v)}(w_v)}{\pi_{1,j(v)}(w_v)}\right].
 \label{eq:cutaffinitychange}
\end{equation}
Uniform threshold densities cancel from the likelihood ratio.
For a pair in the same leaf, stop the product at that leaf.
Likelihood weighting estimates an average under another cut law;
changing the ruler on fixed cuts (Section~\ref{sec:rulerenergy})
needs no weights. Equation~\eqref{eq:cutaffinitychange} is an
expectation identity: finite weighted pair estimates need not form a
positive semidefinite matrix. Section~\ref{sec:fan} uses the query
weights to compare cut laws with the same inverse-depth payoff.

\subsection{Affine Dimension and Global Stretch}
\label{sec:globalgeometry}
Having fixed the cut law, we ask which distances its boxes retain
and how many levels are needed to localize observations. For
coordinate vectors we write $\ell_p$; $L^p(\mu)$ denotes a function
space with a declared measure. We begin with the original RCF
stretch argument, then show how affine dimension gives control
independently of the number of coordinates.

\paragraph{The original stretch identity.}
At a common cell of a stored pair, put $h_j=|a_j-b_j|$ and
$\delta=\|a-b\|_p$. For $p\geq1$, its separation probability
$q_p$ satisfies
\begin{equation}
 R_p(B)q_p
 =\frac{\sum_{j:w_j>0}w_j^{p-1}h_j}{R_p(B)^{p-1}}
 \leq\delta.
 \label{eq:matchedcutdistortion}
\end{equation}
At $p=1$ this is equality. Consequently, with $\tau_{ab}$ counting
the common cuts through separation, the original RCF identity
\citep[Supplement, Lemma~7]{guha2016rrcf} is
\begin{equation}
 \frac{\E_1 d_{T,1}(a,b)}{\|a-b\|_1}=\E_1\tau_{ab}
 .
 \label{eq:originalstretchidentity}
\end{equation}
Writing $R_0=R(B_{\rm root})$ and $\delta=\|a-b\|_1$, the
ambient bound from the same argument is
\begin{equation}
 \frac{\E_1 d_{T,1}(a,b)}{\delta}
 =O\!\left(d[1+\log(R_0/\delta)]\right).
 \label{eq:originalstretchbound}
\end{equation}
For example, following the child containing $a$, a uniform cut
removes at least one quarter of the selected width in expectation.
Thus the expected range decreases by at least
$\sum_jw_j^2/(4R)\geq R/(4d)$ at each step. Once its range is
less than $\delta$ the pair has separated, giving the bound.

\paragraph{Control by affine dimension.}
The following range lemma replaces ambient coordinate counting
with a property of the support. The localization theorem combines
it with the net and cluster-pair argument of
\citet{dasgupta2008random}, using finite-dimensional $L^p$
geometry \citep{lewis1978finite,cohen2015lewis}.

\begin{lemma}[Range of an affine set]
\label{lem:affinerange}
Let $S\subset\R^d$ lie in an affine subspace of dimension $r\geq1$,
fix $1\leq p<\infty$, and put $\Lambda_{r,p}=r^{\max\{1,p/2\}}$.
Every nonempty $A\subseteq S$ satisfies
\begin{equation}
 R_p(\bbox(A))^p\leq\Lambda_{r,p}\,\Delta_p(A)^p.
 \label{eq:affinerange}
\end{equation}
In particular $\Lambda_{r,p}=r$ for $1\leq p\leq2$. At $p=1$ the
summed widths of a bounding box are at most $r$ times its $\ell_1$
diameter, with equality for the cross-polytope
$A=\{\pm e_1,\ldots,\pm e_r\}$.
\end{lemma}
\begin{proof}
Lewis's normalization of the $r$-dimensional direction
space gives a parametrization
$z_j=\lambda_j^{1/p}\langle u_j,t\rangle$, where
$\|u_j\|_2=1$ and $\sum_j\lambda_ju_ju_j^\top=I_r$;
coordinates that vanish on the space can be omitted
\citep{lewis1978finite,cohen2015lewis}. Taking traces gives
$\sum_j\lambda_j=r$. On $\|t\|_2=1$, for $p\leq2$ the sum
$\sum_j\lambda_j|\langle u_j,t\rangle|^p$ is at least one;
for $p\geq2$, the power-mean inequality bounds it below by
$r^{1-p/2}$. Thus
$\|t\|_2\leq r^{(1/2-1/p)_+}\|z\|_p$.
Each coordinate width of $A$ is at most
$\lambda_j^{1/p}r^{(1/2-1/p)_+}\Delta_p(A)$.
Summing their $p$th powers proves \eqref{eq:affinerange}. For the equality, the cross-polytope has every width equal to $2$ and $\ell_1$ diameter $2$.
At $p=2$, the $\lambda_j$ are the coordinate leverage scores
of the subspace and sum to $r$.
\end{proof}

\begin{theorem}[Affine dimension, localization, and stretch]
\label{prop:affinegeometry}
Fix $1\leq p<\infty$. Let $S\subset\R^d$ have at least two distinct
locations and lie in an affine subspace of dimension $r\geq1$.
Put $\Delta=\Delta_p(S)$ and, with $\Lambda_{r,p}$ of
Lemma~\ref{lem:affinerange},
\[
 \kappa_{r,p}=4^p\Lambda_{r,p},\qquad
 \mathbb K_{r,p}(\eta)=\left\lceil\kappa_{r,p}
 \left[2r\log(1+16\kappa_{r,p})+\log(1/\eta)\right]\right\rceil.
\]
In the random tree of the $p$-cut law,
with probability at least $1-\eta$, every depth-$\mathbb K_{r,p}(\eta)$
descendant has data diameter at most $\Delta/2$, for $0<\eta<1$.
Terminal leaves have diameter zero. Moreover, for $a,b\in S$ with
$\delta=\|a-b\|_p>0$,
\begin{equation}
 d_{T,p}(a,b)\geq\delta,\qquad
 \frac{\E d_{T,p}(a,b)}{\delta}
 \leq2\mathbb K_{r,p}(1/2)
       \left(1+\left\lceil\log_2(\Delta/\delta)\right\rceil\right).
 \label{eq:affinestretch}
\end{equation}
For fixed $p$ the constant-probability depth is
$O_p(r^{1+\max\{1,p/2\}}\log(1+r))$, independently of ambient
dimension. At $p=1$ this reduces to
$\mathbb K_{r,1}(\eta)=\lceil4r[2r\log(1+64r)+\log(1/\eta)]\rceil$.
The construction need not identify or rotate the support.
\end{theorem}
\begin{proof}
Write $\Lambda=\Lambda_{r,p}$ and $\kappa=\kappa_{r,p}$; the geometric
input is \eqref{eq:affinerange}.
Take a maximal $a_0$-separated net in $S$, with
$a_0=\Delta/(8\kappa)$, and assign each point to a covering
ball. Volume packing in the $r$-dimensional normed support
gives at most $M=(1+16\kappa)^r$ clusters $C_i$, each
containing its center $c_i$, with diameter at most $2a_0$.
If points of $C_i,C_j$ can be more than $\Delta/2$ apart,
their centers are more than $\Delta/2-2a_0$ apart.
Let $g_j$ and $o_j$ be the coordinate gap and overlap lengths
of their two boxes. Equation~\eqref{eq:affinerange} gives
\[
 \|\boldsymbol g\|_p\geq
 \Delta/2-(2+4\Lambda^{1/p})a_0\geq5\Delta/16,
 \qquad \|\boldsymbol o\|_p\leq2\Lambda^{1/p}a_0.
\]

Call a cell mixed when it contains points from both clusters.
At a mixed cell with widths $w$, a cut in a global gap interval
leaves no mixed child. Two mixed children require a cut in an
overlap interval. The corresponding cut densities are
$w_j^{p-1}/R_p(B)^p$. Since $w_j\geq g_j$ and
$R_p(B)\leq\Lambda^{1/p}\Delta$, H\"older's inequality gives
\[
 \sum_{j:w_j>0}w_j^{p-1}(g_j-o_j)
 \geq\left[(5/16)^p-4^{-(p+1)}\right]\Delta^p
 \geq4^{-p}\Delta^p.
\]
The expected number of mixed children is therefore at most
$1-1/\kappa$. The probability this pair has a mixed cell
at depth $k$ is at most $e^{-k/\kappa}$.
A union bound over the $M^2$ cluster pairs proves localization.

Conditional on previous cuts, the same result applies to the
pair's current common cell. In a block of $\mathbb K_{r,p}(1/2)$ levels,
with conditional probability at least $1/2$, the pair separates
or that cell's data diameter halves. After
$1+\lceil\log_2(\Delta/\delta)\rceil$ successful blocks it must
separate, so the expected number of shared cuts is at most
twice that number of blocks. If $h_j=|a_j-b_j|$, the separation
probability $q$ at a common cell satisfies
\[
 R_p(B)q=\frac{\sum_{j:w_j>0}w_j^{p-1}h_j}{R_p(B)^{p-1}}
 \leq\|a-b\|_p=\delta.
\]
Summing conditional separation payoffs proves the upper
bound in \eqref{eq:affinestretch}; domination follows because
the separating box contains both points. At $p=1$, the last
inequality is the exact separation identity of
\citet[Supplement, Lemma~7]{guha2016rrcf}.
\end{proof}

For $p=1$ on affine data, both \eqref{eq:originalstretchbound} and
\eqref{eq:affinestretch} apply, so one can use their minimum.
The ambient estimate can be smaller when $r$ is comparable to $d$;
the affine estimate removes ambient dimension when many coordinates
describe a low-dimensional support.

\paragraph{The range of $p$.}
The theorem needs no assumption beyond the affine support. Domination
holds for every $p$ because the separating box contains both points,
and the upper bound uses only H\"older's inequality and
Lemma~\ref{lem:affinerange}. For $1\leq p\leq2$ the lemma's constant
is $r$, so the depth is $O(4^p r^2\log(1+r))$ throughout that range;
for $p>2$ the lemma's factor is $r^{p/2}$, giving the depth
stated in the theorem. What singles out $p=1$ is
maintenance: by Proposition~\ref{prop:maintainable} it is the only
exponent whose trees survive deletion, and the other laws are
simulated from it by likelihood weights (Section~\ref{sec:cutgeometry}).

\paragraph{What changes below one.}
Positive powers suppress vanishing coordinate widths, but the
stretch estimate needs $p\geq1$. For $0<p<1$, take
$S=\{a=(0,0),b=(0,\varepsilon),c=(1,0)\}$, with
$0<\varepsilon<1$. The root separates $a,b$ with probability
$\varepsilon^p/(1+\varepsilon^p)$ and then assigns the ruler
$R_p(B_{\rm root})=(1+\varepsilon^p)^{1/p}\geq1$.
Their own separation is $\|a-b\|_p=\varepsilon$, so
\begin{equation}
 \frac{\E_p d_{T,p}(a,b)}{\varepsilon}
 \geq\frac{\varepsilon^{p-1}}{1+\varepsilon^p}.
 \label{eq:subunitstretch}
\end{equation}
This grows polynomially as $\varepsilon\downarrow0$, already on
three points in affine dimension two. Thus the logarithmic
scale-ratio bound does not extend to these exponents. Here $R_p$
is only a formal box ruler; $\ell_p$ is not a norm below one.

\begin{corollary}[Affine localization above a noise scale]
\label{cor:noisygeometry}
Suppose each $x\in S$ has a fixed representation $x=s_x+n_x$,
where the $s_x$ lie in one $r$-dimensional affine subspace,
$r\geq1$, and $|(n_x)_j|\leq\sigma_j$. Put $W=\sum_j\sigma_j$.
Every nonempty $A\subseteq S$ satisfies
\begin{equation}
 R(\bbox A)\leq r\,\Delta_1(A)+2(r+1)W.
 \label{eq:noisyaffinerange}
\end{equation}
For ordinary width-proportional cuts, let $\Delta=\Delta_1(S)>0$
and suppose $\Delta\geq16W$. For $0<\eta<1$, with
\[
 \mathbb K^{\rm noise}_r(\eta)=
 \left\lceil10r\left[2r\log(1+72r)+\log(1/\eta)\right]\right\rceil,
\]
every depth-$\mathbb K^{\rm noise}_r(\eta)$ descendant has data
diameter at most $\Delta/2$ with probability at least $1-\eta$;
terminal leaves have diameter zero.
Moreover, for a stored pair with $\delta=\|a-b\|_1>0$ and
$\delta\geq16W$,
\begin{equation}
 d_{T,1}(a,b)\geq\delta,\qquad
 \frac{\E d_{T,1}(a,b)}{\delta}
 \leq2\mathbb K^{\rm noise}_r(1/2)
       \left[1+\left\lceil\log_2(\Delta/\delta)\right\rceil\right].
 \label{eq:noisyaffinestretch}
\end{equation}
\end{corollary}
\begin{proof}
For $A_0=\{s_x:x\in A\}$, the coordinate noise adds at most $2W$
to the summed widths, and
$\Delta_1(A_0)\leq\Delta_1(A)+2W$. Lemma~\ref{lem:affinerange}
proves \eqref{eq:noisyaffinerange}.
The noisy observations need not have a low-dimensional affine span,
so cover their affine parts instead. These have diameter at most
$\Delta+2W\leq9\Delta/8$. A maximal
$a_0$-separated net with $a_0=\Delta/(32r)$ has at most
$M=(1+72r)^r$ centers. Assign observations through their affine
parts to covering centers. Each resulting cluster $C_i$ satisfies
\[
 R_i:=R(\bbox C_i)\leq2r a_0+2W\leq3\Delta/16.
\]
For a pair of clusters containing observations more than $\Delta/2$
apart, let $g_j,o_j$ be the coordinate gap and overlap lengths of
their boxes. The interval identity for hull width gives
\[
 \sum_j(g_j-o_j)
 =R(\bbox(C_i\cup C_l))-R_i-R_l>\Delta/8.
\]
At a cell containing both cluster types, a cut in a global gap
leaves no mixed child; two mixed children require a cut in a global
overlap. Since its data box has
$R(B)\leq r\Delta+2(r+1)W\leq5r\Delta/4$,
the expected number of mixed children is at most $1-1/(10r)$.
The probability a fixed pair has any mixed cell at depth $k$
is at most $e^{-k/(10r)}$. A union bound over $M^2$ pairs proves
the diameter claim. A single cluster has diameter at most $3\Delta/16$.

Conditional on earlier cuts, the same argument applies to any
current common cell while its diameter is at least
$\delta\geq16W$. Each block of $\mathbb K^{\rm noise}_r(1/2)$
levels has conditional probability at least $1/2$ of separating
the pair or halving that diameter. After
$1+\lceil\log_2(\Delta/\delta)\rceil$ successes the pair must
separate. The exact separation probability at a common cell is
$\delta/R(B)$, so the expected separating range divided by
$\delta$ equals the expected number of shared cuts, including
the separating cut. The block bound proves
\eqref{eq:noisyaffinestretch}; the separating box contains both
points, proving domination.
\end{proof}

Theorem~\ref{thm:geometryenergy} transfers this affine control, including
the noise floor, to the averaged Hamiltonian.
Its Fisher-metric interpretation in Section~\ref{sec:exact} bounds the
statistical distinguishability of Gaussian mean shifts.
Thus $O(r^2\log(1+r))$ levels give constant-probability contraction
above a conservative noise scale, without identifying or rotating
the support. Below that scale the noise may itself be high
dimensional. Section~\ref{sec:fan} uses a similar perturbation
model when it pads the star with bounded noise coordinates.

\paragraph{Uniform coordinate allocation.}
Uniform coordinate allocation, the exponent-zero cut law of
isolation forests, can require ambient-dimension depth even on
affine dimension two. Fix $0<\varepsilon<1$ and take
\[
 S_m=\{(a,\varepsilon t/m,\ldots,\varepsilon t/m):
          a\in\{0,1\},\ t\in\{0,2^{-m},\ldots,2^{-1},1\}\}
       \subset\R^{m+1}.
\]
Its $\ell_1$ diameter is $1+\varepsilon$. The two points with $t=0$
and opposite $a$ remain together until a cut on the first coordinate.
Along their common branch, while $t$ still varies, that coordinate
is selected with probability $1/(m+1)$. Every other cut replaces
the maximum $2^{-j}$ by $2^{-(j+K)}$, or zero if $j+K>m$,
where $\Prob(K=k)=2^{-k}$ and $\E K=2$.
After $k$ cuts, a union bound and Markov's inequality give probability
at least $1-k/(m+1)-2k/m$ that neither a first-coordinate cut nor
collapse of the $t$ range has occurred. For $k\leq m/12$ this is
at least $3/4$. The common cell then has diameter at least one,
greater than half the original diameter. Thus no bound depending
only on affine dimension can hold for this uniform cut law.

\paragraph{Affine dimension versus doubling dimension.}
Affine dimension controls the coordinate ranges in
\eqref{eq:affinerange}. Bounded doubling dimension alone cannot
replace it: even at $p=1$ a metric of doubling dimension at most two
can have growing expected stretch under the coordinate cut law.
For example, take the $n=2^m$ leaves of a
complete binary hierarchy and one coordinate for each
nonroot vertex $v$. With $\rho=2^{-1/2}$, put
$x_z(v)=\rho^j$ when $v$ is on leaf $z$'s root path at
depth $j$, and zero otherwise. Leaves whose common ancestor
is at depth $j$ have distance $2\sum_{l=j+1}^m\rho^l$.
This is an ultrametric of doubling dimension at most two:
four grandchild clusters cover each cluster at half its
diameter. Its diameter is bounded, but its range is
$R_m=\sum_{j=1}^m(2\rho)^j=\Theta(\sqrt n)$.

Fix two leaves in opposite root branches. Until they
separate, a nonseparating coordinate cut removes a descendant
cluster. Write $c=\rho/(1-\rho)$ and let
$P_v=\sum_{l=j}^m2^{l-j}\rho^l$ sum the coordinate weights
indexed by a depth-$j$ vertex and its descendants. Such a removal
loses at most $P_v+c$ range, including emptied ancestors,
and $\sum_v\rho^{\operatorname{depth}(v)}P_v\leq cR_m$
by summing first over the descendant coordinates.
While the remaining range is at least $R_m/2$, the expected
nonseparating range loss per cut is thus at most $3c$,
and separation probability at most $4c/R_m$. Stopping at
separation or when range first falls below $R_m/2$, Markov's inequality and a union
bound give probability at least $1-10ck/R_m$ that neither
has occurred after $k$ cuts. Thus $\Omega(\sqrt n)$ levels
can pass with constant probability and no diameter decrease;
the separation identity gives $\Omega(\sqrt n)$ expected
stretch as well.

Thus a bound on doubling dimension alone does not supply the
localization or stretch guarantee of
Theorem~\ref{prop:affinegeometry}. The distinction concerns the
coordinate representation of the metric: its doubling dimension
does not control the summed widths used by the cuts.

\subsection{Example I: From a Score to a Replacement}
\label{sec:completion}
The upward recurrence \eqref{eq:recurrence} accumulates a score and its face contributions together:
credit each separation payoff to the face on which it occurs.
The contributions sum to the scalar score, and averaging them
with the same weights across trees preserves this identity.
Section~\ref{sec:escape} defines the inverse-depth choice and its
escape interpretation. Here we use the retained face and reference
box to decide what to inspect: sum the two faces to rank coordinates,
and keep their split to locate the direction contributing the score.

Let $J$ contain the coordinates selected for investigation. A counterfactual replacement masks them and holds the remaining values fixed. At a cut on an observed coordinate the query follows its child; at a cut on $J$ both children remain feasible. Leaves supply candidate values for $J$, and completed queries can be scored by the same recurrence. One selection rule compares branch candidates by score plus an independent uniform priority, favoring lower scores while retaining alternatives. Repeated selections are summarized by coordinatewise medians. The selection weights determine the distribution being summarized.

The replacement is in the observation's units. Rescoring it checks whether the selected contributions fall, or the anomaly shifts to other directions. A reporting rule can also require a material change, such as a $20\%$ deviation from the replacement, with an absolute-scale convention near zero. This chain of attribution, conditional replacement, and rescoring is used in OpenSearch \citep{opensearchad,opensearchrcf,li2025falsepositives}.

Imputation starts with a supplied mask and preserves the observed coordinates before detection. A prescribed fill rule is also possible: carrying a previous value forward bridges an expected reporting gap, while filling a missing count with zero can expose a cessation of activity \citep{li2025missing}. The mask and fill choice remain part of the evidence for the resulting score.

\subsection{Example II: Forecasting by Completion}
\label{sec:forecast}
A shingle of length $k+1$ pairs the history $Z_t=(X_{t-k+1},\ldots,X_t)$ with the continuation $Y_t=X_{t+1}$. A forecast fixes the history at its observed value $z_t$ and masks $Y_t$; the same traversal supplies conditional continuations, summarized by their median \citep{opensearchforecast}. Masking an earlier block instead gives a retrospective replacement, which can use readings already present later in the shingle. The mask specifies the question and the available evidence.

The cut law determines which distinctions this traversal can expose. Take the periodic signal $0^k1^k$, $k\geq2$, with independent continuous median-zero noise supported on $[-\sigma,\sigma]$, $0<\sigma<1/4$. Thresholding the last $k$ readings at $1/2$ identifies the next noiseless bit. At a node where $s>0$ of the $d=k+1$ coordinates still distinguish phases, those widths are at least $1-2\sigma$; the other widths are at most $2\sigma$. All widths are positive almost surely. Uniform coordinate allocation, as in isolation trees and fully randomized Extra-Trees, and the positive-power cut laws give \citep{liu2008isolation,geurts2006extremely,guha2016rrcf}
\begin{equation}
 p_{\rm noise}^{(0)}=\frac{d-s}{d},\qquad
 p_{\rm noise}^{(p)}
 \leq\frac{(d-s)(2\sigma)^p}
              {s(1-2\sigma)^p+(d-s)(2\sigma)^p},\qquad p>0
 \label{eq:blocknoise}
\end{equation}
for the share of cuts spent on noise-only coordinates. At $p=1$ this is length allocation. At fixed $d$, every $p>0$ makes this share vanish with $\sigma$; uniform allocation does not. The bound follows by maximizing the noise widths and minimizing the phase-varying widths.

At a node restricted to the patterns $(0^k,1)$ and $(1,0^{k-1},0)$, only the oldest lag and the continuation have unit-scale variation. As $\sigma\downarrow0$, a clean oldest-lag cut has probability $1/d$ under uniform allocation and $1/2$ under width-proportional allocation. That cut distinguishes the observed histories; a cut on the missing continuation requires branching. Thus vanishing noise can retain a fixed share of uniform cuts. Within a phase, common positive scaling of the noise leaves each normalized power cut law unchanged, so repeated observations still permit refinement of the phase cloud. The completion weights determine how that resolution enters the forecast. Proposition~\ref{prop:blockallocation} states these facts.

\begin{proposition}[Cut allocation on the block signal]
\label{prop:blockallocation}
At a node holding records from a set of phases, call a coordinate phase-varying when both binary values occur among its records, and let $s$ be their number.
\begin{enumerate}\setlength{\itemsep}{1pt}
\item[(i)] At a node with $s>0$, a cut falls on one of the other $d-s$ coordinates with probability $(d-s)/d$ under uniform allocation, and with probability at most the right side of \eqref{eq:blocknoise} for every fixed $p>0$.
\item[(ii)] In a box holding records of the two patterns $a=(0^k,1)$ and $b=(1,0^{k-1},0)$ only, a clean oldest-lag cut, which chooses coordinate one and a threshold in $[\sigma,1-\sigma]$, separates the two histories; its probabilities satisfy
\begin{equation}
 p_{\rm clean}^{(0)}\geq\frac{1-2\sigma}{d(1+2\sigma)},\qquad p_{\rm clean}^{(1)}\geq\frac{1-2\sigma}{2+2d\sigma},
 \label{eq:blockclean}
\end{equation}
which tend to $1/d$ and $1/2$ as $\sigma\downarrow0$. More generally, every fixed $p>0$ gives clean-cut probability tending to $1/2$ at fixed $d$.
\item[(iii)] At a node whose records all come from one phase, the cut law for every fixed $p\geq0$, with its threshold normalized to the box, does not depend on $\sigma>0$.
\end{enumerate}
\end{proposition}
\begin{proof}
(i) With bounded noise, each phase-varying width is at least $1-2\sigma$ and each other width is at most $2\sigma$; continuous noise makes every width positive almost surely at a node with two distinct records. Uniform allocation assigns $(d-s)/d$ to the noise-only coordinates. Under exponent $p>0$, their share is largest when their widths are $2\sigma$ and the others $1-2\sigma$, which gives \eqref{eq:blocknoise}. (ii) In this box the oldest-lag and continuation widths lie in $[1-2\sigma,1+2\sigma]$ and each of the other $k-1$ widths is at most $2\sigma$. Under uniform allocation, coordinate one is chosen with probability $1/d$ and a clean threshold then with probability at least $(1-2\sigma)/(1+2\sigma)$. Under width-proportional allocation the coordinate weight cancels its width, leaving the length $1-2\sigma$ of the clean interval over the total extent, at most $2+2d\sigma$. For every fixed $p>0$, the two phase-varying widths tend to one, the other widths to zero, and a clean threshold occupies an asymptotically full share of the oldest-lag interval. This gives the same $1/2$ limit. (iii) Records of one phase have the form $b_\phi+\sigma u_i$, so every coordinate width is $\sigma$ times the corresponding width of the $u_i$; all power-based coordinate probabilities and normalized thresholds are unchanged by a common positive factor, and the same holds at every node below.
\end{proof}

\section{Hitting Time Functionals}
\label{sec:existing}

A forest is an ensemble of controlled random partitions of the space, and random walks and their hitting functionals are natural tools for analyzing such structures. The partitions are maintained and algorithmically designed, so they supply query-relative measurements of mass, extent, gaps, and scale through related stochastic experiments, and the relation among the experiments makes the measurements interpretable.

A surprising connection of studying hitting time is that some of the existing scores of RCF released in open source, \citet{guha2019rcfblog}, can be reinterpreted as an escape score --- 
indicating a curious view that anomaly attempts to escape being characterized by the remainder of data.
Section~\ref{sec:escape} reads inverse depth, displacement, and range
escape as one escape measured in three lengths.
Section~\ref{sec:games} reads escape at a declared mass, providing
resilience to near duplicates without deleting records.
Section~\ref{sec:density} then retains the separation locations as a
hitting bundle: cut, passage, and calibration boxes with directional
extents and variation across trees. The same additive path lengths
will supply the pair interactions used by Section~\ref{sec:exact}.

\paragraph{Sampling and conditioning.}
\label{sec:sampling}
Let $\mathsf Q$ be a declared law on finite samples $S$.
It may subsample or resample a fixed batch, or draw from a
specified population; a point mass at the full batch gives
the fixed-sample case. Conditional on $S$, draw a tree from the
build law $\mathsf B(\,\cdot\mid S)$ of
Section~\ref{sec:cutgeometry}. A query then supplies its separation
law $\mu_{T,x}$. Unless an additional expectation is stated,
a path identity holds conditional on the sampled records and
the realized cuts. A theorem that averages cuts while fixing
$S$ keeps this conditioning explicit.

A forest averages over the sampled records and the trees built
on them. Different samples change the nearby observations,
boxes, masses, and possible attachments. This sampling variation
is present even in batch data, and is distinct from bounded
coordinate perturbations in Section~\ref{sec:globalgeometry}.
We use $T$ as shorthand for a tree realization together with
its retained sample; its ensemble law is
$\mathsf Q(dS)\mathsf B(dT\mid S)$.
Independent draws give a Monte Carlo approximation to this
declared average. The finite-tree identities and empirical
averages themselves require no independence. These snapshot
calculations do not specify how a stream maintains its samples.

\subsection{Escape in Depth, Range, and Mass}
\label{sec:escape}
For an unseen query followed to a singleton-location leaf,
use the separation law \eqref{eq:measure} from
Section~\ref{sec:halfdimensions}. Its attachment probability
\(s_iq_i\) places \(x\) beside \(v_i\).
In this virtual
insertion a new parent \(u\) joins \(x\) and \(v_i\), and the
query becomes a leaf at depth \(i+1\). Write \(T_i\) for this
augmented tree and \(r_i\) for its root. The new parent is
also the root when \(v_i\) was the old root. For a seen query, consider removing the query and its prior copies and perform a shadow (pretend) operation of inserting all the copies simultaneously -- but now the shadow insertion can be at any allowed level reproducing the distribution.

\begin{definition}[Escape on the augmented tree]
\label{def:escape}
In an augmented tree with root \(r\), give each edge a positive
length and use its reciprocal as conductance, so a random walk
chooses among neighboring edges in proportion to their conductances.
Write \(L(w)\) for the total length from \(x\) to \(w\) along the
root path, with \(L(x)=0\), and \(\tau_x^+\) for the first positive
return time to \(x\). The walk started at \(x\) \emph{escapes} if it
reaches \(r\) before it returns to \(x\).
\end{definition}

\begin{theorem}[Scores as escape probabilities]
\label{thm:escape}
Let \(u\) be the parent of \(x\) in the augmented tree. Then
\begin{equation}
 \Prob_x(\tau_r<\tau_x^+)=\frac{L(u)}{L(r)}.
 \label{eq:heights}
\end{equation}
Write \(C_x=R(M_{\rm root}(x))\). When \(x\) attaches beside \(v_i\),
the three lengths of this section give
\begin{center}
\begin{tabular}{@{}lll@{}}
\toprule
Length & Each edge measures & Escape when \(x\) attaches beside \(v_i\)\\
\midrule
Depth & One branch decision & \(1/(i+1)\)\\
Range & Increase in summed box widths & \(R(M_{v_i}(x))/C_x\)\\
Mass & Original sample mass added & \(m_{v_i}/\psi\)\\
\bottomrule
\end{tabular}
\end{center}
Averaged over the separation law, the first is the unseen-query
inverse-depth score of the library, the second is the range escape
\eqref{eq:rangeescape}, and the third is normalized displacement
(Corollary~\ref{cor:displacement}).
\end{theorem}
\begin{proof}
Hold \(x\) at voltage zero and \(r\) at one. Branches off their
connecting path carry no current, and the voltage rises in proportion
to cumulative resistance. After the obligatory first step to \(u\),
the chance of reaching \(r\) first is its voltage, \(L(u)/L(r)\)
\citep{doyle1984random,lyons2016probability}. With unit lengths the
first edge has length one and the path \(i+1\); with range lengths
they are \(R(M_{v_i}(x))\) and \(C_x\); with mass lengths, excluding
the query, they are \(m_{v_i}\) and \(\psi\).
\end{proof}

\begin{proposition}[Inverse-depth escape]
Give each edge unit length. At an attachment beside \(v_i\),
the walk satisfies
\begin{equation}
\Prob_x^{T_i}(\tau_{r_i}<\tau_x^+)=\frac{1}{i+1}.
 \label{eq:inverse}
\end{equation}
Averaging over the separation law gives
\begin{equation}
 U(T,x)=\sum_is_iq_i\Prob_x^{T_i}(\tau_{r_i}<\tau_x^+)
       =\sum_i\frac{s_iq_i}{i+1}.
 \label{eq:score}
\end{equation}
Thus \(U(T,x)\) is the escape probability when the attachment
is drawn first and the walk is then run on that augmented tree.
\end{proposition}
\begin{proof}
The first edge has length one and the whole path has length
\(i+1\). Apply \eqref{eq:heights} and average, using
\(\sum_is_iq_i=1\).
\end{proof}

A shallow attachment gives a larger escape probability.
Ranking queries by this probability is the inverse-depth
anomaly score. The forest averages \(U(T,x)\) over its trees.
This is the unnormalized unseen-query calculation underlying
the OpenSearch score \citep{opensearchad,opensearchrcf}.
\begin{proposition}[Range escape]
\label{prop:heights}
Use range as the cumulative length on the augmented query
path, with length zero at \(x\); contract any edge with zero
range increment. The first edge has length
\(R(B_u)=R(M_{v_i}(x))\), and the whole path has length
\(C_x\). With the gaps \(G_v\) and survivals \(s_v\)
of Section~\ref{sec:halfdimensions},
\begin{equation}
 \E\bigl[\Prob_x(\tau_r<\tau_x^+)\bigr]
 =\frac{\E R(B_u)}{C_x}
 =\frac{\sum_v s_vG_v}{C_x}.
 \label{eq:rangeescape}
\end{equation}
This is also the probability that an independent random cut
of \(M_{\rm root}(x)\) crosses the sampled separation box \(B_u\).
\end{proposition}
\begin{proof}
Equation~\eqref{eq:heights} gives \(R(B_u)/C_x\) at each
attachment. Its expectation is
\(\sum_vs_vq_vR(M_v(x))/C_x=\sum_vs_vG_v/C_x\).
An independent cut chooses a coordinate in proportion to its
width and a uniform position on that coordinate. The fraction
of those positions crossing \(B_u\) is \(R(B_u)/C_x\).
\end{proof}

Range escape measures the fraction of the tree's spatial
extent still in play at separation, averaged over the
attachment. Replacing \(G_v\) in \eqref{eq:rangeescape} by
each of its face gaps gives contributions that sum to the
same probability.
The box ranges also give the tree distance used in the RCF
analysis \citep{guha2016rrcf}. Put \(H_v=R(B_v)/2\) on a
sample tree and give an edge from parent \(p\) to child \(v\)
length \(H_p-H_v\). Stored locations have height zero, so
\begin{equation}
 d_T(a,b)=2H_{\lca(a,b)}
          =R(B_{\lca(a,b)})\geq\|a-b\|_1.
 \label{eq:rcfrangemetric}
\end{equation}
For the escape, the cumulative range length is \(L=2H\)
on the augmented query path. This common factor cancels
from the ratio in \eqref{eq:heights}.

\begin{corollary}[Displacement and CoDisp]
\label{cor:displacement}
Use original sample mass as the cumulative length on the
query path, excluding the inserted query. When \(x\) attaches
beside \(v_i\), the first edge has length \(m_{v_i}\) and
the whole path has length \(\psi\). The escape probability
is therefore \(m_{v_i}/\psi\). Exactly the old records below
\(v_i\) gain a branch decision on insertion, so normalized
displacement is the expected mass escape
\citep{guha2016rrcf}.

For CoDisp, consider a tree with root \(r\) containing \(x\), and at each cut
on its path take the whole child subtree containing \(x\) as
the departing coalition. If its mass is \(a\) and its sibling's
mass is \(b\), deleting that coalition saves one branch decision
for each record in the sibling. The saving per departing record
is \(b/a\). CoDisp takes the largest such ratio over the path:
\[
 \operatorname{CoDisp}(T,x)
   =\max_{w\ne r\text{ on the path of }x}
           \frac{m_{w'}}{m_w},
\]
where \(m_w\) is the mass below \(w\) and \(w'\) is its
sibling. The forest averages
these per-tree maxima.
\end{corollary}
\begin{proof}
For displacement, the mass increments along the path are the
masses of the siblings that join it. Apply \eqref{eq:heights}.
For CoDisp, contraction of the departing subtree's parent
reduces the sibling's depths by one and leaves all other
surviving depths fixed, giving the saving \(b\).

The ratio has a walk interpretation on the mass path as well.
Place a ground below \(x\), and give each path vertex cumulative
resistance equal to its subtree mass, now including \(x\).
From \(w\), the probability of reaching its parent before
ground is \(p=a/(a+b)\). The complementary odds are
\((1-p)/p=b/a\).
\end{proof}

\begin{corollary}[Affine control of the three escapes]
\label{cor:affineescapes}
Fix the stored sample $S$ and an unseen query $x$, and put
$\delta=\min_{y\in S}\|x-y\|_1>0$,
$\Delta=\operatorname{diam}_1(S\cup\{x\})$, and
$C_x=R(\bbox(S\cup\{x\}))$.
Suppose $S\cup\{x\}$ has affine dimension $r\geq1$, and set
\[
 a=2\mathbb K_{r,1}(1/2)
       \left[1+\left\lceil\log_2(\Delta/\delta)\right\rceil\right].
\]
Draw $T$ on $S$ and then the virtual attachment of $x$.
Let $D_x$ be its augmented depth, $R_*$ the range of its first
ancestor containing stored records, and $m_*$ the stored mass at
that ancestor. The expectations below include both cuts and attachment.
The depth and range escapes satisfy
\begin{equation}
 \begin{gathered}
 \E D_x\leq a,\qquad
 \E_T U(T,x)=\E\frac1{D_x}\geq\frac1a,\\
 \frac{\delta}{C_x}\leq\frac{\E R_*}{C_x}
          \leq\min\left\{1,\frac{a\delta}{C_x}\right\}.
 \end{gathered}
 \label{eq:affineescapes}
\end{equation}
Let $M_x(t)$ be the stored mass within $\ell_1$ distance $t$ of
$x$, and let $\psi$ be the total stored mass. For every $t>0$,
the mass escape satisfies
\begin{equation}
 \frac{\E m_*}{\psi}
 \leq\min\left\{1,\frac{M_x(t)}{\psi}+\frac{a\delta}{t}\right\}.
 \label{eq:affinemassescape}
\end{equation}
If instead the augmented sample meets the noise model of
Corollary~\ref{cor:noisygeometry} and $\delta\geq16W$, the same
statements hold with $\mathbb K_r^{\rm noise}(1/2)$ in place of
$\mathbb K_{r,1}(1/2)$, using the clean affine dimension $r$.
\end{corollary}
\begin{proof}
Insertion consistency makes the augmented tree a fresh random-cut
tree on $S\cup\{x\}$. Apply the conditional contraction blocks of
Theorem~\ref{prop:affinegeometry} along the cell containing $x$.
After $1+\lceil\log_2(\Delta/\delta)\rceil$ successful blocks its
diameter is less than $\delta$, so it contains no stored record.
The expected depth is at most $a$, and convexity of $1/D_x$ gives
the depth escape bound. If $y$ is a nearest stored location, then
$\delta\leq R_*\leq d_T(x,y)$, so the stretch theorem and
$R_*\leq C_x$ give the range bounds. On $R_*\leq t$, every
record at the first ancestor lies within distance $t$ of $x$;
hence $m_*\leq M_x(t)$. On the complementary event use
$m_*\leq\psi$ and Markov's inequality
$\Prob(R_*>t)\leq\E R_*/t\leq a\delta/t$.
The noisy contraction and stretch bounds apply by the same argument
while the common cell has diameter at least $\delta\geq16W$.
\end{proof}

Affine dimension thus controls how many decisions are needed to
resolve the nearest gap and the range at which the query first
meets the sample. The mass reading also depends on how much mass
lies at that scale. Copies change this occupancy without changing
the affine geometry. An arbitrary query increases the affine
dimension of the stored sample by at most one. Below the noise
scale, Theorem~\ref{thm:geometryenergy} supplies an additive range
bound; the depth bound above requires resolution beyond that scale.

\subsection{Nonparametric Escape and Near Duplicates}
\label{sec:games}
Exact duplicates add mass without increasing spatial extent.
Near duplicates concentrate mass in a small range. A declared
mass chooses how much support is required before a query reads
its escape. These statements concern the records retained in
each sampled tree; the sampling law determines how often a cloud
supplies that support.

\begin{definition}[Ancestor at a declared mass]
\label{def:massselector}
On a complete query path $v_0,\ldots,v_h$ of root mass
$\psi\geq1$, the mass selector $c_\mu(x)$ is the deepest vertex
containing at least mass $\mu$, in record units. For $0<\mu\leq\psi$,
\begin{equation}
 i_\mu=\max\{i:m_{v_i}\geq\mu\},\qquad c_\mu(x)=v_{i_\mu}.
 \label{eq:massselector}
\end{equation}
\end{definition}

\paragraph{Escape at a declared mass.}
A query can separate beside a few near duplicates even when the
cloud is far from the rest of the sample. Its separation box can
then have only the range of the cloud, making ordinary range
escape small. Choose a minimum mass $\mu$ and read the escape
at the first ancestor with that much support. This is an
adaptive-neighbor construction in the tree's range metric.
It keeps the sample and the separation law fixed; it requires
one query path and no coalition search. The calibration box of
Section~\ref{sec:density} uses mass in the same way to set the
scale of a comparison.

For an integer $1\leq\mu\leq\psi$, use the mass selector
$c_\mu(x)$ of Definition~\ref{def:massselector}.
Here $\mu$ is a minimum number of stored records.
Stop the separation experiment at $c_\mu(x)$: keep separations
above it, and use its range for all outcomes that reach it.
With $v$ the separating vertex under \eqref{eq:measure} and
$v\vee c$ the higher of two vertices on the path, put
\begin{equation}
 \operatorname{Esc}_\mu(T,x)
 =\frac{\E\,R\bigl(B_{v\vee c_\mu(x)}\cup\{x\}\bigr)}
        {C},\qquad C=R(B_{\rm root}\cup\{x\}).
 \label{eq:massescape}
\end{equation}
For a query that matches a stored location, the terminal
outcome uses the range at $c_\mu(x)$. Set the measurement to
zero when $C=0$.

In the tree augmented by virtual insertion, the numerator
reads the range of the first ancestor of $x$ containing at least
$\mu$ original records, counted with multiplicity. It is the
distance to the $\mu$th stored neighbor in the augmented tree's
range metric \eqref{eq:rcfrangemetric}. Equivalently, put the
query at voltage zero and the root at one, using cumulative
range as resistance. The reading is the voltage at this ancestor,
as in \eqref{eq:heights}; Section~\ref{sec:game} develops the
corresponding potential game.

\begin{theorem}[Escape at a declared mass]
\label{prop:massescape}
Fix the stored sample and query $x$, and let $D$ be a box
containing $x$, with $\delta=R(D)$. Parts (i) and (ii) concern
a fixed tree. Parts (iii)--(v) use the random-cut construction
with coordinate choice proportional to width and a uniform
threshold in the chosen coordinate.
\begin{enumerate}\setlength{\itemsep}{1pt}
\item[(i)] $\operatorname{Esc}_1$ is the range escape
\eqref{eq:rangeescape}, and $\operatorname{Esc}_\mu$ is
nondecreasing in $\mu$. Its numerator is computed by the fold
\eqref{eq:recurrence}, stopped at $c_\mu(x)$.

\item[(ii)] Suppose $C>0$. If $D$ contains fewer than $\mu$ stored records and
every stored point $y$ outside it satisfies
$\sum_j\operatorname{dist}(y_j,D_j)\geq\Gamma$, then
$\operatorname{Esc}_\mu(T,x)\geq\Gamma/C$.

\item[(iii)] Suppose there are stored records both inside and
outside $D$, and the sum $\Gamma'$ of the coordinate gaps
between $D$ and the bounding box of the outside records is
positive. With probability at least
$\Gamma'/(\Gamma'+\delta)$, the query's branch reaches a node
containing just the records in $D$ before any cut on that
branch crosses $D$.

\item[(iv)] Under the gap condition of (iii), if $D$ contains
at least $\mu$ stored records, then
\[
 \E_T\operatorname{Esc}_\mu(T,x)
 \leq \frac{\delta}{C}+\frac{\delta}{\Gamma'+\delta}.
\]

\item[(v)] Let $D$ be the bounding box of $x$ and a group of
at least $\mu$ stored records. For each coordinate and side,
let $b_j^\pm$ be the distance from the face of $D$ to the
corresponding face of the root box with $x$ included. If
$\delta>0$, then
\[
 \E_T\operatorname{Esc}_\mu(T,x)
 \leq \frac{\delta}{C}
 \left(1+\sum_{j,\pm}\log\!\left(1+\frac{b_j^\pm}{\delta}\right)\right).
\]
If $\delta=0$, the measurement is zero. No gap between the
bounding boxes is required.
\end{enumerate}
\end{theorem}

\begin{proof}
(i) For $\mu=1$, $c_1(x)$ is the routed leaf, so the stopping
rule leaves the original range escape unchanged. Increasing
$\mu$ moves $c_\mu(x)$ up the path, and the boxes are nested.
To compute the numerator, initialize the fold at $c_\mu(x)$
with $R(B_{c_\mu(x)}\cup\{x\})$ and use the original separation
payoffs above it. In the augmented tree, if $v$ already contains
$\mu$ stored records, the new parent of $x$ is the first
ancestor meeting the mass requirement. Otherwise that ancestor
is $c_\mu(x)$, whose box now includes $x$. At a match the new
copy shares the original leaf. This gives the stated
tree-neighbor interpretation in each case.

(ii) Every box used for a payoff contains at least $\mu$ stored
records, so it contains some $y$ outside $D$. Its range is at
least $\sum_j|x_j-y_j|\geq\sum_j\operatorname{dist}(y_j,D_j)$,
which proves the floor.

(iii) At a node still containing the cloud and some outside
records, the available separating gaps have total length at
least $\Gamma'$, while thresholds through $D$ have total
length at most $\delta$. Cuts elsewhere remove only outside
records, which cannot shrink the gaps. Conditional on the
first cut in either a separating gap or $D$, it falls in a
gap with probability at least $\Gamma'/(\Gamma'+\delta)$.
Such a cut leaves the records of $D$ together on the query's
branch.

(iv) The insertion law gives a random-cut tree on the augmented
sample \citep{guha2016rrcf}; adding a copy at a matching
location changes masses without changing cuts. Apply (iii) to
this augmented sample. On the event that the cloud is isolated
intact, the first ancestor containing $\mu$ original records
lies within the cloud's subtree and has range at most $\delta$.
The normalized payoff is at most $\delta/C$ on this event and
at most one otherwise.

(v) Realize the cuts by an auxiliary Poisson process on position
and time, with unit intensity over the augmented root box's
coordinate intervals laid end to end. On the query branch, accept
the next mark in the current box's intervals. Its position is
uniform over their total length, giving width-proportional cuts.
Disjoint intervals have independent marks; positions of total
length $r$ have no mark by time $t$ with probability $e^{-tr}$
\citep{last2017poisson}. The first mark through $D$ occurs at time
$\tau\sim\operatorname{Exp}(\delta)$. Until that mark,
the cloud stays together, so the ancestor used by the
measurement is no higher than the node where this cut occurs.

Let $Z_j^\pm$ be that node's extent beyond the corresponding
face of $D$. For $Z_j^\pm$ to exceed $u$, there can have been
no mark before $\tau$ in the adjoining interval of length $u$:
such a mark would have cut off the records beyond it. These
marks are independent of $\tau$, so
\[
 \E Z_j^\pm
 \leq \int_0^{b_j^\pm}\E e^{-u\tau}\,du
 =\delta\log\!\left(1+\frac{b_j^\pm}{\delta}\right).
\]
The node's range is $\delta+\sum_{j,\pm}Z_j^\pm$. Taking
expectations and dividing by $C$ gives the bound. If
$\delta=0$, at least $\mu$ stored records coincide with $x$
and the mass requirement is met at its leaf.
\end{proof}

The two regimes give $\mu$ its meaning. If a cloud contains
fewer than $\mu$ retained records and is separated from the
remaining data, part (ii) gives a floor in every tree. If it
contains at least $\mu$ records, parts (iv)--(v) give vanishing
mean escape as $\delta/C$ tends to zero in fixed dimension.
Thus near-duplicate resilience is a statement at a declared
mass resolution. Once enough copies are present, they supply
the requested support. The threshold applies to each tree's
retained sample, and increasing it can raise the reading for
ordinary points as well as anomalies.

The random-tree bounds use width-proportional cuts.
Section~\ref{sec:cutgeometry} explains why uniform coordinate allocation
can retain a nonvanishing share of cuts on shrinking nuisance coordinates.

\begin{corollary}[Noisy copies at a declared mass]
\label{cor:noisycopies}
Suppose the query and $k\geq1$ stored records are perturbations of
one clean location. The other records have a clean bounding box
at summed coordinate gap $\Gamma_0>2W$ from that location.
Every record and the query are perturbed in coordinate $j$ by
at most $\sigma_j$, where $W=\sum_j\sigma_j$ as in
Corollary~\ref{cor:noisygeometry}. Write
$C=R(B_{\rm root}\cup\{x\})$. For $1\leq\mu\leq\psi$,
\[
 \operatorname{Esc}_\mu(T,x)\geq\frac{\Gamma_0-2W}{C}
       \quad\hbox{if }k<\mu,
 \qquad
 \E_T\operatorname{Esc}_\mu(T,x)
       \leq\frac{2W}{C}+\frac{2W}{\Gamma_0}
       \quad\hbox{if }k\geq\mu.
\]
The first inequality holds on every tree; the expectation in the
second uses width-proportional cuts conditional on the noisy sample.
\end{corollary}
\begin{proof}
The box of the query and its $k$ nearby records has range at
most $2W$. Its summed coordinate gap to the other records' box
is at least $\Gamma_0-2W$. Apply parts (ii) and (iv) of
Theorem~\ref{prop:massescape}; the second bound uses
$\delta/(\Gamma'+\delta)\leq2W/\Gamma_0$.
\end{proof}
Thus insufficient nearby mass forces the reading across the gap,
while sufficient nearby mass permits a reading on the noise-sized
cloud. Sampling changes the retained count $k$ and the gap; the
corollary applies to each sample meeting its conditions.

\subsection{Geometry and Regularization of the Hitting Bundle}
\label{sec:density}\label{sec:faces}
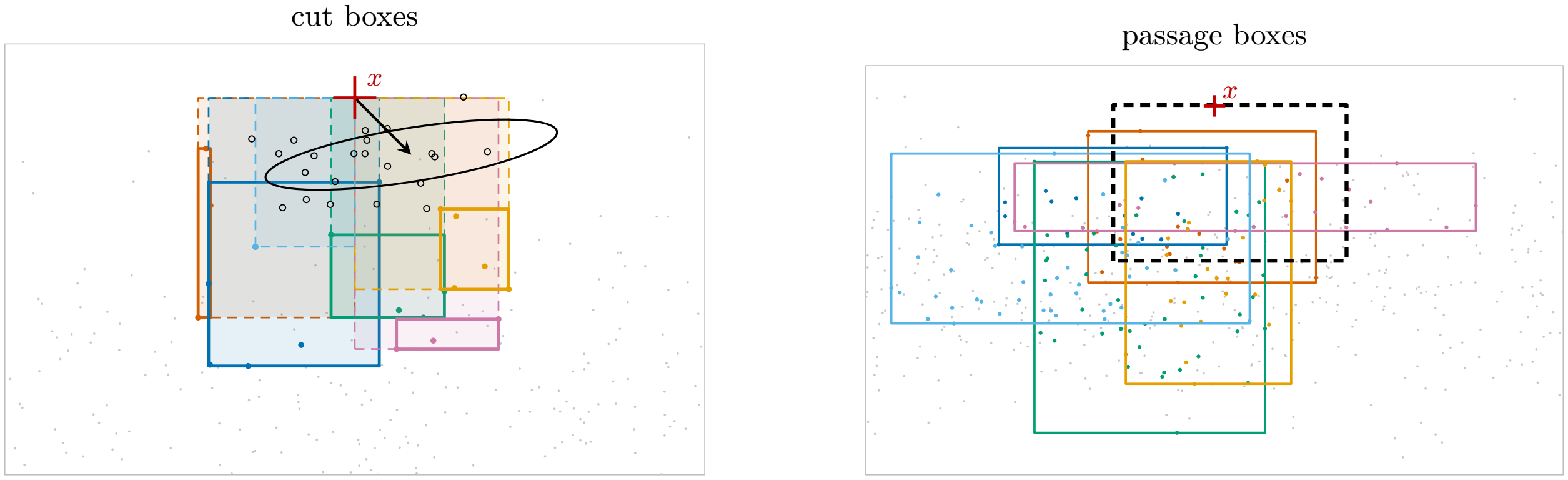
\begin{figure}[!htb]
\centering
\definecolor{tree0}{HTML}{D55E00}
\definecolor{tree1}{HTML}{0072B2}
\definecolor{tree2}{HTML}{009E73}
\definecolor{tree3}{HTML}{CC79A7}
\definecolor{tree4}{HTML}{E69F00}
\definecolor{tree5}{HTML}{56B4E9}
\begin{tikzpicture}[x=2.85cm,y=2.85cm,line join=round]
\node[font=\footnotesize,anchor=south] at (0.00,1.52) {cut boxes};
\draw[black!25,line width=0.3pt] (-1.30,-0.10) rectangle (1.30,1.50);
\begin{scope}
\clip (-1.30,-0.10) rectangle (1.30,1.50);
\fill[black!22] (-0.981,0.391) circle (0.35pt);
\fill[black!22] (-0.104,0.501) circle (0.35pt);
\fill[black!22] (1.139,-0.098) circle (0.35pt);
\fill[black!22] (0.440,-0.065) circle (0.35pt);
\fill[black!22] (-0.765,1.102) circle (0.35pt);
\fill[black!22] (1.041,0.152) circle (0.35pt);
\fill[black!22] (0.187,0.295) circle (0.35pt);
\fill[black!22] (1.038,0.069) circle (0.35pt);
\fill[black!22] (-1.084,0.018) circle (0.35pt);
\fill[black!22] (-0.015,-0.091) circle (0.35pt);
\fill[black!22] (0.651,0.134) circle (0.35pt);
\fill[black!22] (0.876,0.023) circle (0.35pt);
\fill[black!22] (1.163,0.859) circle (0.35pt);
\fill[black!22] (0.424,0.492) circle (0.35pt);
\fill[black!22] (0.542,0.387) circle (0.35pt);
\fill[black!22] (-0.171,0.050) circle (0.35pt);
\fill[black!22] (1.151,0.046) circle (0.35pt);
\fill[black!22] (1.279,0.159) circle (0.35pt);
\fill[black!22] (0.295,-0.035) circle (0.35pt);
\fill[black!22] (0.235,0.009) circle (0.35pt);
\fill[black!22] (0.074,-0.001) circle (0.35pt);
\fill[black!22] (-1.052,0.422) circle (0.35pt);
\fill[black!22] (-0.354,0.251) circle (0.35pt);
\fill[black!22] (0.519,0.665) circle (0.35pt);
\fill[black!22] (0.770,0.021) circle (0.35pt);
\fill[black!22] (-0.211,0.196) circle (0.35pt);
\fill[black!22] (0.905,0.162) circle (0.35pt);
\fill[black!22] (1.165,0.300) circle (0.35pt);
\fill[black!22] (-0.530,0.223) circle (0.35pt);
\fill[black!22] (0.382,0.224) circle (0.35pt);
\fill[black!22] (-0.563,0.561) circle (0.35pt);
\fill[black!22] (-0.379,-0.015) circle (0.35pt);
\fill[black!22] (0.943,0.386) circle (0.35pt);
\fill[black!22] (-0.457,-0.096) circle (0.35pt);
\fill[black!22] (0.787,0.172) circle (0.35pt);
\fill[black!22] (-0.141,0.030) circle (0.35pt);
\fill[black!22] (0.070,0.664) circle (0.35pt);
\fill[black!22] (0.087,-0.099) circle (0.35pt);
\fill[black!22] (1.188,0.099) circle (0.35pt);
\fill[black!22] (-0.536,0.900) circle (0.35pt);
\fill[black!22] (0.859,0.016) circle (0.35pt);
\fill[black!22] (0.985,0.442) circle (0.35pt);
\fill[black!22] (-0.553,1.112) circle (0.35pt);
\fill[black!22] (-0.481,1.002) circle (0.35pt);
\fill[black!22] (-0.225,0.440) circle (0.35pt);
\fill[black!22] (0.265,0.084) circle (0.35pt);
\fill[black!22] (-0.567,0.539) circle (0.35pt);
\fill[black!22] (1.237,0.615) circle (0.35pt);
\fill[black!22] (0.190,0.042) circle (0.35pt);
\fill[black!22] (0.852,0.094) circle (0.35pt);
\fill[black!22] (1.042,-0.019) circle (0.35pt);
\fill[black!22] (0.061,-0.082) circle (0.35pt);
\fill[black!22] (0.970,0.164) circle (0.35pt);
\fill[black!22] (0.853,0.292) circle (0.35pt);
\fill[black!22] (-1.094,0.330) circle (0.35pt);
\fill[black!22] (-1.162,0.535) circle (0.35pt);
\fill[black!22] (-0.875,0.363) circle (0.35pt);
\fill[black!22] (0.658,0.215) circle (0.35pt);
\fill[black!22] (0.133,0.104) circle (0.35pt);
\fill[black!22] (1.200,0.350) circle (0.35pt);
\fill[black!22] (-1.010,-0.032) circle (0.35pt);
\fill[black!22] (-1.004,0.664) circle (0.35pt);
\fill[black!22] (0.167,0.001) circle (0.35pt);
\fill[black!22] (-1.124,0.433) circle (0.35pt);
\fill[black!22] (0.157,0.140) circle (0.35pt);
\fill[black!22] (0.111,0.037) circle (0.35pt);
\fill[black!22] (0.230,-0.095) circle (0.35pt);
\fill[black!22] (-0.921,0.158) circle (0.35pt);
\fill[black!22] (1.015,0.256) circle (0.35pt);
\fill[black!22] (0.319,0.887) circle (0.35pt);
\fill[black!22] (0.869,0.134) circle (0.35pt);
\fill[black!22] (0.811,0.284) circle (0.35pt);
\fill[black!22] (0.309,0.696) circle (0.35pt);
\fill[black!22] (0.128,0.243) circle (0.35pt);
\fill[black!22] (-0.829,0.641) circle (0.35pt);
\fill[black!22] (0.036,0.747) circle (0.35pt);
\fill[black!22] (0.971,0.402) circle (0.35pt);
\fill[black!22] (0.978,0.725) circle (0.35pt);
\fill[black!22] (0.838,0.028) circle (0.35pt);
\fill[black!22] (-1.113,0.069) circle (0.35pt);
\fill[black!22] (1.229,0.071) circle (0.35pt);
\fill[black!22] (0.596,0.651) circle (0.35pt);
\fill[black!22] (0.243,1.072) circle (0.35pt);
\fill[black!22] (0.309,0.275) circle (0.35pt);
\fill[black!22] (0.988,0.725) circle (0.35pt);
\fill[black!22] (-0.164,1.091) circle (0.35pt);
\fill[black!22] (1.122,0.859) circle (0.35pt);
\fill[black!22] (-0.356,0.189) circle (0.35pt);
\fill[black!22] (-0.886,0.053) circle (0.35pt);
\fill[black!22] (-1.030,0.611) circle (0.35pt);
\fill[black!22] (0.907,0.626) circle (0.35pt);
\fill[black!22] (-0.508,0.299) circle (0.35pt);
\fill[black!22] (-1.263,-0.002) circle (0.35pt);
\fill[black!22] (-0.693,0.272) circle (0.35pt);
\fill[black!22] (-0.822,0.990) circle (0.35pt);
\fill[black!22] (0.227,-0.075) circle (0.35pt);
\fill[black!22] (-0.865,-0.042) circle (0.35pt);
\fill[black!22] (-1.235,0.149) circle (0.35pt);
\fill[black!22] (-0.538,0.309) circle (0.35pt);
\fill[black!22] (-0.369,0.747) circle (0.35pt);
\fill[black!22] (0.333,0.582) circle (0.35pt);
\fill[black!22] (0.338,0.225) circle (0.35pt);
\fill[black!22] (-0.501,-0.064) circle (0.35pt);
\fill[black!22] (1.171,0.079) circle (0.35pt);
\fill[black!22] (-1.245,0.163) circle (0.35pt);
\fill[black!22] (0.093,0.067) circle (0.35pt);
\fill[black!22] (0.685,0.773) circle (0.35pt);
\fill[black!22] (-0.950,0.093) circle (0.35pt);
\fill[black!22] (0.960,0.247) circle (0.35pt);
\fill[black!22] (0.216,0.205) circle (0.35pt);
\fill[black!22] (0.534,0.478) circle (0.35pt);
\fill[black!22] (0.379,0.429) circle (0.35pt);
\fill[black!22] (-0.271,0.496) circle (0.35pt);
\fill[black!22] (-0.118,-0.090) circle (0.35pt);
\fill[black!22] (-1.233,1.062) circle (0.35pt);
\fill[black!22] (0.814,0.146) circle (0.35pt);
\fill[black!22] (0.037,0.824) circle (0.35pt);
\fill[black!22] (0.475,0.507) circle (0.35pt);
\fill[black!22] (-0.503,0.091) circle (0.35pt);
\fill[black!22] (-0.681,0.393) circle (0.35pt);
\fill[black!22] (-0.291,0.075) circle (0.35pt);
\fill[black!22] (0.017,-0.066) circle (0.35pt);
\fill[black!22] (-0.080,0.059) circle (0.35pt);
\fill[black!22] (0.213,0.279) circle (0.35pt);
\fill[black!22] (-1.072,-0.018) circle (0.35pt);
\fill[black!22] (1.152,0.123) circle (0.35pt);
\fill[black!22] (0.524,0.216) circle (0.35pt);
\fill[black!22] (0.273,0.558) circle (0.35pt);
\fill[black!22] (0.868,0.314) circle (0.35pt);
\fill[black!22] (-0.087,0.241) circle (0.35pt);
\fill[black!22] (0.336,0.132) circle (0.35pt);
\fill[black!22] (-0.968,0.424) circle (0.35pt);
\fill[black!22] (0.852,0.349) circle (0.35pt);
\fill[black!22] (0.017,0.014) circle (0.35pt);
\fill[black!22] (-0.742,0.357) circle (0.35pt);
\fill[black!22] (0.699,1.291) circle (0.35pt);
\fill[black!22] (0.886,0.512) circle (0.35pt);
\fill[black!22] (-0.146,0.678) circle (0.35pt);
\fill[black!22] (1.265,0.230) circle (0.35pt);
\fill[black!22] (1.165,0.586) circle (0.35pt);
\fill[black!22] (-0.371,-0.012) circle (0.35pt);
\fill[black!22] (0.959,0.595) circle (0.35pt);
\fill[black!22] (-0.135,0.264) circle (0.35pt);
\fill[black!22] (0.744,0.000) circle (0.35pt);
\fill[black!22] (0.101,0.307) circle (0.35pt);
\fill[black!22] (-0.567,0.280) circle (0.35pt);
\fill[black!22] (0.490,0.849) circle (0.35pt);
\fill[black!22] (0.532,0.470) circle (0.35pt);
\fill[black!22] (-0.427,0.068) circle (0.35pt);
\fill[black!22] (1.014,0.541) circle (0.35pt);
\fill[black!22] (1.144,0.421) circle (0.35pt);
\fill[black!22] (-0.031,0.199) circle (0.35pt);
\fill[black!22] (-0.919,0.505) circle (0.35pt);
\fill[black!22] (0.039,0.212) circle (0.35pt);
\fill[black!22] (-0.209,0.170) circle (0.35pt);
\fill[black!22] (-0.205,0.145) circle (0.35pt);
\fill[black!22] (-0.684,-0.001) circle (0.35pt);
\fill[black!22] (1.151,0.209) circle (0.35pt);
\fill[black!22] (-1.280,0.256) circle (0.35pt);
\fill[black!22] (0.742,0.145) circle (0.35pt);
\fill[black!22] (0.043,-0.042) circle (0.35pt);
\fill[black!22] (-0.393,0.177) circle (0.35pt);
\fill[black!22] (-0.971,0.124) circle (0.35pt);
\fill[black!22] (0.278,0.814) circle (0.35pt);
\fill[black!22] (0.499,0.587) circle (0.35pt);
\fill[black!22] (0.027,0.161) circle (0.35pt);
\fill[black!22] (1.242,1.153) circle (0.35pt);
\fill[black!22] (0.061,0.177) circle (0.35pt);
\fill[black!22] (0.758,0.212) circle (0.35pt);
\fill[black!22] (-1.175,0.553) circle (0.35pt);
\fill[black!22] (0.707,-0.031) circle (0.35pt);
\fill[black!22] (0.810,0.048) circle (0.35pt);
\fill[black!22] (-0.838,0.061) circle (0.35pt);
\fill[black!22] (-0.045,0.487) circle (0.35pt);
\fill[black!22] (-1.172,0.285) circle (0.35pt);
\fill[black!22] (0.255,0.484) circle (0.35pt);
\fill[black!22] (0.582,0.512) circle (0.35pt);
\fill[black!22] (-0.011,0.399) circle (0.35pt);
\fill[black!22] (0.198,0.019) circle (0.35pt);
\fill[black!22] (-0.199,0.382) circle (0.35pt);
\fill[black!22] (-0.182,0.299) circle (0.35pt);
\fill[black!22] (-1.019,0.522) circle (0.35pt);
\fill[black!22] (0.613,0.294) circle (0.35pt);
\fill[black!22] (1.287,0.376) circle (0.35pt);
\fill[black!22] (1.276,0.498) circle (0.35pt);
\fill[black!22] (0.763,0.135) circle (0.35pt);
\fill[black!22] (0.227,0.369) circle (0.35pt);
\fill[black!22] (-0.396,0.304) circle (0.35pt);
\fill[black!22] (-1.065,0.361) circle (0.35pt);
\fill[black!22] (-0.900,-0.006) circle (0.35pt);
\fill[black!22] (0.578,0.078) circle (0.35pt);
\fill[black!22] (0.153,0.395) circle (0.35pt);
\fill[black!22] (0.610,0.137) circle (0.35pt);
\fill[black!22] (-0.669,0.480) circle (0.35pt);
\fill[black!22] (0.493,0.453) circle (0.35pt);
\fill[black!22] (0.913,0.861) circle (0.35pt);
\fill[black!22] (-0.647,0.218) circle (0.35pt);
\fill[black!22] (0.376,0.860) circle (0.35pt);
\fill[black!22] (-1.194,0.945) circle (0.35pt);
\fill[black!22] (-0.377,0.544) circle (0.35pt);
\fill[black!22] (-0.702,0.306) circle (0.35pt);
\fill[black!22] (-1.227,-0.063) circle (0.35pt);
\fill[black!22] (0.685,0.225) circle (0.35pt);
\fill[black!22] (1.004,0.675) circle (0.35pt);
\fill[black!22] (-0.272,-0.017) circle (0.35pt);
\fill[black!22] (0.013,0.030) circle (0.35pt);
\fill[black!22] (0.093,0.304) circle (0.35pt);
\fill[black!22] (0.416,0.127) circle (0.35pt);
\fill[black!22] (0.551,0.415) circle (0.35pt);
\fill[black!22] (0.800,0.758) circle (0.35pt);
\fill[black!22] (0.401,0.531) circle (0.35pt);
\fill[black!22] (0.838,0.107) circle (0.35pt);
\fill[black!22] (-0.873,0.461) circle (0.35pt);
\fill[black!22] (-0.052,0.232) circle (0.35pt);
\fill[black!22] (1.079,0.675) circle (0.35pt);
\fill[black!22] (-0.528,1.062) circle (0.35pt);
\fill[black!22] (-0.707,0.561) circle (0.35pt);
\fill[black!22] (1.158,0.314) circle (0.35pt);
\fill[black!22] (-0.778,0.270) circle (0.35pt);
\fill[black!22] (-1.106,0.410) circle (0.35pt);
\fill[tree0,opacity=0.10] (-0.582,0.484) rectangle (0.000,1.300);
\draw[tree0,densely dashed,line width=0.5pt] (-0.582,0.484) rectangle (0.000,1.300);
\draw[tree0,line width=0.9pt] (-0.582,0.484) rectangle (-0.536,1.112);
\fill[tree0] (-0.582,0.484) circle (0.9pt);
\fill[tree0] (-0.553,1.112) circle (0.9pt);
\fill[tree0] (-0.536,0.900) circle (0.9pt);
\fill[tree1,opacity=0.10] (-0.543,0.304) rectangle (0.091,1.300);
\draw[tree1,densely dashed,line width=0.5pt] (-0.543,0.304) rectangle (0.091,1.300);
\draw[tree1,line width=0.9pt] (-0.543,0.304) rectangle (0.091,0.987);
\fill[tree1] (-0.369,0.747) circle (0.9pt);
\fill[tree1] (0.091,0.987) circle (0.9pt);
\fill[tree1] (-0.538,0.309) circle (0.9pt);
\fill[tree1] (-0.199,0.382) circle (0.9pt);
\fill[tree1] (-0.543,0.610) circle (0.9pt);
\fill[tree1] (-0.396,0.304) circle (0.9pt);
\fill[tree2,opacity=0.10] (-0.088,0.484) rectangle (0.333,1.300);
\draw[tree2,densely dashed,line width=0.5pt] (-0.088,0.484) rectangle (0.333,1.300);
\draw[tree2,line width=0.9pt] (-0.088,0.484) rectangle (0.333,0.791);
\fill[tree2] (0.255,0.484) circle (0.9pt);
\fill[tree2] (-0.088,0.791) circle (0.9pt);
\fill[tree2] (0.164,0.511) circle (0.9pt);
\fill[tree2] (0.333,0.582) circle (0.9pt);
\fill[tree3,opacity=0.10] (0.000,0.367) rectangle (0.534,1.300);
\draw[tree3,densely dashed,line width=0.5pt] (0.000,0.367) rectangle (0.534,1.300);
\draw[tree3,line width=0.9pt] (0.155,0.367) rectangle (0.534,0.478);
\fill[tree3] (0.155,0.367) circle (0.9pt);
\fill[tree3] (0.534,0.478) circle (0.9pt);
\fill[tree3] (0.292,0.398) circle (0.9pt);
\fill[tree4,opacity=0.10] (0.000,0.589) rectangle (0.572,1.300);
\draw[tree4,densely dashed,line width=0.5pt] (0.000,0.589) rectangle (0.572,1.300);
\draw[tree4,line width=0.9pt] (0.319,0.589) rectangle (0.572,0.887);
\fill[tree4] (0.376,0.860) circle (0.9pt);
\fill[tree4] (0.483,0.674) circle (0.9pt);
\fill[tree4] (0.319,0.887) circle (0.9pt);
\fill[tree4] (0.572,0.589) circle (0.9pt);
\fill[tree4] (0.370,0.594) circle (0.9pt);
\fill[tree5,opacity=0.10] (-0.369,0.747) rectangle (0.000,1.300);
\draw[tree5,densely dashed,line width=0.5pt] (-0.369,0.747) rectangle (0.000,1.300);
\draw[tree5,line width=0.9pt] (-0.369,0.747) rectangle (-0.369,0.747);
\fill[tree5] (-0.369,0.747) circle (0.9pt);
% the swing: covariance of the 25 gap-box centres at one standard deviation, about the end of the lead
\draw[black,line width=0.35pt] (-0.268,0.892) circle (0.9pt);
\draw[black,line width=0.35pt] (-0.226,1.143) circle (0.9pt);
\draw[black,line width=0.35pt] (0.122,1.046) circle (0.9pt);
\draw[black,line width=0.35pt] (0.267,0.889) circle (0.9pt);
\draw[black,line width=0.35pt] (0.286,1.093) circle (0.9pt);
\draw[black,line width=0.35pt] (-0.184,1.023) circle (0.9pt);
\draw[black,line width=0.35pt] (1.470,1.245) circle (0.9pt);
\draw[black,line width=0.35pt] (-0.151,1.085) circle (0.9pt);
\draw[black,line width=0.35pt] (-0.091,0.904) circle (0.9pt);
\draw[black,line width=0.35pt] (0.245,0.983) circle (0.9pt);
\draw[black,line width=0.35pt] (1.605,1.358) circle (0.9pt);
\draw[black,line width=0.35pt] (-0.383,1.148) circle (0.9pt);
\draw[black,line width=0.35pt] (1.579,1.323) circle (0.9pt);
\draw[black,line width=0.35pt] (0.039,1.179) circle (0.9pt);
\draw[black,line width=0.35pt] (0.038,1.093) circle (0.9pt);
\draw[black,line width=0.35pt] (0.493,1.100) circle (0.9pt);
\draw[black,line width=0.35pt] (0.121,1.186) circle (0.9pt);
\draw[black,line width=0.35pt] (0.404,1.303) circle (0.9pt);
\draw[black,line width=0.35pt] (-0.073,0.989) circle (0.9pt);
\draw[black,line width=0.35pt] (-0.003,1.093) circle (0.9pt);
\draw[black,line width=0.35pt] (0.297,1.081) circle (0.9pt);
\draw[black,line width=0.35pt] (0.082,0.905) circle (0.9pt);
\draw[black,line width=0.35pt] (-0.282,1.093) circle (0.9pt);
\draw[black,line width=0.35pt] (-0.180,0.922) circle (0.9pt);
\draw[black,line width=0.35pt] (0.045,1.143) circle (0.9pt);
\draw[black,line width=0.6pt,rotate around={9.03:(0.210,1.089)}] (0.210,1.089) ellipse (0.548 and 0.0996);
\draw[->,black,line width=0.8pt,>=stealth] (0.000,1.300) -- ++(0.210,-0.211);
\end{scope}
\draw[red!75!black,line width=0.9pt] (0.000,1.300) ++(-0.08,0)--++(0.16,0) (0.000,1.300) ++(0,-0.08)--++(0,0.16);
\node[font=\scriptsize,anchor=south west,red!75!black,inner sep=1pt] at (0.030,1.330) {$x$};
\end{tikzpicture}
\hfill
\begin{tikzpicture}[x=1.42cm,y=1.42cm,line join=round]
\node[font=\footnotesize,anchor=south] at (0.00,1.62) {passage boxes};
\draw[black!25,line width=0.3pt] (-2.60,-1.45) rectangle (2.60,1.60);
\begin{scope}
\clip (-2.60,-1.45) rectangle (2.60,1.60);
\fill[black!22] (-2.542,-0.430) circle (0.35pt);
\fill[black!22] (2.293,0.407) circle (0.35pt);
\fill[black!22] (2.080,-0.727) circle (0.35pt);
\fill[black!22] (0.018,0.319) circle (0.35pt);
\fill[black!22] (1.213,-0.050) circle (0.35pt);
\fill[black!22] (-0.393,0.177) circle (0.35pt);
\fill[black!22] (1.446,0.789) circle (0.35pt);
\fill[black!22] (-2.348,0.235) circle (0.35pt);
\fill[black!22] (-1.601,-0.250) circle (0.35pt);
\fill[black!22] (1.838,-0.016) circle (0.35pt);
\fill[black!22] (-1.483,0.159) circle (0.35pt);
\fill[black!22] (-0.966,-0.370) circle (0.35pt);
\fill[black!22] (-0.396,0.304) circle (0.35pt);
\fill[black!22] (-2.535,-0.707) circle (0.35pt);
\fill[black!22] (-2.213,-0.580) circle (0.35pt);
\fill[black!22] (-1.509,-0.019) circle (0.35pt);
\fill[black!22] (1.166,0.343) circle (0.35pt);
\fill[black!22] (-1.758,-0.213) circle (0.35pt);
\fill[black!22] (-1.655,0.090) circle (0.35pt);
\fill[black!22] (-2.279,0.458) circle (0.35pt);
\fill[black!22] (-1.438,-0.298) circle (0.35pt);
\fill[black!22] (-1.873,0.603) circle (0.35pt);
\fill[black!22] (2.596,0.359) circle (0.35pt);
\fill[black!22] (-2.207,0.524) circle (0.35pt);
\fill[black!22] (1.440,0.185) circle (0.35pt);
\fill[black!22] (0.535,-0.252) circle (0.35pt);
\fill[black!22] (-0.968,0.424) circle (0.35pt);
\fill[black!22] (0.292,0.398) circle (0.35pt);
\fill[black!22] (-0.926,-0.264) circle (0.35pt);
\fill[black!22] (1.270,0.683) circle (0.35pt);
\fill[black!22] (0.093,0.067) circle (0.35pt);
\fill[black!22] (-2.518,-0.638) circle (0.35pt);
\fill[black!22] (-1.840,0.348) circle (0.35pt);
\fill[black!22] (-1.280,0.256) circle (0.35pt);
\fill[black!22] (-1.815,-0.705) circle (0.35pt);
\fill[black!22] (1.273,0.053) circle (0.35pt);
\fill[black!22] (2.340,0.209) circle (0.35pt);
\fill[black!22] (1.281,0.080) circle (0.35pt);
\fill[black!22] (-2.044,-0.955) circle (0.35pt);
\fill[black!22] (2.180,-0.640) circle (0.35pt);
\fill[black!22] (0.876,0.023) circle (0.35pt);
\fill[black!22] (1.518,-0.370) circle (0.35pt);
\fill[black!22] (-2.189,1.055) circle (0.35pt);
\fill[black!22] (0.379,0.429) circle (0.35pt);
\fill[black!22] (-0.688,0.207) circle (0.35pt);
\fill[black!22] (-0.748,-0.245) circle (0.35pt);
\fill[black!22] (2.107,-0.160) circle (0.35pt);
\fill[black!22] (0.303,-0.106) circle (0.35pt);
\fill[black!22] (-2.176,-0.180) circle (0.35pt);
\fill[black!22] (0.033,-0.643) circle (0.35pt);
\fill[black!22] (-1.846,0.278) circle (0.35pt);
\fill[black!22] (1.302,0.174) circle (0.35pt);
\fill[black!22] (-0.474,-0.299) circle (0.35pt);
\fill[black!22] (-0.536,0.900) circle (0.35pt);
\fill[black!22] (-1.072,-0.619) circle (0.35pt);
\fill[black!22] (1.848,-0.464) circle (0.35pt);
\fill[black!22] (0.445,-0.808) circle (0.35pt);
\fill[black!22] (1.167,-0.932) circle (0.35pt);
\fill[black!22] (0.155,0.613) circle (0.35pt);
\fill[black!22] (0.289,-0.481) circle (0.35pt);
\fill[black!22] (0.579,-0.221) circle (0.35pt);
\fill[black!22] (-2.071,0.181) circle (0.35pt);
\fill[black!22] (0.985,0.442) circle (0.35pt);
\fill[black!22] (-0.807,-0.105) circle (0.35pt);
\fill[black!22] (-2.434,0.377) circle (0.35pt);
\fill[black!22] (0.061,0.177) circle (0.35pt);
\fill[black!22] (-1.425,-0.013) circle (0.35pt);
\fill[black!22] (-1.615,0.516) circle (0.35pt);
\fill[black!22] (0.004,0.342) circle (0.35pt);
\fill[black!22] (1.930,0.848) circle (0.35pt);
\fill[black!22] (-2.406,0.622) circle (0.35pt);
\fill[black!22] (-1.290,-0.681) circle (0.35pt);
\fill[black!22] (-2.531,-1.145) circle (0.35pt);
\fill[black!22] (-0.466,-0.270) circle (0.35pt);
\fill[black!22] (-0.811,-0.008) circle (0.35pt);
\fill[black!22] (-1.645,0.306) circle (0.35pt);
\fill[black!22] (-1.550,-0.136) circle (0.35pt);
\fill[black!22] (-0.901,-0.416) circle (0.35pt);
\fill[black!22] (-0.970,-0.481) circle (0.35pt);
\fill[black!22] (0.227,0.369) circle (0.35pt);
\fill[black!22] (2.433,0.216) circle (0.35pt);
\fill[black!22] (-0.805,-0.712) circle (0.35pt);
\fill[black!22] (1.434,0.804) circle (0.35pt);
\fill[black!22] (-0.298,-0.137) circle (0.35pt);
\fill[black!22] (1.525,0.642) circle (0.35pt);
\fill[black!22] (1.514,-0.641) circle (0.35pt);
\fill[black!22] (-0.089,0.870) circle (0.35pt);
\fill[black!22] (-1.665,-0.469) circle (0.35pt);
\fill[black!22] (1.931,0.947) circle (0.35pt);
\fill[black!22] (-2.058,0.638) circle (0.35pt);
\fill[black!22] (-1.223,-0.208) circle (0.35pt);
\fill[black!22] (0.867,0.184) circle (0.35pt);
\fill[black!22] (-2.132,-1.062) circle (0.35pt);
\fill[black!22] (1.405,-1.320) circle (0.35pt);
\fill[black!22] (-0.384,0.228) circle (0.35pt);
\fill[black!22] (2.433,-0.099) circle (0.35pt);
\fill[black!22] (-0.654,0.997) circle (0.35pt);
\fill[black!22] (0.490,0.849) circle (0.35pt);
\fill[black!22] (-2.234,0.051) circle (0.35pt);
\fill[black!22] (-0.454,-0.240) circle (0.35pt);
\fill[black!22] (-0.886,0.053) circle (0.35pt);
\fill[black!22] (-0.633,0.028) circle (0.35pt);
\fill[black!22] (-1.173,0.017) circle (0.35pt);
\fill[black!22] (-0.503,-0.035) circle (0.35pt);
\fill[black!22] (-2.317,-0.105) circle (0.35pt);
\fill[black!22] (-0.035,0.507) circle (0.35pt);
\fill[black!22] (0.651,0.134) circle (0.35pt);
\fill[black!22] (0.551,0.415) circle (0.35pt);
\fill[black!22] (1.243,0.928) circle (0.35pt);
\fill[black!22] (-0.647,0.218) circle (0.35pt);
\fill[black!22] (-0.270,-0.819) circle (0.35pt);
\fill[black!22] (2.206,-0.598) circle (0.35pt);
\fill[black!22] (2.400,-0.514) circle (0.35pt);
\fill[black!22] (-1.942,0.068) circle (0.35pt);
\fill[black!22] (-2.137,-0.289) circle (0.35pt);
\fill[black!22] (-2.587,-0.088) circle (0.35pt);
\fill[black!22] (-2.511,0.511) circle (0.35pt);
\fill[black!22] (1.704,-0.541) circle (0.35pt);
\fill[black!22] (2.576,-0.394) circle (0.35pt);
\fill[black!22] (-0.899,0.350) circle (0.35pt);
\fill[black!22] (-2.033,-0.368) circle (0.35pt);
\fill[black!22] (0.865,0.417) circle (0.35pt);
\fill[black!22] (-0.554,-0.693) circle (0.35pt);
\fill[black!22] (2.432,-0.001) circle (0.35pt);
\fill[black!22] (0.527,-0.556) circle (0.35pt);
\fill[black!22] (-0.625,0.304) circle (0.35pt);
\fill[black!22] (-0.962,-0.733) circle (0.35pt);
\fill[black!22] (1.984,-0.368) circle (0.35pt);
\fill[black!22] (2.375,-0.115) circle (0.35pt);
\fill[black!22] (0.834,-0.164) circle (0.35pt);
\fill[black!22] (0.770,0.021) circle (0.35pt);
\fill[black!22] (0.852,0.094) circle (0.35pt);
\fill[black!22] (0.280,-0.627) circle (0.35pt);
\fill[black!22] (0.183,0.134) circle (0.35pt);
\fill[black!22] (-1.411,0.906) circle (0.35pt);
\fill[black!22] (-1.964,0.077) circle (0.35pt);
\fill[black!22] (0.151,-0.223) circle (0.35pt);
\fill[black!22] (-2.152,0.233) circle (0.35pt);
\fill[black!22] (2.237,1.351) circle (0.35pt);
\fill[black!22] (0.912,-0.232) circle (0.35pt);
\fill[black!22] (-1.273,0.218) circle (0.35pt);
\fill[black!22] (2.244,0.556) circle (0.35pt);
\fill[black!22] (-2.048,-0.094) circle (0.35pt);
\fill[black!22] (-0.127,-0.672) circle (0.35pt);
\fill[black!22] (-2.499,-0.326) circle (0.35pt);
\fill[black!22] (-1.057,0.480) circle (0.35pt);
\fill[black!22] (0.317,-0.414) circle (0.35pt);
\fill[black!22] (2.143,-0.086) circle (0.35pt);
\fill[black!22] (1.954,0.260) circle (0.35pt);
\fill[black!22] (-2.480,0.176) circle (0.35pt);
\fill[black!22] (-2.596,-0.122) circle (0.35pt);
\fill[black!22] (1.867,-0.654) circle (0.35pt);
\fill[black!22] (1.488,0.707) circle (0.35pt);
\fill[black!22] (-1.748,0.809) circle (0.35pt);
\fill[black!22] (0.082,0.387) circle (0.35pt);
\fill[black!22] (-0.846,-0.356) circle (0.35pt);
\fill[black!22] (-1.192,-0.451) circle (0.35pt);
\fill[black!22] (-0.657,-0.274) circle (0.35pt);
\fill[black!22] (0.754,-0.763) circle (0.35pt);
\fill[black!22] (0.243,0.768) circle (0.35pt);
\fill[black!22] (1.036,0.749) circle (0.35pt);
\fill[black!22] (-0.480,-0.359) circle (0.35pt);
\fill[black!22] (0.732,-0.163) circle (0.35pt);
\fill[black!22] (2.527,0.470) circle (0.35pt);
\fill[black!22] (1.647,-0.208) circle (0.35pt);
\fill[black!22] (-0.858,-0.180) circle (0.35pt);
\fill[black!22] (0.630,-0.319) circle (0.35pt);
\fill[black!22] (-1.297,-0.392) circle (0.35pt);
\fill[black!22] (0.077,-0.123) circle (0.35pt);
\fill[black!22] (-2.475,-0.243) circle (0.35pt);
\fill[black!22] (1.055,-0.296) circle (0.35pt);
\fill[black!22] (-0.671,-0.360) circle (0.35pt);
\fill[black!22] (0.708,0.022) circle (0.35pt);
\fill[black!22] (1.198,0.257) circle (0.35pt);
\fill[black!22] (-0.693,-0.204) circle (0.35pt);
\fill[black!22] (2.527,-0.238) circle (0.35pt);
\fill[black!22] (0.959,0.595) circle (0.35pt);
\fill[black!22] (1.723,-0.814) circle (0.35pt);
\fill[black!22] (-0.904,-0.421) circle (0.35pt);
\fill[black!22] (1.765,-0.036) circle (0.35pt);
\fill[black!22] (2.504,-0.393) circle (0.35pt);
\fill[black!22] (0.971,0.402) circle (0.35pt);
\fill[black!22] (0.061,-0.082) circle (0.35pt);
\fill[black!22] (0.067,-0.178) circle (0.35pt);
\fill[black!22] (-1.232,-0.300) circle (0.35pt);
\fill[black!22] (2.221,0.384) circle (0.35pt);
\fill[black!22] (0.943,0.386) circle (0.35pt);
\fill[black!22] (-0.644,-0.645) circle (0.35pt);
\fill[black!22] (2.337,-0.275) circle (0.35pt);
\fill[black!22] (0.685,0.186) circle (0.35pt);
\fill[black!22] (-2.095,0.135) circle (0.35pt);
\fill[black!22] (-0.191,0.430) circle (0.35pt);
\fill[black!22] (-1.943,-0.322) circle (0.35pt);
\fill[black!22] (-1.392,-0.494) circle (0.35pt);
\fill[black!22] (-1.294,-0.158) circle (0.35pt);
\fill[black!22] (-0.257,-0.668) circle (0.35pt);
\fill[black!22] (-1.901,-0.415) circle (0.35pt);
\fill[black!22] (-1.849,-0.460) circle (0.35pt);
\fill[black!22] (-1.258,0.146) circle (0.35pt);
\fill[black!22] (2.342,0.232) circle (0.35pt);
\fill[black!22] (0.838,0.107) circle (0.35pt);
\fill[black!22] (-0.776,-0.176) circle (0.35pt);
\fill[black!22] (-1.555,-0.188) circle (0.35pt);
\fill[black!22] (-1.911,1.141) circle (0.35pt);
\fill[black!22] (2.385,0.946) circle (0.35pt);
\fill[black!22] (0.544,0.104) circle (0.35pt);
\fill[black!22] (2.369,-0.049) circle (0.35pt);
\fill[black!22] (-2.541,0.451) circle (0.35pt);
\fill[black!22] (-1.892,-0.559) circle (0.35pt);
\fill[black!22] (-0.045,0.487) circle (0.35pt);
\fill[black!22] (1.903,0.446) circle (0.35pt);
\fill[black!22] (1.260,-0.218) circle (0.35pt);
\fill[black!22] (-1.356,0.430) circle (0.35pt);
\fill[black!22] (1.042,-0.019) circle (0.35pt);
\fill[black!22] (2.410,-0.764) circle (0.35pt);
\fill[black!22] (1.151,0.046) circle (0.35pt);
\fill[black!22] (-0.446,-0.337) circle (0.35pt);
\fill[black!22] (0.507,0.471) circle (0.35pt);
\fill[black!22] (-0.423,0.063) circle (0.35pt);
\fill[black!22] (1.347,-1.179) circle (0.35pt);
\fill[black!22] (-2.583,-1.056) circle (0.35pt);
\fill[black!22] (0.924,-0.043) circle (0.35pt);
\fill[black!22] (0.532,-0.567) circle (0.35pt);
\fill[black!22] (1.649,-1.149) circle (0.35pt);
\fill[black!22] (-1.435,-0.243) circle (0.35pt);
\fill[black!22] (-0.544,-0.692) circle (0.35pt);
\fill[black!22] (-1.964,0.061) circle (0.35pt);
\fill[black!22] (1.710,-0.260) circle (0.35pt);
\fill[black!22] (0.610,0.137) circle (0.35pt);
\fill[black!22] (2.220,0.229) circle (0.35pt);
\fill[black!22] (-2.038,0.333) circle (0.35pt);
\fill[black!22] (1.670,0.532) circle (0.35pt);
\fill[black!22] (-2.549,0.391) circle (0.35pt);
\fill[black!22] (-0.786,0.476) circle (0.35pt);
\fill[black!22] (2.213,0.341) circle (0.35pt);
\fill[black!22] (-1.495,-0.517) circle (0.35pt);
\fill[black!22] (2.297,-0.053) circle (0.35pt);
\fill[black!22] (-0.521,-0.562) circle (0.35pt);
\fill[black!22] (-0.712,-0.599) circle (0.35pt);
\fill[black!22] (-0.187,-1.031) circle (0.35pt);
\fill[black!22] (-2.456,-0.227) circle (0.35pt);
\fill[black!22] (-0.112,0.114) circle (0.35pt);
\fill[black!22] (-1.487,-0.908) circle (0.35pt);
\fill[black!22] (1.743,0.172) circle (0.35pt);
\fill[black!22] (-1.692,0.165) circle (0.35pt);
\fill[black!22] (0.265,0.084) circle (0.35pt);
\fill[black!22] (1.053,-0.555) circle (0.35pt);
\fill[black!22] (-1.887,0.585) circle (0.35pt);
\fill[black!22] (-1.879,-0.198) circle (0.35pt);
\fill[black!22] (2.404,-0.194) circle (0.35pt);
\fill[black!22] (0.120,-0.007) circle (0.35pt);
\fill[black!22] (-0.689,-0.186) circle (0.35pt);
\fill[black!22] (-2.058,-0.747) circle (0.35pt);
\fill[black!22] (-1.711,0.433) circle (0.35pt);
\fill[black!22] (-1.245,0.163) circle (0.35pt);
\fill[black!22] (2.012,0.763) circle (0.35pt);
\fill[black!22] (-2.302,0.993) circle (0.35pt);
\fill[black!22] (0.334,-0.605) circle (0.35pt);
\fill[black!22] (-0.674,0.595) circle (0.35pt);
\fill[black!22] (1.540,-0.094) circle (0.35pt);
\fill[black!22] (2.430,0.730) circle (0.35pt);
\fill[black!22] (2.446,-0.110) circle (0.35pt);
\fill[black!22] (1.760,-0.503) circle (0.35pt);
\fill[black!22] (-1.738,-0.159) circle (0.35pt);
\fill[black!22] (0.110,-0.358) circle (0.35pt);
\fill[black!22] (-1.571,0.683) circle (0.35pt);
\fill[black!22] (2.293,-0.627) circle (0.35pt);
\fill[black!22] (0.440,-0.204) circle (0.35pt);
\fill[black!22] (2.389,0.293) circle (0.35pt);
\fill[black!22] (1.516,0.794) circle (0.35pt);
\fill[black!22] (-0.625,0.084) circle (0.35pt);
\fill[black!22] (0.913,0.861) circle (0.35pt);
\fill[black!22] (2.517,0.582) circle (0.35pt);
\fill[black!22] (-1.643,-0.160) circle (0.35pt);
\fill[black!22] (0.204,-0.346) circle (0.35pt);
\fill[black!22] (2.545,0.854) circle (0.35pt);
\fill[black!22] (-2.051,0.415) circle (0.35pt);
\fill[black!22] (-0.225,0.440) circle (0.35pt);
\fill[black!22] (0.582,-0.443) circle (0.35pt);
\fill[black!22] (0.375,-0.628) circle (0.35pt);
\fill[black!22] (-2.507,0.164) circle (0.35pt);
\fill[black!22] (1.158,0.314) circle (0.35pt);
\fill[black!22] (-2.587,0.096) circle (0.35pt);
\fill[black!22] (1.981,0.199) circle (0.35pt);
\fill[black!22] (0.582,0.512) circle (0.35pt);
\fill[black!22] (0.588,-1.019) circle (0.35pt);
\fill[black!22] (2.284,-0.386) circle (0.35pt);
\fill[black!22] (1.540,-1.062) circle (0.35pt);
\fill[black!22] (0.037,0.824) circle (0.35pt);
\fill[black!22] (2.455,-0.421) circle (0.35pt);
\fill[black!22] (-0.389,-0.719) circle (0.35pt);
\fill[black!22] (1.580,-0.085) circle (0.35pt);
\fill[black!22] (1.525,-0.051) circle (0.35pt);
\fill[black!22] (-2.082,-0.570) circle (0.35pt);
\fill[black!22] (1.200,0.350) circle (0.35pt);
\fill[black!22] (-1.468,-0.176) circle (0.35pt);
\fill[black!22] (0.275,-0.636) circle (0.35pt);
\fill[black!22] (-2.014,-0.347) circle (0.35pt);
\fill[black!22] (-0.082,0.990) circle (0.35pt);
\fill[black!22] (-0.103,0.183) circle (0.35pt);
\fill[black!22] (-1.752,0.366) circle (0.35pt);
\fill[black!22] (-1.446,-0.467) circle (0.35pt);
\fill[black!22] (2.372,-0.608) circle (0.35pt);
\fill[black!22] (-2.502,-0.821) circle (0.35pt);
\fill[black!22] (1.360,0.872) circle (0.35pt);
\fill[black!22] (-1.830,0.268) circle (0.35pt);
\fill[black!22] (-0.209,0.170) circle (0.35pt);
\fill[black!22] (1.029,-0.346) circle (0.35pt);
\fill[black!22] (-2.516,-0.572) circle (0.35pt);
\fill[black!22] (-2.309,1.370) circle (0.35pt);
\fill[black!22] (2.573,-0.856) circle (0.35pt);
\fill[black!22] (-0.818,-0.290) circle (0.35pt);
\fill[black!22] (0.211,-0.359) circle (0.35pt);
\fill[black!22] (0.433,-0.511) circle (0.35pt);
\fill[black!22] (-0.141,0.030) circle (0.35pt);
\fill[black!22] (-0.981,0.391) circle (0.35pt);
\fill[black!22] (-2.196,-0.495) circle (0.35pt);
\fill[black!22] (-2.314,-0.714) circle (0.35pt);
\fill[black!22] (1.407,0.779) circle (0.35pt);
\fill[black!22] (1.418,-0.106) circle (0.35pt);
\fill[black!22] (-0.985,-0.219) circle (0.35pt);
\fill[black!22] (-0.358,-1.004) circle (0.35pt);
\fill[black!22] (2.231,0.503) circle (0.35pt);
\fill[black!22] (0.259,0.644) circle (0.35pt);
\fill[black!22] (1.121,0.546) circle (0.35pt);
\fill[black!22] (-2.458,0.072) circle (0.35pt);
\fill[black!22] (1.423,-0.728) circle (0.35pt);
\fill[black!22] (0.987,1.056) circle (0.35pt);
\fill[black!22] (-1.072,-0.018) circle (0.35pt);
\fill[black!22] (-2.154,0.711) circle (0.35pt);
\fill[black!22] (-0.441,0.261) circle (0.35pt);
\fill[black!22] (1.266,-0.194) circle (0.35pt);
\fill[black!22] (-2.090,0.226) circle (0.35pt);
\fill[black!22] (-2.483,-0.067) circle (0.35pt);
\fill[black!22] (-1.890,-0.442) circle (0.35pt);
\fill[black!22] (-2.211,0.122) circle (0.35pt);
\fill[black!22] (1.685,-0.408) circle (0.35pt);
\fill[black!22] (-2.017,0.044) circle (0.35pt);
\fill[black!22] (1.172,-0.156) circle (0.35pt);
\fill[black!22] (0.171,-0.381) circle (0.35pt);
\fill[black!22] (1.883,0.325) circle (0.35pt);
\fill[black!22] (0.013,0.030) circle (0.35pt);
\fill[black!22] (-0.790,-0.303) circle (0.35pt);
\fill[black!22] (1.229,0.071) circle (0.35pt);
\fill[black!22] (0.606,-0.183) circle (0.35pt);
\fill[black!22] (-1.868,-0.902) circle (0.35pt);
\fill[black!22] (-0.651,0.414) circle (0.35pt);
\fill[black!22] (-0.243,-0.824) circle (0.35pt);
\fill[black!22] (-1.898,-0.419) circle (0.35pt);
\fill[black!22] (-2.393,0.443) circle (0.35pt);
\fill[black!22] (0.927,-0.496) circle (0.35pt);
\fill[black!22] (-0.087,0.241) circle (0.35pt);
\fill[black!22] (-0.009,-0.463) circle (0.35pt);
\fill[black!22] (0.519,0.665) circle (0.35pt);
\fill[black!22] (0.859,0.016) circle (0.35pt);
\fill[black!22] (-0.763,0.048) circle (0.35pt);
\fill[black!22] (2.245,0.137) circle (0.35pt);
\fill[black!22] (1.507,-0.064) circle (0.35pt);
\fill[black!22] (-0.742,0.357) circle (0.35pt);
\fill[black!22] (1.973,-0.573) circle (0.35pt);
\fill[black!22] (0.960,0.247) circle (0.35pt);
\fill[black!22] (-2.101,0.320) circle (0.35pt);
\fill[black!22] (-0.052,0.085) circle (0.35pt);
\fill[black!22] (1.730,0.395) circle (0.35pt);
\fill[black!22] (-0.123,-0.235) circle (0.35pt);
\fill[black!22] (-2.514,1.357) circle (0.35pt);
\fill[black!22] (-0.553,1.112) circle (0.35pt);
\fill[black!22] (1.270,-0.122) circle (0.35pt);
\fill[black!22] (-1.408,-0.203) circle (0.35pt);
\fill[black!22] (0.382,0.289) circle (0.35pt);
\fill[black!22] (-0.822,0.990) circle (0.35pt);
\fill[black!22] (0.788,-0.133) circle (0.35pt);
\fill[black!22] (-1.605,0.197) circle (0.35pt);
\fill[black!22] (1.670,-0.048) circle (0.35pt);
\fill[black!22] (-0.052,0.232) circle (0.35pt);
\fill[black!22] (-1.018,-0.328) circle (0.35pt);
\fill[black!22] (-0.298,-0.168) circle (0.35pt);
\fill[black!22] (0.919,-0.182) circle (0.35pt);
\fill[black!22] (1.612,0.451) circle (0.35pt);
\fill[black!22] (-1.194,0.945) circle (0.35pt);
\fill[black!22] (0.966,0.444) circle (0.35pt);
\fill[black!22] (-1.465,0.633) circle (0.35pt);
\fill[black!22] (-1.182,-0.672) circle (0.35pt);
\fill[black!22] (-0.873,0.461) circle (0.35pt);
\fill[black!22] (0.838,0.028) circle (0.35pt);
\fill[black!22] (0.331,-0.117) circle (0.35pt);
\fill[black!22] (-1.198,0.252) circle (0.35pt);
\fill[black!22] (-0.693,0.272) circle (0.35pt);
\fill[black!22] (1.697,-1.209) circle (0.35pt);
\fill[black!22] (-0.907,0.258) circle (0.35pt);
\fill[black!22] (-1.958,-0.275) circle (0.35pt);
\fill[black!22] (-0.732,0.258) circle (0.35pt);
\fill[black!22] (-0.016,-0.773) circle (0.35pt);
\fill[black!22] (-1.801,-0.079) circle (0.35pt);
\fill[black!22] (1.946,-0.017) circle (0.35pt);
\fill[black!22] (1.575,-0.807) circle (0.35pt);
\fill[black!22] (0.869,0.134) circle (0.35pt);
\fill[black!22] (1.647,-0.109) circle (0.35pt);
\fill[black!22] (-0.481,1.002) circle (0.35pt);
\fill[black!22] (0.984,0.393) circle (0.35pt);
\fill[black!22] (0.587,-1.101) circle (0.35pt);
\fill[black!22] (2.270,0.373) circle (0.35pt);
\fill[black!22] (-0.698,-0.362) circle (0.35pt);
\fill[black!22] (-2.194,0.462) circle (0.35pt);
\fill[black!22] (1.164,0.871) circle (0.35pt);
\fill[black!22] (1.311,-0.058) circle (0.35pt);
\fill[black!22] (-1.710,0.151) circle (0.35pt);
\fill[black!22] (0.169,-0.420) circle (0.35pt);
\fill[black!22] (1.557,0.260) circle (0.35pt);
\fill[black!22] (1.339,-0.249) circle (0.35pt);
\fill[black!22] (0.222,-0.791) circle (0.35pt);
\fill[black!22] (-1.238,-0.309) circle (0.35pt);
\fill[black!22] (-0.080,-0.694) circle (0.35pt);
\fill[black!22] (-2.586,0.185) circle (0.35pt);
\fill[black!22] (-1.302,-0.466) circle (0.35pt);
\fill[black!22] (-1.004,-0.138) circle (0.35pt);
\fill[black!22] (1.040,0.313) circle (0.35pt);
\fill[black!22] (-0.929,-0.809) circle (0.35pt);
\fill[black!22] (-2.131,0.002) circle (0.35pt);
\fill[black!22] (-2.365,-0.037) circle (0.35pt);
\draw[tree0,line width=0.7pt] (-0.941,-0.017) rectangle (0.758,1.112);
\fill[tree0] (-0.702,0.306) circle (0.6pt);
\fill[tree0] (-0.333,0.195) circle (0.6pt);
\fill[tree0] (-0.681,0.393) circle (0.6pt);
\fill[tree0] (0.366,0.037) circle (0.6pt);
\fill[tree0] (-0.272,-0.017) circle (0.6pt);
\fill[tree0] (-0.272,0.400) circle (0.6pt);
\fill[tree0] (0.072,0.237) circle (0.6pt);
\fill[tree0] (-0.838,0.061) circle (0.6pt);
\fill[tree0] (0.483,0.674) circle (0.6pt);
\fill[tree0] (0.541,0.744) circle (0.6pt);
\fill[tree0] (-0.114,0.347) circle (0.6pt);
\fill[tree0] (-0.582,0.484) circle (0.6pt);
\fill[tree0] (0.758,0.018) circle (0.6pt);
\fill[tree0] (0.333,0.582) circle (0.6pt);
\fill[tree0] (-0.354,0.251) circle (0.6pt);
\fill[tree0] (0.157,0.140) circle (0.6pt);
\fill[tree0] (-0.305,0.771) circle (0.6pt);
\fill[tree0] (0.183,0.134) circle (0.6pt);
\fill[tree0] (-0.941,1.080) circle (0.6pt);
\fill[tree0] (-0.875,0.363) circle (0.6pt);
\fill[tree0] (-0.553,1.112) circle (0.6pt);
\fill[tree0] (-0.536,0.900) circle (0.6pt);
\draw[tree1,line width=0.7pt] (-1.608,0.267) rectangle (0.091,0.987);
\fill[tree1] (-0.271,0.496) circle (0.6pt);
\fill[tree1] (-0.369,0.747) circle (0.6pt);
\fill[tree1] (-1.608,0.267) circle (0.6pt);
\fill[tree1] (-1.561,0.581) circle (0.6pt);
\fill[tree1] (-0.899,0.350) circle (0.6pt);
\fill[tree1] (-1.098,0.348) circle (0.6pt);
\fill[tree1] (0.091,0.987) circle (0.6pt);
\fill[tree1] (-0.538,0.309) circle (0.6pt);
\fill[tree1] (-0.669,0.480) circle (0.6pt);
\fill[tree1] (-0.606,0.345) circle (0.6pt);
\fill[tree1] (-0.199,0.382) circle (0.6pt);
\fill[tree1] (-0.543,0.610) circle (0.6pt);
\fill[tree1] (-1.215,0.587) circle (0.6pt);
\fill[tree1] (-1.258,0.665) circle (0.6pt);
\fill[tree1] (-1.561,0.477) circle (0.6pt);
\fill[tree1] (-1.030,0.611) circle (0.6pt);
\fill[tree1] (-0.396,0.304) circle (0.6pt);
\draw[tree2,line width=0.7pt] (-1.342,-1.137) rectangle (0.377,0.884);
\fill[tree2] (-0.698,-0.362) circle (0.6pt);
\fill[tree2] (-0.837,-0.397) circle (0.6pt);
\fill[tree2] (-0.087,-0.130) circle (0.6pt);
\fill[tree2] (0.255,0.484) circle (0.6pt);
\fill[tree2] (-0.707,0.561) circle (0.6pt);
\fill[tree2] (-1.245,0.163) circle (0.6pt);
\fill[tree2] (0.251,-0.765) circle (0.6pt);
\fill[tree2] (-0.517,-0.324) circle (0.6pt);
\fill[tree2] (-0.119,-0.568) circle (0.6pt);
\fill[tree2] (-0.370,-0.289) circle (0.6pt);
\fill[tree2] (-0.981,0.391) circle (0.6pt);
\fill[tree2] (0.292,0.398) circle (0.6pt);
\fill[tree2] (-0.389,-0.719) circle (0.6pt);
\fill[tree2] (-0.088,0.791) circle (0.6pt);
\fill[tree2] (-0.559,-0.497) circle (0.6pt);
\fill[tree2] (-0.951,0.228) circle (0.6pt);
\fill[tree2] (0.164,0.511) circle (0.6pt);
\fill[tree2] (-0.644,-0.645) circle (0.6pt);
\fill[tree2] (-1.297,-0.392) circle (0.6pt);
\fill[tree2] (-0.087,0.241) circle (0.6pt);
\fill[tree2] (-0.103,0.183) circle (0.6pt);
\fill[tree2] (-1.192,-0.451) circle (0.6pt);
\fill[tree2] (-0.225,0.440) circle (0.6pt);
\fill[tree2] (-1.258,0.146) circle (0.6pt);
\fill[tree2] (-1.342,0.884) circle (0.6pt);
\fill[tree2] (-0.899,0.350) circle (0.6pt);
\fill[tree2] (-0.279,-1.137) circle (0.6pt);
\fill[tree2] (-0.954,0.044) circle (0.6pt);
\fill[tree2] (-0.732,0.258) circle (0.6pt);
\fill[tree2] (0.333,0.582) circle (0.6pt);
\fill[tree2] (-0.375,-0.688) circle (0.6pt);
\fill[tree2] (-0.661,-0.554) circle (0.6pt);
\fill[tree2] (-1.263,-0.002) circle (0.6pt);
\fill[tree2] (-0.305,0.771) circle (0.6pt);
\fill[tree2] (-0.631,-0.379) circle (0.6pt);
\fill[tree2] (-0.582,0.484) circle (0.6pt);
\fill[tree2] (-1.246,0.336) circle (0.6pt);
\fill[tree2] (-0.257,-0.668) circle (0.6pt);
\fill[tree2] (-0.681,0.393) circle (0.6pt);
\fill[tree2] (-0.985,-0.219) circle (0.6pt);
\fill[tree2] (-1.094,0.330) circle (0.6pt);
\fill[tree2] (0.189,-0.256) circle (0.6pt);
\fill[tree2] (-0.141,0.030) circle (0.6pt);
\fill[tree2] (-0.718,-0.147) circle (0.6pt);
\fill[tree2] (0.377,-0.361) circle (0.6pt);
\fill[tree2] (0.366,0.037) circle (0.6pt);
\fill[tree2] (0.171,-0.381) circle (0.6pt);
\fill[tree2] (0.287,-0.124) circle (0.6pt);
\draw[tree3,line width=0.7pt] (-1.490,0.367) rectangle (1.949,0.872);
\fill[tree3] (0.155,0.367) circle (0.6pt);
\fill[tree3] (-0.567,0.539) circle (0.6pt);
\fill[tree3] (1.360,0.872) circle (0.6pt);
\fill[tree3] (1.165,0.586) circle (0.6pt);
\fill[tree3] (-0.981,0.391) circle (0.6pt);
\fill[tree3] (1.004,0.675) circle (0.6pt);
\fill[tree3] (0.800,0.758) circle (0.6pt);
\fill[tree3] (-1.413,0.399) circle (0.6pt);
\fill[tree3] (-0.298,0.872) circle (0.6pt);
\fill[tree3] (-0.707,0.561) circle (0.6pt);
\fill[tree3] (1.949,0.556) circle (0.6pt);
\fill[tree3] (0.636,0.793) circle (0.6pt);
\fill[tree3] (0.534,0.478) circle (0.6pt);
\fill[tree3] (1.287,0.376) circle (0.6pt);
\fill[tree3] (0.984,0.393) circle (0.6pt);
\fill[tree3] (-1.490,0.436) circle (0.6pt);
\fill[tree3] (0.292,0.398) circle (0.6pt);
\fill[tree3] (1.730,0.395) circle (0.6pt);
\fill[tree3] (0.751,0.508) circle (0.6pt);
\draw[tree4,line width=0.7pt] (-0.661,-0.772) rectangle (0.572,0.887);
\fill[tree4] (-0.325,-0.156) circle (0.6pt);
\fill[tree4] (-0.101,-0.309) circle (0.6pt);
\fill[tree4] (0.101,0.013) circle (0.6pt);
\fill[tree4] (0.376,0.860) circle (0.6pt);
\fill[tree4] (-0.149,-0.772) circle (0.6pt);
\fill[tree4] (-0.199,0.382) circle (0.6pt);
\fill[tree4] (0.017,0.014) circle (0.6pt);
\fill[tree4] (-0.661,-0.554) circle (0.6pt);
\fill[tree4] (-0.226,-0.164) circle (0.6pt);
\fill[tree4] (0.483,0.674) circle (0.6pt);
\fill[tree4] (-0.210,-0.612) circle (0.6pt);
\fill[tree4] (-0.356,0.189) circle (0.6pt);
\fill[tree4] (-0.247,-0.340) circle (0.6pt);
\fill[tree4] (0.303,-0.106) circle (0.6pt);
\fill[tree4] (0.407,-0.331) circle (0.6pt);
\fill[tree4] (0.213,0.318) circle (0.6pt);
\fill[tree4] (-0.345,0.312) circle (0.6pt);
\fill[tree4] (-0.291,0.075) circle (0.6pt);
\fill[tree4] (0.087,-0.099) circle (0.6pt);
\fill[tree4] (0.319,0.887) circle (0.6pt);
\fill[tree4] (-0.191,0.430) circle (0.6pt);
\fill[tree4] (-0.052,0.085) circle (0.6pt);
\fill[tree4] (0.572,0.589) circle (0.6pt);
\fill[tree4] (0.370,0.594) circle (0.6pt);
\draw[tree5,line width=0.7pt] (-2.410,-0.322) rectangle (0.263,0.945);
\fill[tree5] (-1.274,-0.225) circle (0.6pt);
\fill[tree5] (-0.826,-0.229) circle (0.6pt);
\fill[tree5] (-2.410,-0.057) circle (0.6pt);
\fill[tree5] (-2.078,-0.121) circle (0.6pt);
\fill[tree5] (-1.194,0.945) circle (0.6pt);
\fill[tree5] (-2.137,-0.289) circle (0.6pt);
\fill[tree5] (-1.665,0.355) circle (0.6pt);
\fill[tree5] (-1.181,-0.264) circle (0.6pt);
\fill[tree5] (-0.776,-0.176) circle (0.6pt);
\fill[tree5] (-1.636,0.255) circle (0.6pt);
\fill[tree5] (-1.964,0.077) circle (0.6pt);
\fill[tree5] (-1.771,-0.143) circle (0.6pt);
\fill[tree5] (-0.818,-0.290) circle (0.6pt);
\fill[tree5] (-1.222,-0.106) circle (0.6pt);
\fill[tree5] (0.101,0.307) circle (0.6pt);
\fill[tree5] (-0.291,0.075) circle (0.6pt);
\fill[tree5] (-0.681,0.393) circle (0.6pt);
\fill[tree5] (-2.058,0.638) circle (0.6pt);
\fill[tree5] (-1.010,-0.032) circle (0.6pt);
\fill[tree5] (-1.029,-0.097) circle (0.6pt);
\fill[tree5] (-0.688,0.207) circle (0.6pt);
\fill[tree5] (-1.738,-0.159) circle (0.6pt);
\fill[tree5] (-1.601,-0.250) circle (0.6pt);
\fill[tree5] (-2.209,0.173) circle (0.6pt);
\fill[tree5] (-1.173,0.017) circle (0.6pt);
\fill[tree5] (-0.567,0.280) circle (0.6pt);
\fill[tree5] (-0.693,-0.204) circle (0.6pt);
\fill[tree5] (-1.434,0.252) circle (0.6pt);
\fill[tree5] (-1.813,0.407) circle (0.6pt);
\fill[tree5] (-1.094,0.093) circle (0.6pt);
\fill[tree5] (-1.457,-0.147) circle (0.6pt);
\fill[tree5] (-1.943,-0.322) circle (0.6pt);
\fill[tree5] (-0.369,0.747) circle (0.6pt);
\fill[tree5] (0.263,-0.304) circle (0.6pt);
\fill[tree5] (-2.199,0.432) circle (0.6pt);
\fill[tree5] (-1.219,-0.228) circle (0.6pt);
\fill[tree5] (-0.503,0.091) circle (0.6pt);
\fill[tree5] (-2.347,-0.092) circle (0.6pt);
\fill[tree5] (-0.649,0.186) circle (0.6pt);
\fill[tree5] (-0.633,0.028) circle (0.6pt);
\fill[tree5] (-1.956,-0.205) circle (0.6pt);
\fill[tree5] (-1.425,-0.013) circle (0.6pt);
\fill[tree5] (-0.829,0.641) circle (0.6pt);
\draw[black,line width=1.2pt,densely dashed] (-0.754,0.146) rectangle (0.985,1.307);
\end{scope}
\draw[red!75!black,line width=0.9pt] (0.000,1.300) ++(-0.08,0)--++(0.16,0) (0.000,1.300) ++(0,-0.08)--++(0,0.16);
\node[font=\scriptsize,anchor=south west,red!75!black,inner sep=1pt] at (0.030,1.330) {$x$};
\end{tikzpicture}
\caption{One query and six trees from a $25$-tree bundle, shown in coordinates $1$ and $2$; colors identify trees. Left: cut boxes (solid) and their enlargements by $x$ (dashed); the shaded extensions are empty. Circles mark the gap-box centres on each tree's largest-gap face. The arrow and ellipse give their mean displacement from $x$ and one-standard-deviation spread over all $25$ trees. Right: passage boxes of the same trees, at half the drawing scale. The black dashed box averages the query-relative half dimensions over all $25$ trees.}
\label{fig:passagebox}
\end{figure}

The escape probabilities of Section~\ref{sec:escape} summarize how a query
separates from one tree. The locations of those separations also select
neighborhoods around it. Across a forest, the query paths, with their
separation laws, masses, and boxes, form its \emph{hitting bundle}.
Figure~\ref{fig:escapes} shows one such path; Figure~\ref{fig:passagebox}
shows neighborhoods from several trees. The bundle retains both the
geometry of each neighborhood and the variation between trees.

\paragraph{Neighborhoods on one path.}
A query can traverse large boxes whose local separation probabilities
are small. We start the neighborhood selection near the end of this
part of the path. The cut box then uses a median attachment and the
passage box a low attachment quantile. Both are restricted to lie at
or below a common mass-selected ancestor; calibration selects
the comparison neighborhood by mass.

On the complete path $v_0,\ldots,v_h$, choose
$\varepsilon\in(0,1/2)$ and let $k$ be the deepest index with
$q_k\leq\varepsilon$, taking $k=0$ if there is none. Under the
width-proportional query law, the probabilities $q_i$ do not decrease
down the nested boxes: exterior gaps grow as widths shrink.
Conditional on reaching $v_k$, the cumulative attachment probability
and its quantile levels, for $0<\alpha<1$, are
\[
 F_k(i)=
 \begin{cases}
  1-\displaystyle\prod_{\ell=k}^i(1-q_\ell),&k\leq i<h,\\
  1,&i=h,
 \end{cases}
 \qquad
 Q_\alpha=\min\{i\in\{k,\ldots,h\}:F_k(i)>\alpha\}.
\]
The terminal event is assigned to $v_h$, so $Q_\alpha=h$ at an
exact match. We use the mass selector $c_\mu(x)$ of
Definition~\ref{def:massselector} to restrict the selected ancestors.

\begin{definition}[Cut, passage, and calibration boxes]
\label{def:queryboxes}
Using the path and restarted law above:
\begin{enumerate}\setlength{\itemsep}{3pt}
\item The \emph{cut box} is
$B_{\rm cut}(T,x)$, the data box at the deeper of the
conditional-median vertex $v_{Q_{1/2}}$ and $c_{\sqrt\psi}(x)$.
\item The \emph{passage box} is
$B_{\rm pass}(T,x)$, the data box at the deeper of
$v_{Q_\varepsilon}$ and $c_{\sqrt\psi}(x)$, using the low
attachment quantile with the same cap.
\item For a declared $\mu_{\rm cal}\in[\sqrt\psi,\psi]$, the
\emph{calibration box} is
$B_{\rm cal}(T,x;\mu_{\rm cal})=B_{c_{\mu_{\rm cal}}(x)}$.
The calibration mass can be selected by
\begin{equation}
 \mu(\theta)=\psi^{(1+\theta)/2},\qquad 0\leq\theta\leq1.
 \label{eq:calibrationmass}
\end{equation}
\end{enumerate}
Each box retains its mass and its enlargement by the query,
$M=\bbox(B\cup\{x\})$, together with the face gaps.
\end{definition}
The examples use $\varepsilon=0.05$ and
$\mu_{\rm cal}=\min\{\psi,2\sqrt\psi\}$; the latter corresponds
to $\theta=0.25$ when $\psi=256$. Restarting and the cap select
the reported boxes; the root-based escapes of
Section~\ref{sec:escape} retain their original law.

These are data boxes at three chosen vertices. Their query
enlargements preserve the same nesting:
$B_{\rm cut}\subseteq B_{\rm pass}\subseteq B_{\rm cal}$,
with ties allowed. The low quantile selects the passage box;
the conditional median can select a deeper cut box.
Their shared cap allows the two boxes to coincide. Calibration selects mass independently of the query's
hazard.

Figure~\ref{fig:escapes} illustrates the rules on one path.
For $\psi=8$, the cap $c_{\sqrt8}(x)$ is $v_3$.
The uncapped passage candidate is $v_1$ and the median is $v_3$,
so both reported boxes are $B_{v_3}$. The default calibration
mass exceeds the five records at $v_1$, so its box is the root
$B_{v_0}$. The table also shows the three escape readings
developed in Section~\ref{sec:escape}.

\begin{figure}[!htb]
\centering
\definecolor{lv1}{HTML}{D55E00}
\definecolor{lv2}{HTML}{0072B2}
\definecolor{lv3}{HTML}{009E73}
\definecolor{lv4}{HTML}{CC79A7}
\definecolor{lv5}{HTML}{E69F00}
\tikzset{
  pv/.style={circle,draw=black,line width=0.4pt,inner sep=0pt,minimum size=8.5pt,text=white,
             font=\fontsize{6}{6}\selectfont},
  sib/.style={rectangle,fill=black,inner sep=0pt,minimum size=3.2pt},
  elab/.style={font=\scriptsize,inner sep=1pt},
}
% (a) the data boxes on the path of x
\begin{tikzpicture}[x=0.8cm,y=0.8cm,line join=round,baseline=(basea)]
  \coordinate (basea) at (0,2.9);
  \node[font=\footnotesize,anchor=south] at (3.3,5.15) {(a) boxes on the path of $x$};
  \draw[black!40,line width=0.4pt] (1,1) rectangle (5.6,4.9);
  \node[elab,black!55,anchor=south east] at (5.6,1.0) {$v_0$};
  \fill[lv1,opacity=0.10] (1,1) rectangle (3,3.5);
  \draw[lv1,line width=0.7pt] (1,1) rectangle (3,3.5);
  \fill[lv2,opacity=0.10] (1,1) rectangle (2.5,3);
  \draw[lv2,line width=0.7pt] (1,1) rectangle (2.5,3);
  \fill[lv3,opacity=0.10] (1.5,1.5) rectangle (2.5,3);
  \draw[lv3,line width=0.7pt] (1.5,1.5) rectangle (2.5,3);
  \fill[lv4,opacity=0.10] (2,1.5) rectangle (2.5,2);
  \draw[lv4,line width=0.7pt] (2,1.5) rectangle (2.5,2);
  % the box of v_4 with x; shaded, where a cut of it separates x
  \fill[lv4,opacity=0.22] (2,2) -- (2.5,2) -- (2.5,1.5) -- (4,1.5) -- (4,2.5) -- (2,2.5) -- cycle;
  \draw[lv4,densely dashed,line width=0.5pt] (2,1.5) rectangle (4,2.5);
  \draw[lv4!75!black,<->,>=stealth,line width=0.45pt] (2.5,1.75) -- (4,1.75);
  \node[elab,lv4!75!black,anchor=north] at (3.25,1.72) {$g_1^+$};
  \draw[lv4!75!black,<->,>=stealth,line width=0.45pt] (3.2,2) -- (3.2,2.5);
  \node[elab,lv4!75!black,anchor=west] at (3.23,2.25) {$g_2^+$};
  % half dimensions of that box about x
  \draw[black,->,>=stealth,line width=0.5pt] (4,2.5) -- (2,2.5);
  \node[elab,anchor=south] at (3.0,2.52) {$a_1^-$};
  \draw[black,->,>=stealth,line width=0.5pt] (4,2.5) -- (4,1.5);
  \node[elab,anchor=west] at (4.03,2.0) {$a_2^-$};
  % vertex labels
  \node[elab,lv1,anchor=south west] at (1.0,3.5) {$v_1$};
  \node[elab,lv2,anchor=north west] at (1.03,2.97) {$v_2$};
  \node[elab,lv3,anchor=north east] at (2.47,2.97) {$v_3$};
  \node[elab,lv4,anchor=east] at (2.0,1.75) {$v_4$};
  % the sample
  \fill (1.0,1.0) circle (1.4pt); \node[elab,anchor=north east] at (1.0,1.0) {$a_1$};
  \fill (1.5,3.0) circle (1.4pt); \node[elab,anchor=south] at (1.5,3.02) {$a_2$};
  \fill (2.5,2.0) circle (1.4pt); \draw[lv5,line width=0.8pt] (2.5,2.0) circle (3.2pt);
  \node[elab,anchor=south west] at (2.56,2.04) {$a_3$};
  \fill (3.0,3.5) circle (1.4pt); \node[elab,anchor=south west] at (3.0,3.5) {$a_4$};
  \fill (2.0,1.5) circle (1.4pt); \node[elab,anchor=north] at (2.0,1.47) {$a_5$};
  \fill (4.6,4.4) circle (1.4pt);
  \fill (5.2,4.9) circle (1.4pt);
  \fill (5.6,4.2) circle (1.4pt);
  \node[elab,anchor=north] at (5.1,4.1) {$b_1,b_2,b_3$};
  % the query
  \draw[red!75!black,line width=0.9pt] (4,2.5) ++(-0.22,0) -- ++(0.44,0) (4,2.5) ++(0,-0.22) -- ++(0,0.44);
  \node[elab,red!75!black,anchor=south west] at (4.06,2.56) {$x$};
\end{tikzpicture}%
\hspace{18pt}%
% (b) the tree, and where separation at each v_i would attach x
\begin{tikzpicture}[x=0.55cm,y=0.55cm,line join=round,baseline=(baseb)]
  \coordinate (baseb) at (0,-2.4);
  \node[font=\footnotesize,anchor=south] at (2.4,0.75) {(b) the tree, and where $x$ attaches};
  \coordinate (v0) at (0,0); \coordinate (v1) at (1,-1); \coordinate (v2) at (2,-2);
  \coordinate (v3) at (3,-3); \coordinate (v4) at (4,-4); \coordinate (v5) at (5,-5);
  \coordinate (sb) at (-1,-1); \coordinate (s4) at (0,-2); \coordinate (s1) at (1,-3);
  \coordinate (s2) at (2,-4); \coordinate (s5) at (3,-5);
  \coordinate (qx) at (5.4,-0.9);
  \draw[lv1,densely dashed,line width=1.21pt] (qx) -- ($(v0)!0.5!(v1)$);
  \draw[lv2,densely dashed,line width=1.53pt] (qx) -- ($(v1)!0.5!(v2)$);
  \draw[lv3,densely dashed,line width=1.37pt] (qx) -- ($(v2)!0.5!(v3)$);
  \draw[lv4,densely dashed,line width=1.49pt] (qx) -- ($(v3)!0.5!(v4)$);
  \draw[lv5,densely dashed,line width=0.90pt] (qx) -- ($(v4)!0.5!(v5)$);
  \draw[line width=0.6pt] (v0) -- (v1) -- (v2) -- (v3) -- (v4) -- (v5);
  \draw[line width=0.6pt] (v0) -- (sb) (v1) -- (s4) (v2) -- (s1) (v3) -- (s2) (v4) -- (s5);
  \draw[fill=black!15,line width=0.4pt] (sb) -- ++(-0.45,-0.8) -- ++(0.9,0) -- cycle;
  \node[elab,anchor=east] at (-1.5,-1.45) {$b_1,b_2,b_3$};
  \foreach \p/\n in {s4/a_4,s1/a_1,s2/a_2,s5/a_5} {
    \node[sib] at (\p) {};
    \node[elab,anchor=east] at ($(\p)+(-0.12,0)$) {$\n$};
  }
  \node[pv,fill=black!45] at (v0) {0};
  \node[pv,fill=lv1] at (v1) {1};
  \node[pv,fill=lv2] at (v2) {2};
  \node[pv,fill=lv3] at (v3) {3};
  \node[pv,fill=lv4] at (v4) {4};
  \node[pv,fill=lv5] at (v5) {5};
  \node[elab,anchor=north] at ($(v5)+(0,-0.2)$) {$a_3$};
  \draw[red!75!black,line width=0.9pt] (qx) ++(-0.2,0) -- ++(0.4,0) (qx) ++(0,-0.2) -- ++(0,0.4);
  \node[elab,red!75!black,anchor=south west] at ($(qx)+(0.06,0.06)$) {$x$};
\end{tikzpicture}

\medskip
{\footnotesize (c) the separation law and the payoffs of separation\par\smallskip
\begin{tabular}{@{}lrrlrrrccc@{}}
\toprule
 & & & & & & & \multicolumn{3}{c}{payoff of separating at $v_i$}\\
\cmidrule(l){8-10}
 & $m_{v_i}$ & $R(B_{v_i})$ & gaps of $x$ & $q_i$ & $s_i$ & $s_iq_i$ & $1/(i+1)$ & $R(B_{v_i}\cup\{x\})/R_0$ & $m_{v_i}/\psi$\\
\midrule
$v_0$ & $8$ & $8.5$ & none & $0$ & $1$ & $0$ & & & \\
$v_1$ & $5$ & $4.5$ & $g_1^+=1$ & $0.182$ & $1$ & $0.182$ & $1/2$ & $0.647$ & $0.625$\\
$v_2$ & $4$ & $3.5$ & $g_1^+=1.5$ & $0.300$ & $0.818$ & $0.245$ & $1/3$ & $0.588$ & $0.500$\\
$v_3$ & $3$ & $2.5$ & $g_1^+=1.5$ & $0.375$ & $0.573$ & $0.215$ & $1/4$ & $0.471$ & $0.375$\\
$v_4$ & $2$ & $1$ & $g_1^+=1.5$, $g_2^+=0.5$ & $0.667$ & $0.358$ & $0.239$ & $1/5$ & $0.353$ & $0.250$\\
$v_5$ & $1$ & $0$ & $g_1^+=1.5$, $g_2^+=0.5$ & $1$ & $0.119$ & $0.119$ & $1/6$ & $0.235$ & $0.125$\\
\midrule
\multicolumn{6}{@{}l}{expectation under the separation law} & $1$ & $0.294$ & $0.475$ & $0.391$\\
\bottomrule
\end{tabular}}
\caption{One query path through eight observations. (a)~Nested data boxes and $x=(4,2.5)$; the dashed box enlarges $B_{v_4}$ by $x$, and its shaded extensions are empty. (b)~The same ancestry, with attachment widths proportional to $s_iq_i$. (c)~The separation law and the three escape readings of Section~\ref{sec:escape}, with $R_0=8.5$. Under Definition~\ref{def:queryboxes}, the cut and passage boxes both equal $B_{v_3}$, and the calibration box is $B_{v_0}$. Colors identify ancestors.}
\label{fig:escapes}
\end{figure}

\paragraph{Geometry across trees.}
The trees can select different neighborhoods for the same query,
both because they retain different records and because their
cuts differ conditional on those records.

\begin{definition}[Directional geometry of the hitting bundle]
\label{def:bundlegeometry}
Fix coordinate units and a box-selection rule. For each realized
tree $T_t$, let $\boldsymbol a_t(x)\in\R_+^{2d}$ contain the
query-relative half dimensions $a_{tj}^-,a_{tj}^+$ of its selected
box, as in Definition~\ref{def:halfdimensions}. The vector retains
elongation across coordinates and imbalance between opposite
faces. Differences among the widths $a_{tj}^-+a_{tj}^+$ record
the selected box's anisotropy in these units; differences between
$a_{tj}^-$ and $a_{tj}^+$ record its asymmetry about the query.
Its representative box has half dimensions
\[
 \bar a_j^\pm=\frac1N\sum_{t=1}^N a_{tj}^\pm.
\]
Equal coordinate widths do not by themselves establish rotational
isotropy of the underlying data.
\end{definition}

\begin{definition}[Mean gap displacement and its covariance: lead and swing]
\label{def:leadswing}
At each selected cut box choose a positive-gap face by a declared
rule, and let $c_T(x)$ be the centre of the gap box spanned by
that face and $x$. Set $c_T(x)=x$ if there is no positive gap.
For $Z_T(x)=c_T(x)-x$, define the \emph{lead} and \emph{swing} by
\[
 \boldsymbol\ell_{\rm gap}(x)=\E_T Z_T(x),\qquad
 \Sigma_{\rm gap}(x)=
 \E_T\!\left[(Z_T(x)-\boldsymbol\ell_{\rm gap}(x))
            (Z_T(x)-\boldsymbol\ell_{\rm gap}(x))^\top\right],
\]
when the second moment exists. In a finite forest, use its
empirical averages. The figure chooses a largest-gap face.
\end{definition}

Lead is mean gap displacement; swing is its covariance across
trees. A long lead with small swing means the sampled trees
locate similar empty margins. Both sampling and cut variation
contribute to swing. For a displayed coordinate projection $P$,
the mean arrow is $P\boldsymbol\ell_{\rm gap}$ and the covariance
ellipse uses $P\Sigma_{\rm gap}P^\top$. These moments describe
selected gap locations; the half dimensions retain the boxes'
extents as well.

Figure~\ref{fig:passagebox} shows a three-dimensional \emph{blade}:
coordinate $1$ is uniform on $[-5,5]$ and the transverse standard
deviations are $0.5$ and $0.12$. The query $x=(0,1.3,0)$ lies beside
its narrow side; each of $25$ trees holds $\psi=256$ observations.
The mean passage box reaches $1.15$ toward the data and only $0.01$
away in coordinate $2$. The swing is mainly along the blade, with
standard deviation $0.55$ against $0.10$ across it: the trees agree
on the separation side while their gap locations vary along the blade.

\paragraph{Directional readings.}
The same separating-face record also explains a scalar escape.
For inverse depth, keeping that face gives its contributions:
\begin{equation}
 A_{j,\pm}(T,x)=\sum_i\frac{s_i p_{i,j,\pm}}{i+1},
 \qquad \sum_{j,\pm}A_{j,\pm}(T,x)=U(T,x).
 \label{eq:escapeattribution}
\end{equation}
The fold \eqref{eq:recurrence} computes the probability and
these contributions together. The library's attribution visitor
carries out this calculation before normalization.

\paragraph{Anisotropic balloon estimates.}
\label{sec:massextent}
Mass per unit extent gives a local density reading of the selected
neighborhoods. Selecting a neighborhood at the query is the \emph{balloon}
view of density estimation \citep{terrell1992variable}.
The mass criterion plays the role of the neighbor count
\citep{loftsgaarden1965nonparametric}; the tree supplies an
anisotropic window. For a sample $S$ of total mass $n$, a selected
window $A_x$, and a reference measure $\lambda$ with $\lambda(A_x)>0$, its empirical
occupancy per unit extent is
\begin{equation}
 \widehat f_{\rm occ}(x)=\frac{m(S\cap A_x)/n}{\lambda(A_x)}.
 \label{eq:occupancy}
\end{equation}
For example, $\lambda$ may be ambient volume or a measure on an
identified support. For a fixed-shape window of volume $Ch^d$,
locally constant density gives $k/n\simeq f(x)Ch^d$:
choosing mass chooses a bandwidth, while rectangular windows
retain different lengths in different directions.

\begin{samepage}
A scalar summary of the bundle also requires an aggregation rule.
For the separation-law mass reading, fix a probe mass $\eta\geq0$,
put $\xi_v=m_v+\eta$, and define
\begin{equation}
 a(x)=\frac{\sum_{t=1}^N\int\xi_v\,d\mu_{T_t,x}}
                  {\sum_{t=1}^N n_t},
 \label{eq:passagemass}
\end{equation}
where $n_t$ is tree $t$'s sample mass and the terminal event receives
its declared mass payoff. For a selected cut, passage, or calibration
box, use its mass payoff in place of the integral. This specifies
which mass reading accompanies the geometric summary.\par
\end{samepage}

\paragraph{Regularization of the bundle estimate.}
\label{app:densityestimates}
The representative geometry and the chosen mass reading specify
the raw estimate. A regularization specifies how a vanishing
extent is treated.

\begin{definition}[Regularization of mass per unit extent]
\label{def:bundleregularization}
For the representative half dimensions of
Definition~\ref{def:bundlegeometry}, put
\[
 V_{\rm box}(x)=\prod_{j=1}^d(\bar a_j^-+\bar a_j^+).
\]
Given $a(x)>0$, the additive regularization $\mathcal R_\gamma$
adds $\gamma\geq0$ to volume per unit normalized mass:
\begin{equation}
 \widehat f^{\mathcal R_\gamma}(x)
 =\frac{1}{V_{\rm box}(x)/a(x)+\gamma}
 =\frac{a(x)}{V_{\rm box}(x)+\gamma a(x)},
 \qquad V_{\rm box}(x)+\gamma a(x)>0.
 \label{eq:boxdensity}
\end{equation}
Since $a(x)$ is dimensionless, $\gamma$ has volume units.
\end{definition}
Thus $\gamma$ sets a floor on volume per unit mass. Taking
$\gamma=0$ gives the raw ratio where the volume is positive;
taking $\gamma>0$ caps it at $1/\gamma$, including when the
observed volume vanishes. This operation regularizes a geometric
reading; the measurement-error variance of
Section~\ref{sec:green} belongs to a different model.
The order of aggregation is part of this estimate: average the
half dimensions, form a volume, and divide the chosen mass reading
by its regularized value. Averaging per-tree ratios is a different
choice. To report empirical occupancy of the representative box
itself, count its contents and use \eqref{eq:occupancy}.

On lower-dimensional data, a density on the support requires a
measure on that support; an ambient-volume comparison can instead
use the declared regularization above. Dropping zero-width
coordinates amounts to choosing a coordinate support. These
finite neighborhood readings retain their mass, extents, and gaps
as geometric evidence. As with other balloon estimates, global
normalization is a separate requirement
\citep{terrell1992variable,hoffmann2015unified}.

\paragraph{Directional mass-to-extent estimates.}

The balloon reading can also use each separating face and its
associated length.
Normalize the nonterminal directional law, retaining its terminal
probability separately. For $\sigma\in\{-,+\}$, let $w_{j\sigma}$
be the probability of separation on face $(j,\sigma)$ and
$D_{j\sigma}$ the integral of the expanded width $a_j^-+a_j^+$
on those events. Put $L_{j\sigma}=D_{j\sigma}/w_{j\sigma}$
when $w_{j\sigma}>0$. The face identifies the side of separation;
the width supplies its length scale. For a declared moment order
$p>0$, independent of the cut and field-energy exponents,
the same additive regularization gives
\begin{equation}
 V_{\rm dir}^{(p)}(x)
 =\sum_{j,\sigma:w_{j\sigma}>0}w_{j\sigma}L_{j\sigma}^p,
 \qquad
 \widehat f_{\rm dir}^{\mathcal R_\gamma,p}(x)
 =\frac{a(x)}{V_{\rm dir}^{(p)}(x)+\gamma a(x)}.
 \label{eq:dirdensity}
\end{equation}
Pooling opposite faces first gives $w_j=w_{j-}+w_{j+}$ and
$D_j=D_{j-}+D_{j+}$. Applying the power to $L_j=D_j/w_j$
generally changes the estimate when $p\ne1$. Keeping the faces
therefore matters even for a scalar summary. In common coordinate
units, $V_{\rm dir}^{(p)}$ and its regularization scale $\gamma$
have units of length to the power $p$. Taking $p=d$ gives volume
units to this mixture of directional lengths; its reference
measure and exponent remain declared choices.

\section{Laplacians and Hamiltonians on a Tree}
\label{sec:exact}
Section~\ref{sec:existing} asks whether a walk escapes before it returns.
To predict a value, supply values at reference vertices and let the
walk return the first prescribed value it reaches. An escape probability is the case in which one
boundary has value one and the other zero. This hitting functional
therefore gives both a probability and a weighted prediction. The
chosen energy below specifies the walk's transition weights; the
prescribed boundary values determine which question is being asked.

To couple values across the observations, charge disagreements
according to how strongly the geometry connects them. The graph
Laplacian records these couplings; its quadratic form is the
Hamiltonian. Minimizing that energy over unknown values gives
the prediction with the least disagreement allowed by the evidence.
This passage from sampled geometry to diffusion energy has a long
history. Laplacian eigenmaps and diffusion maps construct weighted
interactions from observations and use their Laplacians to describe
geometry and diffusion
\citep{belkin2003laplacian,coifman2006diffusion}.
The sampling measure and the normalization affect which manifold
operator a graph approximates \citep{hein2007graph}.
Here the declared sample and the maintained tree determine the
finite geometry; the affine bounds connect its interaction energy
directly to distances between the observed locations.

Shared ancestry explains why a tree can also describe uncertainty.
Imagine an uncertain value at the root, followed by an independent,
mean-zero change on each edge. Two observations inherit the same changes
until their paths part, then acquire separate changes. Their covariance
is precisely the variance accumulated on that shared path. A measurement
on one branch gives evidence about this shared part of a second branch's
value. Changes on an unobserved branch remain uncertain.

We first condition on the retained sample. Theorem~\ref{thm:geometryenergy}
averages the cut law on this fixed sample; each network construction
is then computed on one realized tree. Sampling different records
changes the geometry as well as the answer, according to the law of
Section~\ref{sec:sampling}. Section~\ref{sec:extension} puts trees
with different retained samples on common reference observations
before combining their answers. The Gaussian uncertainty below is
conditional uncertainty in values on the chosen geometry.

\paragraph{Additive geometry and interaction.}
\label{sec:rangeaffinity}
An \emph{affinity} is a weight coupling two observation values.
The tree supplies their unique path and its additive length.
To make the affinity independent of path subdivision, require
survival through consecutive pieces to multiply. Its length-based
factor then satisfies
\[
 S(0)=1,\qquad 0<S(\ell)\leq1,\qquad S(a+b)=S(a)S(b).
\]
For a positive, continuous, nonconstant survival law, taking
$-\log S$ gives a continuous additive function. Hence, for
some length scale $h>0$,
\[
 S(\ell)=e^{-\ell/h}.
\]
The requirement concerns composition on a fixed geometry.
It specifies how the geometric contributions at different levels
combine before boundary values or labels are supplied.

On the range tree of \eqref{eq:rcfrangemetric}, with edge lengths
$\ell_e$, the pair affinity is
\[
 K_{ij}=\prod_{e\text{ on the path }i\leftrightarrow j}
                    e^{-\ell_e/h}
        =e^{-d_T(i,j)/h}.
\]
The scale $h$ is fixed for one field: an affinity falls by a factor
$e$ over distance $h$. A declared calibration box
(Definition~\ref{def:queryboxes}) can supply that length;
Section~\ref{sec:routed} sets a common scale across the forest.
Changing $h$ changes the coupling, while rescaling both $h$ and
the geometric lengths only changes units.

\paragraph{Range escape and affinity.}
Sample $U$ under the attachment law of Section~\ref{sec:escape},
put $C_x=R(M_{\rm root}(x))>0$, and let
$Z=R(B_U)/C_x$. The affinity at spatial scale $h$ is $e^{-\lambda Z}$,
where $\lambda=C_x/h$ is the dimensionless inverse scale. Hence
\begin{equation}
 F_x(\lambda)=\E[1-e^{-\lambda Z}]
             =\lambda\,\E Z+O(\lambda^2).
 \label{eq:escapeslope}
\end{equation}
The initial slope is range escape \eqref{eq:rangeescape}; the finite
quantity records affinity lost at the chosen spatial scale. For a
declared-mass first-passage trial, use the normalized range $Z_\mu$
of its first augmented ancestor with at least $\mu$ original records:
$F_{x,\mu}'(0)=\operatorname{Esc}_\mu(T,x)$.
For an unseen query, Corollary~\ref{cor:affineescapes} controls
the expected initial slope through its nearest geometric gap and
the affine dimension of the augmented sample.

Section~\ref{sec:game} computes predictions and responses of a
network energy; Section~\ref{sec:starkernel} realizes the complete
affinity energy on a sparse tree network.

\subsection{Shared Ancestry, Prediction, and Equilibrium}
\label{sec:game}\label{sec:green}
A \emph{reference set} $P$ is the collection of observations against
which an answer is read. For now these observations have vertices on
one tree. Some may supply measured values; in a grouping problem their
values may instead be the class indicators being tested. They are not
automatically all fixed. A \emph{clamp} prescribes a value, such as a
measurement, a proposed class, or a zero prior mean. A \emph{free vertex}
is one whose value is still to be determined. It may be an unmeasured
observation, a query, or an internal vertex carrying the shared part of
its descendants' values. These are values on the retained geometry;
solving for them does not move the data or rebuild the cuts.

An edge's conductance $c_{uv}$ specifies how strongly its two values
are coupled. Its reciprocal is resistance: a large resistance permits
a larger difference at lower energy cost. Geometric lengths determine
these couplings through the constructions below.

\begin{definition}[Hamiltonian $\mathcal H$ and prescribed values]
\label{def:networkenergy}\label{def:affinityenergy}
For an undirected network with conductances $c_{uv}>0$ and Laplacian
$L$, the Hamiltonian $\mathcal H$ of the field $f$ is its total
weighted disagreement energy:
\begin{equation}
 \mathcal H(f)=\tfrac12 f^\top Lf
              =\tfrac12\sum_{\{u,v\}}c_{uv}(f_u-f_v)^2,
 \label{eq:hamiltonian}
\end{equation}
counting each edge once. Prescribe $f_A=y_A$ at a set of vertices $A$,
and write $F$ for the remaining vertices. Each connected component
must reach a prescribed value to determine its absolute level; a
\emph{ground} supplies a fixed zero value when a prior mean is required.
The equilibrium minimizes $\mathcal H$ over $f_F$.
For pair affinities $K$ on observations $P$, coupling every distinct
pair directly with conductance $K_{ij}$ gives the agreement energy
\[
 \mathcal H_{\rm aff}(y)=\tfrac12 y^\top L_Ky
   =\tfrac12\sum_{i<j}K_{ij}(y_i-y_j)^2,\qquad
 L_K=\operatorname{diag}(K\one)-K.
\]
Self-affinities cancel from $L_K$.
\end{definition}

\begin{theorem}[Affine geometry controls the Hamiltonian]
\label{thm:geometryenergy}
Fix a retained sample $S$ with at least two distinct locations,
write $\delta_{ij}=\|x_i-x_j\|_1$,
$\Delta=\operatorname{diam}_1(S)$, and let
$\delta_*=\min_{\delta_{ij}>0}\delta_{ij}$.
For $s>0$ define the direct distance energy
\[
 \mathcal H^{(1)}_s(f)
   =\tfrac12\sum_{i<j}e^{-\delta_{ij}/s}(f_i-f_j)^2,
\]
and let $\mathcal H^T_h$ use the tree affinities
$e^{-d_T(x_i,x_j)/h}$ at a fixed $h>0$.
If $S$ lies in an $r$-dimensional affine subspace, put
\[
 a=2\mathbb K_{r,1}(1/2)
       \left[1+\left\lceil\log_2(\Delta/\delta_*)\right\rceil\right].
\]
For every field $f$, averaging width-proportional trees conditional
on $S$ gives
\begin{equation}
 \mathcal H^{(1)}_{h/a}(f)
       \leq\E_T\mathcal H^T_h(f)
       \leq\mathcal H^{(1)}_h(f).
 \label{eq:affineenergy}
\end{equation}
Equivalently, the corresponding Laplacians are ordered as
quadratic forms. This is a comparison of the averaged energy
before minimization.

In the noise model of Corollary~\ref{cor:noisygeometry}, let $r$
be the clean affine dimension and $W>0$. Using the observed
distances above, set
\[
 \begin{gathered}
 t_W=\max\{\delta_*,16W\},\qquad
 b_W=20rW\,\one\{\delta_*<16W\},\\
 a_W=2\mathbb K_r^{\rm noise}(1/2)
   \left[1+\left\lceil
      \log_2\frac{\max\{\Delta,t_W\}}{t_W}
    \right\rceil\right].
 \end{gathered}
\]
Then
\begin{equation}
 e^{-b_W/h}\mathcal H^{(1)}_{h/a_W}(f)
       \leq\E_T\mathcal H^T_h(f)
       \leq\mathcal H^{(1)}_h(f).
 \label{eq:noisyaffineenergy}
\end{equation}
In particular, when all distinct locations are separated by at
least $16W$, the additional attenuation is absent.
\end{theorem}
\begin{proof}
For $\delta_{ij}>0$, Theorem~\ref{prop:affinegeometry} gives
$d_T(x_i,x_j)\geq\delta_{ij}$ and
$\E_Td_T(x_i,x_j)\leq a\delta_{ij}$. Since $e^{-z/h}$ is convex,
\[
 e^{-a\delta_{ij}/h}
 \leq e^{-\E_Td_T(x_i,x_j)/h}
 \leq\E_T e^{-d_T(x_i,x_j)/h}
 \leq e^{-\delta_{ij}/h}.
\]
Coincident records have distance zero and affinity one in both
constructions. Multiply by $(f_i-f_j)^2/2$ and sum.

For the noisy case, follow a distinct pair's common cell until it
separates or the cell's data diameter falls below $t_W$.
Corollary~\ref{cor:noisygeometry}, applied conditionally in blocks,
bounds the expected number of these cuts by $a_W$.
At each such cut the separation probability is
$\delta_{ij}/R(B)$, so its expected separating-range payoff is
$\delta_{ij}$. Summing the stopped payoffs bounds the contribution
before this stopping point by $a_W\delta_{ij}$.
If $t_W=\delta_*\geq16W$, no distinct pair remains together in
a cell of diameter less than $t_W$.
Otherwise $t_W=16W$. If $A$ is the set of records in the stopped
cell, every subsequent separating range is at most
\[
 R(\bbox A)\leq r\operatorname{diam}_1(A)+2(r+1)W
       \leq(18r+2)W\leq20rW,
\]
by \eqref{eq:noisyaffinerange}. Thus
$\E_Td_T(x_i,x_j)\leq a_W\delta_{ij}+b_W$.
Jensen's inequality and summation prove
\eqref{eq:noisyaffineenergy}.
\end{proof}

The affine theorem therefore bounds the strength of the averaged
coupling by changing the spatial scale of a direct distance energy.
Below the noise floor, unresolved geometry adds the attenuation
$e^{-b_W/h}$. The comparison uses the same stored observations in
every tree; routing a different reference set is the further
construction of Section~\ref{sec:extension}.

\paragraph{Information geometry of field responses.}
Fix the retained observations, bandwidth $h$, and a nonempty prescribed
boundary $A$, with nonempty free set $F$. Pool the observation
Laplacians as $\bar L_h=\E_T L_h^T$ and put
$Q_h=(\bar L_h)_{FF}$, which is positive definite. At temperature
$\vartheta>0$, consider the Gaussian mean family
$P_m=\mathcal N(m,\vartheta Q_h^{-1})$ on the free values. For a
mean shift $\xi\in\R^F$, extend $\xi$ by zero on $A$ to obtain
$\tilde\xi$. The Fisher metric of this family is
\begin{equation}
 g_m(\xi,\xi)
   =\frac{1}{\vartheta}\xi^\top Q_h\xi
   =\frac{2}{\vartheta}\E_T\mathcal H_h^T(\tilde\xi),
 \label{eq:fieldfisher}
\end{equation}
and, since the covariance is fixed,
$\operatorname{KL}(P_m\|P_{m+\xi})=\tfrac12g_m(\xi,\xi)$
\citep{amari2000methods,amari2016information}.
The field energy therefore measures the statistical distinguishability
of mean shifts. Under the affine-support assumptions of
Theorem~\ref{thm:geometryenergy},
\begin{equation}
 \frac{2}{\vartheta}\mathcal H^{(1)}_{h/a}(\tilde\xi)
 \;\leq\;g_m(\xi,\xi)\;\leq\;
 \frac{2}{\vartheta}\mathcal H^{(1)}_h(\tilde\xi).
 \label{eq:affinefisher}
\end{equation}
The noisy version replaces $a$ by $a_W$ and multiplies the lower
bound by $e^{-b_W/h}$, as in \eqref{eq:noisyaffineenergy}.
This construction forms the Gaussian law after pooling energies.
\emph{Affine structure controls tree geometry; tree geometry controls
energy; energy controls distinguishability of field responses.}

The equilibrium can be found by local agreement: each free value
equals the conductance-weighted average of its neighbors. It is also
a potential game. A player controls one free value and pays for its
disagreement with its neighbors; a unilateral change has exactly the
same cost in the player's objective and in the total energy.

\begin{proposition}[A potential game]
\label{prop:potentialgame}
Let each free vertex $u$ be a player that chooses $f_u\in\mathbb R$ at the cost $J_u(f)=\frac12\sum_vc_{uv}(f_u-f_v)^2$, with $f_A=y_A$.
\begin{enumerate}\setlength{\itemsep}{1pt}
\item[(i)] A change of $f_u$ alone changes $J_u$ and $\mathcal H$ by the same amount, so the game is an exact potential game with potential $\mathcal H$ \citep{monderer1996potential}. Each best response is the conductance-weighted average of the neighbors, and the unique Nash equilibrium \citep{nash1951noncooperative} is the harmonic extension $f^\star_F=-L_{FF}^{-1}L_{FA}y_A$, the minimizer of $\mathcal H$. Equivalently, a walk with transition probabilities $c_{uv}/\sum_wc_{uw}$ returns the prescribed value at its first hit of $A$, and $f^\star_u=\E_u[y_{X_{\tau_A}}]$ for $u\in F$.
\item[(ii)] On the tree of Proposition~\ref{prop:heights}, with the query at zero and the root at one, the equilibrium value at the vertex above the query is the escape \eqref{eq:heights}, and at every vertex of the path it is the probability that a walk from there reaches the root before the query.
\end{enumerate}
\end{proposition}
\begin{proof}
(i) $J_u$ and $\mathcal H$ share the terms on the edges at $u$, and no other term of either depends on $f_u$. Setting $\partial J_u/\partial f_u=\sum_vc_{uv}(f_u-f_v)$ to zero gives the weighted average. Since every component of $F$ reaches $A$, $L_{FF}$ is positive definite and $\mathcal H$ is strictly convex in $f_F$. A profile at which no player can lower its own cost has $\partial\mathcal H/\partial f_u=\partial J_u/\partial f_u=0$ for every $u$, so it is the unique minimizer, and the minimizer is such a profile. The first-step equation for the expected boundary value is the same weighted-average equation, with the same prescribed values on $A$, so uniqueness gives the hitting representation. (ii) Branches off the path end in free vertices and carry no current, so on the path the equilibrium is linear in the cumulative resistance $L$: it is $L(w)/L({\rm root})$ at $w$, the probability that the walk from $w$ meets the root first, and at the vertex above the query it is \eqref{eq:heights}.
\end{proof}

The matrix $G=(L_{FF})^{-1}$ records how these predictions respond
to additional evidence. Its column at $q$ is the change caused by
a unit current applied at $q$, with the prescribed values held fixed.
This response matrix is the \emph{grounded Green's function}.
It too is a hitting functional. Run a continuous-time walk that jumps
from $u$ to $v$ at rate $c_{uv}$ and stops on reaching $A$. Then
\begin{equation}
 G(u,v)=\E_u\!\left[\int_0^{\tau_A}
                       \one\{X_t=v\}\,dt\right].
 \label{eq:greenoccupation}
\end{equation}
Its entries count time spent at a specified vertex before the fixed
evidence is reached, rather than the whole duration of the walk.
The stopped transition matrix is $e^{-tL_{FF}}$, whose time integral
is $L_{FF}^{-1}$. The jump probabilities at departures are those
of Proposition~\ref{prop:potentialgame}; the mean holding time at $u$ is
$1/\sum_vc_{uv}$. In particular, the diagonal records time at
the query, the same response that will give its model variance.
The following construction identifies it directly from shared ancestry.

\paragraph{A grounded network for shared variation.}
Let $T=(W,\mathcal E)$ have positive geometric edge lengths $\ell_e$,
after zero-length edges are contracted. Put
$\delta_u=d_T({\rm root},u)$ and $\psi(u)=e^{2\delta_u/h}$ for a
declared spatial scale $h>0$. Give the root an independent Gaussian
value of variance one; along $p\to c$ add an independent Gaussian
increment of variance $\psi(c)-\psi(p)$. The corresponding grounded
network $N$ has root-to-ground resistance one and edge resistance
$\psi(c)-\psi(p)$. Thus each resistance is the variance of the
change carried by that edge. Define
\begin{equation}
 R_{uv}=e^{-d_T(u,v)/h},\qquad
 \rho_e=e^{-\ell_e/h},\qquad
 K_{ij}=R_{ij}\ (i,j\in P),\qquad
 b_u=\sum_{j\in P}R_{uj}.
 \label{eq:treelaplace}
\end{equation}
Rescaling each Gaussian value to variance one will give $R$.
On the RCF range geometry already defined in
\eqref{eq:rcfrangemetric}, the same construction can use cumulative
variance $\varphi(v)=e^{-R(B_v)/h}$. Nested ranges make its increments
nonnegative, and each observation has variance one. The identity
$d_T(i,j)=R(B_{\lca(i,j)})$ then gives the observation covariance
without a rescaling.

\begin{proposition}[The grounded Green's function]
\label{prop:green}
Let $L_N$ be the Laplacian with the ground held at zero, and $G^{\rm prior}=L_N^{-1}$. Then $G^{\rm prior}(u,v)=\psi(\lca(u,v))$ and $R=E G^{\rm prior} E$ for $E=\operatorname{diag}(e^{-\delta_u/h})$. On the range metric of \eqref{eq:rcfrangemetric}, use $\varphi(v)=e^{-R(B_v)/h}$ in place of $\psi$, and root resistance $\varphi({\rm root})$. Then $E=\operatorname{diag}(e^{H_v/h})$, and $G^{\rm prior}$ itself restricts to $K$ on the observations:
\begin{equation}
 K_{ij}=\varphi(\lca(i,j)),\qquad \varphi(v)=e^{-R(B_v)/h}.
 \label{eq:ultrakernel}
\end{equation}
\end{proposition}
\begin{proof}
A unit current from $u$ to the ground flows along its path to the root and through the ground edge. The potential at $v$ is the resistance of the part of that path shared with the path from $v$ to ground. That resistance telescopes to $\psi(\lca(u,v))$. The identity for $R$ follows from $d_T(u,v)=\delta_u+\delta_v-2\delta_{\lca(u,v)}$. On the range tree the same sum telescopes to $\varphi(\lca(u,v))$, and $d_T(u,v)=2H_{\lca(u,v)}-H_u-H_v$ gives the indicated rescaling. Since $H_i=0$ for each observation, that rescaling is one on $P$.
\end{proof}

In the Gaussian reading, the covariance is the sum of increment
variances shared by the two paths. In the electrical reading, a unit
current to ground follows the unique path, and its potential on another
branch is the shared resistance. These are the same sum. Upward fusion
of the evidence and downward prediction therefore compute conditioning
on a tree in linear time \citep{chou1994multiscale}.
For the range construction,
$h[\varphi(c)-\varphi(p)]\to R(B_p)-R(B_c)$ as $h\to\infty$.
This recovers the resistance increments used for the escapes in
Section~\ref{sec:escape}; prescribing the query and root values gives
that earlier hitting experiment.

\paragraph{Prediction, uncertainty, and testing a value.}
Now suppose the references $P$ carry measured values $y_P$. On the range tree, conditioning this Gaussian model gives simple Kriging \citep{stein1999interpolation,rasmussen2006gaussian}: the measurements are held fixed, the remaining values are predicted, and the prior ground stays at zero. Equivalently, a walk returns the value of the measurement or ground that it hits first. Independent measurement error of variance $\epsilon$, the Kriging nugget, is represented by a resistance $\epsilon$ between each latent reference value and its prescribed measurement. At zero nugget they coincide. The prediction is $G^{\rm prior}(q,P)(K+\epsilon I)^{-1}y_P$, and the conditioned Green's function is
\begin{equation}
 C^{\rm prior}_{P,\epsilon}(u,v)=G^{\rm prior}(u,v)-G^{\rm prior}(u,P)(K+\epsilon I)^{-1}G^{\rm prior}(P,v).
 \label{eq:conditionedgreen}
\end{equation}
At $\epsilon=0$ one can use the pseudoinverse for redundant coincident measurements. The diagonal $C^{\rm prior}_{P,\epsilon}(q,q)$ is the remaining variance of the latent value at $q$ under this model. Electrically it is the effective resistance to the prescribed measurements and ground: a weakly supported value can change substantially in response to a small applied current. If it is positive, clamping $q$ a distance $\delta$ from its equilibrium and letting the other free values respond gives
\begin{equation}
 \Delta f(v)=\delta\,\frac{C^{\rm prior}_{P,\epsilon}(v,q)}{C^{\rm prior}_{P,\epsilon}(q,q)},\qquad
 \Delta\mathcal H_{\min}=\frac{\delta^2}{2C^{\rm prior}_{P,\epsilon}(q,q)}.
 \label{eq:greenclamp}
\end{equation}
Indeed a unit current at $q$ gives the column $C^{\rm prior}_{P,\epsilon}(\cdot,q)$; scaling that current imposes the desired change, and its energy is half the current times the change. Across an edge of conductance $c$, if the far branch has effective conductance $a$ to its fixed evidence, the change is multiplied by $c/(c+a)$: the edge current must equal the current into the branch. A weak connection attenuates a change when the branch has its own evidence. A branch with no fixed evidence follows the change at equilibrium even across a weak edge. The response thus records both the geometry of the gap and the evidence beyond it.

\paragraph{Field fluctuations and Langevin dynamics.}
Allow the inferred values to fluctuate about the prediction while
holding the sampled geometry and prescribed values fixed.
In overdamped Langevin dynamics \citep[Chapters~3--4]{pavliotis2014stochastic},
the free values descend the energy while receiving independent noise:
\begin{equation}
 df_F=-\bigl(L_{FF}f_F+L_{FA}y_A\bigr)\,dt
                       +\sqrt{2\vartheta}\,dW_t.
 \label{eq:langevin}
\end{equation}
Here $\vartheta>0$ sets the fluctuation level of the field values
and $W$ is standard Brownian motion on $\R^F$. Its stationary law is
$\mathcal N(f_F^\star,\vartheta L_{FF}^{-1})$: the hitting prediction
is its mean, and the response matrix is its covariance at unit noise
level. Fixing all other values gives conditional variance
$\vartheta/d_u$ at $u$, where $d_u=\sum_vc_{uv}$; letting the other
free values fluctuate gives marginal variance $\vartheta G(u,u)$.
The deterministic mean relaxes as $e^{-tL_{FF}}$.
On the grounded Gaussian tree above, unit noise gives the independent
increment model itself. Resampling one vertex at a time from its
conditional Gaussian law also samples this joint law. The Gaussian reading, the hitting prediction, and the
potential game thus describe one specified energy and boundary.

\subsection{Realizing the Energy of Agreement}
\label{sec:starkernel}
For proposed observation values $y$, Definition~\ref{def:networkenergy}
gives the agreement cost $\mathcal H_{\rm aff}(y)$; twice this cost
is the cut weight when $y$ is a class indicator. The cost involves
every pair, although the geometry is a tree. The next theorem
realizes it by a sparse network.

A separate terminal carries each observation's prescribed value.
The other vertices transmit agreement and are free to settle.
Eliminating them means minimizing over their values while retaining
the exact energy and response at the observation terminals.

\begin{theorem}[Tree realization of Laplace affinities]
\label{prop:laplacerealization}
Give each edge $\{u,v\}\in\mathcal E$ conductance
\begin{equation}
 c_{uv}=\frac{b_u b_v\rho_{uv}}{1-\rho_{uv}^2},
 \label{eq:laplaceconductance}
\end{equation}
and attach a separate terminal $j'$ to each observation $j\in P$ by an edge of conductance $b_j$. Eliminating the free vertices $W$ leaves $L_K$ on the terminal set $P'=\{j':j\in P\}$. Equivalently, at terminal potentials $y$ the minimum energy of the sparse network is
\begin{equation}
 \min_{f_{P'}=y}\mathcal H(f)=\tfrac12 y^\top L_K y.
 \label{eq:affinityhamiltonian}
\end{equation}
Write $L^{\rm aff}$ for the Laplacian of the full sparse network. The reduction preserves terminal currents, harmonic equilibria, and effective capacities.
\end{theorem}
\begin{proof}
Let $s=\one_P$ and $b=Rs$, and set $g=E^{-1}b=G^{\rm prior}Es$. The rescaled matrix
$Q=\operatorname{diag}(g)L_N\operatorname{diag}(g)$
has off-diagonal entry $-g_cg_p/(\psi(c)-\psi(p))$ on an edge, the negative of \eqref{eq:laplaceconductance}. Its row sum at $u$ is $g_u(L_Ng)_u=b_us_u$. The rescaling is a Doob transform: the row sums are the conductances to the new terminals, with no leakage elsewhere. Hence $Q$ is the interior block of the proposed network. Write $D=\operatorname{diag}(b)$, so $Q^{-1}=D^{-1}RD^{-1}$. With $H_{j,u}=b_j\one\{u=j\}$ the terminal-to-interior conductance matrix, the Schur complement is
\[
 \operatorname{diag}(b_P)-HQ^{-1}H^\top
 =\operatorname{diag}(K\one)-R_{PP}=L_K.
\]
Eliminating by minimizing the quadratic energy gives \eqref{eq:affinityhamiltonian}.
\end{proof}

The resulting terminal matrix is the Schur complement, also called
Kron reduction \citep{dorfler2013kron}; star--mesh transformations
use the same elimination \citep{wagner2018temporal}.
This is an exact representation of the chosen affinity energy, so
the equilibrium and the compensating current of a clamp can be
computed by tree elimination without forming the complete graph.
Section~\ref{app:example} works it on six observations.

The distinction from the preceding covariance calculation matters.
Reducing the original grounded Gaussian tree gives $K^{-1}$ when
$K$ is nonsingular. Reducing the transformed, ungrounded network
above gives $L_K$. The former retains a prior mean and describes
prediction error; the latter penalizes differences between
observation values. Formula~\eqref{eq:greenclamp} applies to either
calculation using its own grounded response matrix and prescribed values.

\paragraph{Total affinity in two passes.}
The total affinities $b_u$ measure how much reference mass is coupled to each vertex; they need no pairwise matrix. Root the tree and write $U_v$ for the affinity mass contributed by observations in the descendant subtree. An upward pass followed by a downward pass gives
\begin{equation}
 U_v=\one\{v\in P\}+\sum_{c:\,c\text{ child of }v}\rho_{vc}U_c,\qquad b_{\rm root}=U_{\rm root},\qquad b_v=U_v+\rho_{pv}(b_p-\rho_{pv}U_v).
 \label{eq:kernelmassfold}
\end{equation}
Hence \eqref{eq:laplaceconductance} is available after linear work in $|W|$.

\paragraph{Changing the geometric ruler.}
\label{sec:rulerenergy}
The realization uses distances on the retained tree, so the same
cuts can support another choice of geometric lengths.
Let $w_v=(R_j(B_v))_{j=1}^d$ be the coordinate widths of a stored box.
Using heights $H_v=\|w_v\|_2/2$ gives nonnegative edge lengths,
since the boxes are nested. With singleton-location leaves, the tree
distance of stored observations is then
$\|w_{\lca(i,j)}\|_2$, which dominates their Euclidean distance.
The contribution of coordinate $j$ to the edge from parent $p$ to
child $v$ is
\begin{equation}
 \frac{(w_{pj}+w_{vj})(w_{pj}-w_{vj})}
      {2(\|w_p\|_2+\|w_v\|_2)} .
 \label{eq:euclideanedge}
\end{equation}
The entries are nonnegative and sum to $H_p-H_v$. Splitting each
width decrement into the two occupied half-width decrements of
Section~\ref{sec:gaugemass} retains all $2d$ directions.
Recomputing affinities from these lengths gives another field on the
same maintained hierarchy; Theorem~\ref{prop:laplacerealization}
still supplies its exact sparse network.

The quadratic power in the field energy acts on differences of
values independently of this spatial ruler. Section~\ref{sec:codispstability}
uses it to read $\sqrt{\operatorname{CoDisp}}$ as a relative
field-change magnitude.

\subsection{Energy within Nested Neighborhoods}
\label{sec:nested}
The tree realization explains why the dense energy can be solved
sparsely. Its nested neighborhoods also explain what the energy
charges for, and how a proposed grouping can be evaluated by counts.
We use the RCF range metric of \eqref{eq:rcfrangemetric} throughout,
with $\varphi(v)=e^{-R(B_v)/h}$; two observations have affinity
$\varphi(\lca(i,j))$.
Write $P_v$ for the reference observations below $v$ and $n_v$ for
their number. Here the tree retains the occupied descendant sets;
Section~\ref{sec:acrosstrees} constructs this representation when
the references are routed through a separately sampled tree.

Put $K_{ii}=1$, the affinity of an observation with itself at its leaf, and $L_K=\operatorname{diag}(K\one)-K$, in which the diagonal cancels. For a vertex $v$, its \emph{degree} is the total affinity of a new observation leaf joining at $v$. It couples to descendants at $\varphi(v)$ and to the remaining references through the ancestors of $v$:
\begin{equation}
 \begin{aligned}
 \deg(v)&=\sum_{j\in P}\varphi(\lca(v,j))=\varphi(v)\,n_v+\Out_v,\\
 \Out_{\rm root}&=0,\qquad \Out_c=\Out_v+\varphi(v)\,(n_v-n_c),
 \end{aligned}
 \label{eq:degree}
\end{equation}
where $c$ is a child of $v$, $\lca(v,j)$ is the lowest common ancestor of $v$ and the leaf of $j$, and $\varphi=1$ at a leaf. One upward pass for the counts and one downward pass for $\Out$ give every degree; the degree of a leaf is the row sum of $K$ for each reference at that leaf.

\paragraph{The same counts construct the network.}
Since $d_T(v,j)=2H_{\lca(v,j)}-H_v$, the total affinities and conductances of Theorem~\ref{prop:laplacerealization} are determined by the degrees \eqref{eq:degree}:
\begin{equation}
 b_v=e^{H_v/h}\deg(v),\qquad
 c_{pc}=\frac{\deg(p)\deg(c)}{\varphi(c)-\varphi(p)},
 \label{eq:circuitspectrum}
\end{equation}
for every vertex $v$ and every edge from a child $c$ to its parent $p$. At the leaf of an observation $j$ that is alone there, $H=0$ and $b_j=1+\sum_{i\neq j}K_{ij}$. Observations that share a leaf $u$ hang from it by separate pendant edges, each of conductance $\deg(u)$: the proof of Theorem~\ref{prop:laplacerealization} goes through with $s$ counting the observations at each vertex. One array of degrees, computed by the two passes of \eqref{eq:degree}, therefore gives the sparse network whose Schur complement is $L_K$, and each induced tree of Section~\ref{sec:acrosstrees} has one.

\begin{proposition}[Computing with nested neighborhoods]
\label{prop:nested}
Use $\varphi(v)=e^{-R(B_v)/h}$ and $K_{ij}=\varphi(\lca(i,j))$ on the range tree of \eqref{eq:rcfrangemetric}, with $\varphi=1$ at observation leaves. Write $\operatorname{par}(v)$ for the parent of $v$, with $\varphi(\operatorname{par}({\rm root}))=0$. Then
\begin{equation}
 K=\sum_v\bigl(\varphi(v)-\varphi(\operatorname{par}(v))\bigr)\,\one_{P_v}\one_{P_v}^\top,
 \label{eq:nested}
\end{equation}
the sum running over all vertices, leaves included. Consequently:
\begin{enumerate}\setlength{\itemsep}{1pt}
\item[(i)] $K$ is positive semidefinite.
\item[(ii)] For $x\in\R^{P\times r}$, let $X_v$ sum the rows of $x$ over $P_v$, upward, and put $O_{\rm root}=0$ and $O_c=O_v+\varphi(v)X_{c'}$ for each child $c$ of $v$ with sibling $c'$, downward. Then $(Kx)_i=X_u+O_u$ for $i$ at leaf $u$. The work is $O(r|T|)$, and $x=\one$ gives the degrees \eqref{eq:degree}.
\item[(iii)] For a labeling of $P$ into $k$ classes, let $n_v(c)$ count the observations of class $c$ in $P_v$. Pair each internal vertex $v$ with the ordered pairs $(a,b)$ and $(b,a)$ of its children, and each leaf $u$ with $(u,u)$. The cut and volume of class $c$,
\[
 \operatorname{cut}(c)=\sum_v\varphi(v)\sum_{(a,b)}n_a(c)\bigl(n_b-n_b(c)\bigr),\qquad
 \operatorname{vol}(c)=\sum_{i:\,c(i)=c}(K\one)_i,
\]
follow from one upward pass of the counts, in $O(k|T|)$.
\end{enumerate}
\end{proposition}
\begin{proof}
For $i\neq j$ the vertices whose sets contain both are the ancestors of $\lca(i,j)$, itself included, and the coefficients in \eqref{eq:nested} telescope to $\varphi(\lca(i,j))$; for $i=j$ they telescope to the value one at the leaf. Every coefficient is nonnegative because $\varphi$ does not decrease downward, so $K$ is a nonnegative combination of positive semidefinite matrices. For (ii), the observations $j$ with $\lca(i,j)=v$, for $v$ a proper ancestor of the leaf $u$ of $i$, are those below the child of $v$ off the path to $u$, and those at $u$ itself have affinity one; summing $\varphi(v)$ times their sums along the path is the downward recursion. For (iii), group the ordered pairs $(i,j)$ with $c(i)=c\neq c(j)$ by $\lca(i,j)$.
\end{proof}

The nonnegative block weights make $K$ a valid covariance, and
this property survives averaging. More directly, the two passes give
the interactions needed for a proposed assignment without writing the
pairwise matrix. The argument uses the retained descendant sets;
Section~\ref{sec:acrosstrees} constructs them for references routed
through a tree that was built from another sample. For weighted
observations, use weighted sums and the product of the two weights
in each pair's energy, including for time-decay weights. The
same formulas hold for any declared node affinity whose increments
are nonnegative, as they are automatically for nested box ranges.

\paragraph{Which neighborhoods charge for disagreement.}
The block decomposition \eqref{eq:nested} expresses the affinity energy as disagreement within the retained neighborhoods. Let $\bar f_v$ be the mean potential on the $n_v$ observations below $v$. Then
\begin{equation}
 \mathcal H_T(f)=\tfrac12\sum_v
    \bigl(\varphi(v)-\varphi(\operatorname{par}(v))\bigr)\,
    n_v\sum_{i\in P_v}(f_i-\bar f_v)^2 .
 \label{eq:neighborhoodhamiltonian}
\end{equation}
Indeed the Laplacian of the block $\one_{P_v}\one_{P_v}^{\top}$ has quadratic form $n_v\sum_{i\in P_v}(f_i-\bar f_v)^2$. For $\varphi(v)=e^{-R(B_v)/h}$, draw a resolution from $\operatorname{Exp}(1/h)$ and take the maximal boxes whose ranges are at most that resolution. A box is selected with probability $\varphi(v)-\varphi(\operatorname{par}(v))$, the chance that the resolution lies between its range and its parent's. The energy is half the expected mass-weighted variation within these boxes. This is where the cut geometry enters a field calculation: it selects the neighborhoods within which disagreement costs energy.

\paragraph{What each merge contributes.}
Write $c_v=n_v[\varphi(v)-\varphi(\operatorname{par}(v))]$. The affinity operator is a finite hierarchical Laplacian \citep{bendikov2014isotropic}: if each observation in $P_v$ is refreshed at rate $c_v$ by a draw uniform on $P_v$, then $i$ jumps to $j\neq i$ at rate $\sum_{v:\,i,j\in P_v}c_v/n_v=\varphi(\lca(i,j))=K_{ij}$, so $-L_K$ is the generator, and its Dirichlet energy is \eqref{eq:neighborhoodhamiltonian}. Split each within-box variation at the merge of the two children $a,b$ of $v$,
\[
 \sum_{i\in P_v}(f_i-\bar f_v)^2=\sum_{i\in P_a}(f_i-\bar f_a)^2+\sum_{i\in P_b}(f_i-\bar f_b)^2+\frac{n_an_b}{n_v}(\bar f_a-\bar f_b)^2,
\]
and collect the coefficient of each merge, $\sum_{w\succeq v}c_w=\deg(v)$ by \eqref{eq:degree}, the sum over $v$ and its ancestors. With one observation at each leaf,
\begin{equation}
 2\mathcal H_T(f)=\sum_{v:\,a,b\text{ children}}\deg(v)\,\frac{n_an_b}{n_v}\,(\bar f_a-\bar f_b)^2;
 \label{eq:wardenergy}
\end{equation}
a leaf $u$ holding several adds $\deg(u)$ times its own within-leaf variation. The factor after $\deg(v)$ is the increase in the within-group sum of squares when the two children merge, the criterion of \citet{ward1963hierarchical}. For $f=\one_A$ the child means are the shares of $A$ in each child, and the left side is the numerator of the normalized cut \eqref{eq:ncut}: each merge charges for how differently its two children support membership in $A$, weighted by the affinity of a virtual observation placed at the merge.

\paragraph{Relaxation across a merge.}
Consider the observation energy $\mathcal H_T(f)=\tfrac12f^\top L_Kf$
with one field coordinate per record. For the children $a,b$ of $v$,
the normalized difference of their means is
\[
 A_v(t)=\sqrt{\frac{n_an_b}{n_v}}\,
              \bigl(\bar f_a(t)-\bar f_b(t)\bigr).
\]
Its contribution in \eqref{eq:wardenergy} is
$\tfrac12\deg(v)A_v^2$. The neighborhood averages within either child
leave this contrast unchanged, whereas averaging over $v$ or an
ancestor sets it to zero. Thus the heat flow generated by $-L_K$
relaxes this contrast at the rate
\begin{equation}
 \E A_v(t)=e^{-t\deg(v)}A_v(0),\qquad
 \deg(v)=\sum_{w\succeq v}c_w.
 \label{eq:mergemodes}
\end{equation}
For Langevin dynamics of these free observation values, projecting
\eqref{eq:langevin} onto the normalized contrast gives
\[
 dA_v(t)=-\deg(v)A_v(t)\,dt+\sqrt{2\vartheta}\,dB_v(t),
\]
where $B_v$ is standard Brownian motion. This is an
Ornstein--Uhlenbeck process: its relaxation time is $1/\deg(v)$
and its stationary variance is $\vartheta/\deg(v)$.
The common offset requires anchoring. Fixing individual observation
values changes the contrast dynamics; the conditioned response in
Section~\ref{sec:game} then gives the uncertainty.

These relaxation rates belong to the observation Laplacian $L_K$.
The sparse realization of Theorem~\ref{prop:laplacerealization}
preserves its equilibrium responses; running dynamics on all
auxiliary vertices need not preserve the same transient rates.

\paragraph{Updating the energy as the data change.}
\label{sec:sampleinsertion}
An insertion or deletion changes the stored boxes and masses along
its path. The new box ranges determine the affinities. Once the
references are located, two passes for the updated counts and degrees
give the sparse-network conductances and merge relaxation rates.
One array therefore links the current geometry, its predictions,
and its dynamics.

For a committed insertion, the effect on existing affinities is
explicit. At fixed $h$, attach $x$ above $v$ and put
$G_w=R(B_w\cup\{x\})-R(B_w)$. An old stored pair keeps its lowest
common ancestor $w$: its affinity is multiplied by $e^{-G_w/h}$
when $w$ is strictly above $v$, and is unchanged otherwise.
Insertion above the old root therefore deepens the old leaves
while preserving their pairwise affinities.
With $s_w=\Prob(v\text{ strictly below }w)$ under the separation law,
the expected old-pair affinity at $w$ is
\[
 e^{-R(B_w)/h}\,[1-s_w(1-e^{-G_w/h})].
\]
It retains the nested form of Proposition~\ref{prop:nested}.
The expected loss summed over unordered pairs of stored copies is
\[
 \sum_{w\text{ on the query path}}m_{w'}m_{w''}\,
 e^{-R(B_w)/h}\,s_w(1-e^{-G_w/h}),
\]
where $w',w''$ are the children of $w$. Thus the changed stored
boxes lie on the insertion path. External references can also be
rerouted by the new cut. Insertion can change the network support,
which the conductance comparisons of Section~\ref{app:recoverynoise}
hold fixed.

Even a fixed mode with drift $-\lambda(A_v-m(t))$ toward a moving
prescribed value $m(t)$ can lag behind it. For $\lambda=\deg(v)>0$,
the mean error $e(t)=\E A_v(t)-m(t)$ obeys
$e'=-\lambda e-m'$. When $|m'|\leq v_0$,
\[
 |e(t)|\leq e^{-\lambda t}|e(0)|
       +\frac{v_0}{\lambda}(1-e^{-\lambda t}).
\]
Tracking changes of the hierarchy further requires a common
representation of the values and control of the changing equilibrium.
Section~\ref{sec:extension} places the trees on common reference
observations; Section~\ref{sec:stability} bounds the response to
finite conductance changes. Section~\ref{sec:referenceupdates}
gives a residual bound for reusing a field on the new references.

\section{Landmarks, Routing, and Normalized Cuts}
\label{sec:extension}
In this section we discuss streaming segmentation based on energy
and Laplacians; in particular we focus on normalized cuts. The
challenge is to create a common reference set $P$, often called
landmarks, which supplies the observations being compared. When
reference values are supplied, the covariance interpretation gives
Kriging, or Gaussian process prediction
\citep{stein1999interpolation,rasmussen2006gaussian}. For grouping,
the affinities define a Laplacian on the same observations.
Routing $P$ gives nested reference sets and counts within each
hierarchy. Inherited box ranges, measured at a common spatial
scale, give the affinities and degrees; their average defines one
energy on $P$. Its leading spectral modes are computed iteratively
from tree products, and Nystr\"om extends the representation to new observations
\citep{fowlkes2004spectral}. Recursive landmarks and multi-centroid
representations have been used in stream segmentation through
weighted center summaries \citep{guha1998cure,guha2003streams}.
Here the RCF supplies the geometry used both to group the references
and to read a new observation, providing an alternative based on
the maintained energy.

Section~\ref{sec:acrosstrees} constructs that geometry.
Section~\ref{sec:cutenergy} reads separation, support, and pooled
prediction uncertainty. Section~\ref{sec:ncut} evaluates normalized
cuts and extends a fixed reference partition.
Algorithm~\ref{alg:referenceclustering} collects the construction
and assignment interface. Section~\ref{sec:transduction} then distinguishes
what a query may change, and Section~\ref{sec:recursiveclamping}
attributes boundary costs and tests changes of membership.
Every tree calculation remains exact within its own hierarchy;
partition optimization uses the chosen normalized-cut procedure.

\subsection{Putting the Trees on Common Observations}
\label{sec:acrosstrees}\label{sec:routed}

The reference set \(P\) contains the observations we want to compare, group, or use as prediction evidence. It can be chosen from a current sample or supplied separately. It need not be the sample that built any one tree, and its members need not have measured response values. \emph{Routing} sends each reference through the existing split decisions to its reached leaf, as in ordinary tree prediction \citep{breiman1984classification,breiman2001random}. The cuts and stored boxes remain fixed. Repeating this in every tree places its affinities on the same observations. For distinct references \(i,j\), let \(v_t(i,j)\)
be their parting vertex in tree \(t\) and define
\begin{equation}
 K_t(i,j)=\varphi_t(v_t(i,j)),\qquad \varphi_t(v)=e^{-R(B^t_v)/h},
 \label{eq:routedkernel}
\end{equation}
Here \(B_v^t\) is the data box of \(v\) in tree \(t\), computed from
that tree's own sample. Set \(K_t(i,j)=1\) when \(i\) and \(j\) reach
the same leaf, including \(i=j\). The forest affinity on \(P\) is
\[
 \bar K=\frac1N\sum_{t=1}^NK_t,\qquad \bar d=\bar K\one,\qquad \bar L=\operatorname{diag}(\bar d)-\bar K.
\]
The sets of references routed through descendants are exactly nested
along each path. Their membership is determined by the partition cuts,
so a routed point can lie outside the data box of a vertex it passes.
For example, in one dimension two points arbitrarily far to the right
of every stored sample reach the same extreme leaf in every tree,
giving affinity one regardless of their separation. The randomized
passive placement uses the query's enlargement of the attachment box
to measure support outside the sample.

\begin{lemma}[Induced trees]
\label{lem:induced}
Route $P$ through tree $t$, and keep the vertices at which points of $P$ part and the leaves they reach. This induced tree $T_t(P)$ has at most $2|P|-1$ vertices, its leaves hold the points of $P$ that reach a common leaf of $t$, and $K_t$ has the form of Proposition~\ref{prop:nested} on it, with $\varphi_t$ inherited from $t$ and one at its leaves. Hence $K_t\succeq0$, $\bar K\succeq0$ with unit diagonal, every point has positive affinity to every other, and a product $\bar Kx$ with $r$ columns costs $O(Nr|P|)$ once the induced trees are built. With the conductances \eqref{eq:circuitspectrum} computed from the degrees of $T_t(P)$, Theorem~\ref{prop:laplacerealization} realizes each $L_t$ exactly.
\end{lemma}
\begin{proof}
A vertex of $T_t(P)$ is a vertex of $t$, and a descendant in $T_t(P)$ is a descendant in $t$, whose data box is nested in its ancestor's; so $\varphi_t$ does not decrease down $T_t(P)$, and $K_t(i,j)=\varphi_t(\lca_{T_t(P)}(i,j))$ by the definition of $v_t$. Every internal vertex of $T_t(P)$ splits its points into two nonempty sets, which bounds the size. Positivity of the affinities follows from $\varphi_t>0$. The rest is Proposition~\ref{prop:nested} and \eqref{eq:circuitspectrum} applied to $T_t(P)$.
\end{proof}

The induced tree uses the original cut and box at each retained
vertex. Its reference counts $n_v=|P_v|$ are computed after routing, while the retained boxes still describe the tree's construction sample. The reference counts of Section~\ref{sec:nested} and the stored sample masses $m_v$ therefore have separate roles. An
out-of-sample row uses the same paths. For a query \(q\), each
sibling that parts from its path supplies its references at the
parting affinity, and references at its leaf have affinity one.
This gives the row and its products with reference vectors in one
traversal per tree, using subtree sums when only the products are
needed. Routing can assign affinity one to two points outside the
sample that reach the same leaf. The randomized placement of
Section~\ref{sec:support} additionally measures enlargement of the
attachment box.

\paragraph{A common spatial scale.}
The scale $h$ is fixed for all pairs being compared. The calibration
boxes of Section~\ref{sec:density} offer a choice tied to mass: take
the median, over trees and reference observations, of the calibration
range at mass $\mu(\theta)$ in \eqref{eq:calibrationmass}. This uses the
same declared neighborhood mass as the ratios of
Section~\ref{sec:ratios}. A query-specific passage box supplies a
local scale for that query; a common affinity field requires a common
choice of $h$.
The experiments use $h$ equal to five percent of the median root
range of the selected trees. The same positive spatial scale is used in the trees being compared.
Compute it once for a reference model and share it across that model's
queries. Across updates, retaining $h$ fixes the comparison scale;
reselecting it changes that scale.

\paragraph{Moving references and reusing computations.}
\label{sec:referenceupdates}
The sampling target matters. For repeated data $D_s=g_s(D)$ with
known point identities, one can sample identities $P\subseteq D$
once and move their coordinates as $P_s=g_s(P)$. A uniform identity
sample remains uniform over the current indexed records.
A reservoir over arriving records instead samples the observed
history \citep{vitter1985reservoir}; these are different reference
laws. Section~\ref{sec:sampling} allows either, once declared.

Cached routes remain valid while their split decisions remain
valid. At an unchanged cut $x_j\lessgtr\tau$, the sufficient condition
$|\Delta x_j|<|x_j-\tau|$ certifies that a moving point stays on its
old side. A changed cut or a structural update requires rechecking
the affected route.
The route and its affinity values are separate: box ranges,
bandwidth, reference counts, and class or spectral sums can change
even when the route does not. Rotation preserves Euclidean distances
but generally changes axis-parallel box ranges.

Previous partitions, spectral vectors, or fields can initialize the
next solve after identifying or evaluating them on the new references.
For a current grounded field system
$L_{FF}f_F^\star=b$, let $g$ be such an initial field and
$r=b-L_{FF}g$ its current residual. If $L_{FF}\succeq\lambda I$,
$\lambda>0$, then
\begin{equation}
 \|g-f_F^\star\|_2\leq\frac{\|r\|_2}{\lambda},\qquad
 \|\E f_F(t)-f_F^\star\|_2
 \leq e^{-\lambda t}\frac{\|r\|_2}{\lambda}
 \quad\text{when }\E f_F(0)=g .
 \label{eq:cachedfield}
\end{equation}
The first bound follows from $f_F^\star-g=L_{FF}^{-1}r$;
the second applies the Langevin mean relaxation of
\eqref{eq:langevin} with the current operator held fixed between
updates. Thus the residual measures what must be corrected, and
the restoring rate determines how much relaxation is needed.

\paragraph{Randomized placement of a passive probe.}
\label{sec:support}
The simplest passive placement puts \(q\) at the leaf reached by routing \(x\). For a randomized
placement, draw the separation vertex \(v\) as in Section~\ref{sec:escape}
and put \(H=R(B_v\cup\{x\})/2\). Place an attachment vertex at height \(H\)
on the ancestral path above \(v\), extending it above the root if needed.
Attach \(q\) there as a height-zero leaf through an edge of length \(H\).
If a reference \(j\) parts from this path at \(w_j\), with \(w_j=v\) for
descendants of \(v\), its passive row entry is
\[
 K_{qj}=\exp\!\left(-\frac{\max\{R(B_v\cup\{x\}),R(B_{w_j})\}}{h}\right).
\]
Subdividing the path preserves all old pairwise distances. The first
contact scale \(D_*\) used in the next subsection is the minimum of
these ranges over the reference set. With a reference below \(v\) it is
\(R(B_v\cup\{x\})\), the same augmented range used in range escape. A
query equal to a stored location uses its leaf, contracting the
zero-length attachment.

For a fixed query, adjoining \(x\) to every box on a chain preserves
containment: the box enclosing \(B_v\cup\{x\}\) is contained in the
one enclosing \(B_p\cup\{x\}\) whenever \(v\) is a child of \(p\).
Subdivision preserves the descendant hierarchy of the references.
A clamp fixes values with the tree and gauge held fixed. As the query,
its placement, or the tree changes, the neighborhoods being compared
can belong to different chains. Exact identities are applied within a
realization before the responses or affinities are pooled.

\subsection{Separation, Support, and Pooling}
\label{sec:cutenergy}
The trees can agree about a prediction while disagreeing about its support. A query may join the references at a small box in one realization and only at a large box in another. We now keep the reference set fixed and average over both the tree and any randomized placement. A mean separation scale alone loses the distinction between these realizations. Let $D$ be the range at which a pair, or a probe and a reference set, first share a box, a random scale over the maintained trees and, for a passive probe, its placements, with law $P_D$. Its pooled affinity defines a separation cost:
\begin{equation}
 \mathcal F_h=-h\log\E e^{-D/h}
 =\inf_{Q\ll P_D}
   \left\{\E_Q D+h\,\operatorname{KL}(Q\|P_D)\right\}.
 \label{eq:separationfreeenergy}
\end{equation}
The minimizing law $Q_h$ tilts $P_D$ by $e^{-D/h}/\E e^{-D/h}$. Subtracting the infimum from the expression in braces gives $h\,\operatorname{KL}(Q\|Q_h)$. Thus short separation paths lower the cost, but favoring a rare path costs information. If a box of range $d$ is found with probability $p$ and the other scales are large against $h$, then $\mathcal F_h\approx d+h\log(1/p)$: finding a close neighbor once is different from finding one reliably.

For a passive probe $q$, fix a common nonempty reference set $P$. Let $\xi$ denote a draw of a tree and a realized height-zero placement, as specified in Section~\ref{sec:support}. In the resulting range tree, $u_*(\xi)$ is the first ancestor of $q$ with a reference descendant, including the leaf when it already contains a reference. Write $D_*(\xi)$ for its range, twice its height. The reference affinities are fixed under placement in each tree. Let $Q_\xi=K_\xi(P,P)+\epsilon I$, $k_\xi=K_\xi(P,q)$ and let $w_\xi$ minimize the per-realization error
\[
 J_\xi(w)=1-2k_\xi^\top w+w^\top Q_\xi w .
\]
A weight $w_i$ specifies the coefficient of reference $i$ in a linear prediction. Choosing a different vector $w_\xi$ in each realization conditions there before averaging. Pooling first instead asks for one common weight vector, $w_\star$, minimizing $\E_\xi J_\xi(w)$. A bar denotes this same average on $K$ and $k$. The pooled variance is
\begin{equation}
 V_\epsilon^{\rm pool}(q)=\min_w\E_\xi J_\xi(w)
        =1-\bar k^\top(\bar K+\epsilon I)^+\bar k.
 \label{eq:proberesiduals}
\end{equation}
The inverse is ordinary for $\epsilon>0$ and may be a pseudoinverse at zero nugget.

The next decomposition answers three questions in order: how far the
query is from any reference on its branch, how firmly the references
constrain it after first contact, and how much is lost by requiring
one predictor to work across the realizations.

\begin{theorem}[Support and pooled Kriging variance]
\label{prop:krigingsupport}
Use the grounded range circuit of Proposition~\ref{prop:green}, with each reference measurement connected through resistance $\epsilon$ and held at zero for the resistance calculation. Let $R^{\rm rem}_{\xi,\epsilon}$ be the effective resistance from $u_*(\xi)$ to those measurements and prior ground after removing the unsupported probe branch below $u_*(\xi)$. Then
\begin{equation}
 \begin{split}
 V_\epsilon^{\rm pool}(q)={}&\E_\xi[1-e^{-D_*(\xi)/h}]
 +\E_\xi R^{\rm rem}_{\xi,\epsilon}\\
 &+\E_\xi[(w_\star-w_\xi)^\top Q_\xi(w_\star-w_\xi)].
 \end{split}
 \label{eq:krigingsupport}
\end{equation}
At positive nugget the last term vanishes exactly when the optimal weights are common to the realizations. The first term is the probability of no reference in the probe's block at exponential resolution. In particular,
\begin{equation}
 V_\epsilon^{\rm pool}(q)\geq 1-e^{-\mathcal F_h(q;P)/h},
 \qquad \mathcal F_h(q;P)=-h\log\E_\xi e^{-D_*(\xi)/h}.
 \label{eq:supportbound}
\end{equation}

At zero nugget, hold the finite ensemble, placement law and ranges fixed as $h\to\infty$. For an edge $e=(p,c)$ of the realized tree let $S_e$ be the probe and reference indices below $c$, and write $R_p,R_c$ for the corresponding vertex ranges, including the attachment heights. Then
\begin{equation}
 \lim_{h\to\infty} hV_0^{\rm pool}(q)
 =\min_{\one^\top w=1}\E_\xi\sum_{e=(p,c)}
 (R_p-R_c)\left(\one\{q\in S_e\}-\sum_{i\in P\cap S_e}w_i\right)^2.
 \label{eq:krigingrangeenergy}
\end{equation}
This is the least mean dissipation of a unit probe current with common signed return currents at the references.
\end{theorem}
\begin{proof}
In the grounded circuit of a realization the diagonal of the conditioned Green's function is the effective resistance from $q$ to the fixed measurements and ground. Every edge from $q$ to $u_*$ carries the full unit current, since no branch off that segment contains a fixed measurement. Its resistance telescopes to $1-e^{-D_*/h}$ and the rest is $R^{\rm rem}_{\xi,\epsilon}$. This gives $J_\xi(w_\xi)$. Expanding the quadratic around $w_\xi$ gives $J_\xi(w)-J_\xi(w_\xi)=(w-w_\xi)^\top Q_\xi(w-w_\xi)$, also when $Q_\xi$ is singular. Averaging at $w_\star$ proves \eqref{eq:krigingsupport}. A resolution $S\sim\operatorname{Exp}(1/h)$ places a reference in the probe's block exactly when $S\geq D_*$, proving the probability claim.

For the limit, inject $1$ at $q$ and $-w_i$ at each reference. The quadratic $J_\xi(w)$ at zero nugget is the power in the grounded tree. The current on each edge is the sum of injections below it, and the current to ground is $1-\one^\top w$. If $R_{\max}$ bounds the ranges in the finite ensemble, the edge resistances multiplied by $h$ lie between $e^{-R_{\max}/h}(R_p-R_c)$ and $R_p-R_c$. The scaled ground resistance is between $he^{-R_{\max}/h}$ and $h$. Thus the minimum is bounded above by the right side of \eqref{eq:krigingrangeenergy}, and bounded below by $e^{-R_{\max}/h}$ times the minimum of that quadratic edge cost plus $h(1-\one^\top w)^2$. These penalized minima tend to the equality-constrained minimum. Indeed, the possible vectors of edge currents and the current to ground form a finite-dimensional closed affine space, after weighting the edge currents by the square roots of their lengths and ensemble probabilities. A bounded objective keeps the weighted currents bounded and drives the ground current to zero. Any limit point is feasible for the equality constraint.
\end{proof}

The decomposition separates unsupported distance, the strength of the reference connection after first contact, and incompatibility of the reconstruction weights across trees. The slope of the unsupported term in $1/h$ at zero is $\E_\xi D_*$. In a tree with its stored sample as reference and uncapped placement, $D_*=R(B_v\cup\{x\})$ at the sampled separation vertex $v$. Dividing the slope by $R(B_{\rm root}\cup\{x\})$ gives the range escape \eqref{eq:rangeescape}. With one reference at separation scale $D$ and zero nugget, the unsupported term is $1-e^{-D/h}$ and the full variance is $1-e^{-2D/h}$. The latter includes resistance beyond first contact. A placement cap or a different reference set changes the first contact scale used in all these formulas.

For this stored-reference, uncapped placement of an unseen query,
Corollary~\ref{cor:affineescapes} gives $D_*\geq\delta$ and
$\E_\xi D_*\leq a\delta$, with the nearest distance $\delta$ and
affine factor $a$ defined there. Concavity of $1-e^{-z/h}$ gives
\begin{equation}
 1-e^{-\delta/h}
 \leq\E_\xi[1-e^{-D_*/h}]
 \leq1-e^{-a\delta/h}.
 \label{eq:affinekrigingsupport}
\end{equation}
The full pooled variance is at least the left side; the upper
bound here concerns its unsupported summand. In the noisy case,
the stopped-range estimate in the proof of
Theorem~\ref{thm:geometryenergy}, applied to the augmented sample,
replaces $a\delta$ by $a_W\delta+b_W$. Thus the affine geometry
also controls the first contribution to uncertainty.

\paragraph{Nystr\"om coordinates and support.}
Let $\bar D=\operatorname{diag}(\bar d)$ and let $z_l$ be an
orthonormal eigenvector of
$\bar D^{-1/2}\bar K\bar D^{-1/2}$ with eigenvalue $\mu_l>0$.
For a probe row $k_q=\bar K(P,q)$ and $d_q=\one^\top k_q$,
its Nystr\"om coordinate is
\begin{equation}
 z_l(q)=\frac1{\mu_l}\sum_{j\in P}
       \frac{k_q(j)z_l(j)}{\sqrt{d_q\bar d_j}}.
 \label{eq:nystromquery}
\end{equation}
For a positive semidefinite augmented affinity with probe diagonal
one, the variance not represented by retained modes $I$ is
\begin{equation}
 V_I^{\rm spec}(q)=1-d_q\sum_{l\in I}\mu_l z_l(q)^2.
 \label{eq:nystromsupport}
\end{equation}
With all positive modes this is the zero-nugget Kriging variance
\eqref{eq:proberesiduals} for the same row. This is a support
diagnostic of the declared geometry. The rotating-star assignment
uses routed rows; its two pre-frame support diagnostics share a
randomized-placement row (Section~\ref{sec:fan}).

\paragraph{Which quantities can be averaged directly.}
\label{sec:ensemblegreen}
Fix a query and supplied reference values. Let $m_T(x)$ and $v_T(x)$
be the conditional prediction mean and variance on a sampled tree.
Mixing these conditional prediction laws under the declared ensemble
law gives mean $\E_Tm_T(x)$ and variance
\[
 \E_Tv_T(x)+\operatorname{Var}_T(m_T(x)).
\]
The first term is uncertainty with the sampled geometry fixed;
the second includes changes in predictions caused by sampling
and tree construction. Pooling before prediction defines other models.

The affinity energies average directly because
$L_{\bar K}=\E L_{K_t}$. Their equilibria need not average to the
equilibrium of that common energy. Gaussian prediction also involves
an inverse: averaging the per-tree Gaussian energies gives precision
$\E[(K_t+\epsilon I)^{-1}]$, whereas averaging the covariances gives
precision $(\E K_t+\epsilon I)^{-1}$. For $\epsilon>0$,
\[
 \E[(K_t+\epsilon I)^{-1}]\succeq(\E K_t+\epsilon I)^{-1}.
\]
Thus these two pooling choices define different prediction models.
For the pooled covariance model just analyzed, the exact excess
over the average per-realization conditioned variance is the
nonnegative common-weight cost in \eqref{eq:krigingsupport}.
This is why the order of solving and pooling is part of the estimator,
even though every one-tree calculation is exact.

Here the pooling rule specifies which prediction problem is solved.
Section~\ref{sec:codispstability} considers a further choice after a
per-tree response has been computed: averaging its squared relative
magnitude, or averaging its relative magnitude and then squaring.
For CoDisp the latter gives a smoothed ensemble reading while keeping
each tree's game and energy fixed. The Kriging calculation and this
reporting choice illustrate different places where the order of
operations matters.

\subsection{Cut Energy}
\label{sec:ncut}
Section~\ref{sec:routed} fixes the reference observations and gives
the tree products and query rows. We now use the affinity Hamiltonian
of Section~\ref{sec:exact} to evaluate a partition on those references.
Each tree contributes through its own exact hierarchy. The pooled
affinity can combine incompatible hierarchies, and the optimization
and spectral extension use the sum of the tree operators.

For a partition $A_1,\ldots,A_k$ of $P$, the normalized cut \citep{shi2000normalized} under $\bar K$ is
\begin{equation}
 \operatorname{Ncut}(A)=\sum_{c=1}^k\frac{\operatorname{cut}(A_c)}{\operatorname{vol}(A_c)},\qquad
 \operatorname{vol}(A)=\sum_{i\in A}\bar d_i,\qquad
 \operatorname{cut}(A)=\operatorname{vol}(A)-\sum_{i,j\in A}\bar K_{ij}.
 \label{eq:ncut}
\end{equation}

For the walk with transitions \(\bar K_{ij}/\bar d_i\), the ratio
\(\operatorname{cut}(A)/\operatorname{vol}(A)\) is the probability
of leaving \(A\) in one step conditional on starting in \(A\) under
the stationary law \citep{meila2001random}.

For a public boundary on tree \(t\), clamp each reference terminal
to its membership in \(A\). If \(\mathcal H_t^{\rm aff}\) is the energy
of the sparse affinity network for \(K_t\), then
\begin{equation}
 \operatorname{cut}(A)
 =\one_A^\top\bar L\one_A
 =\frac{2}{N}\sum_{t=1}^N\min_{f_{P'}=\one_A}\mathcal H_t^{\rm aff}(f).
 \label{eq:publicclamp}
\end{equation}
This follows from \eqref{eq:affinityhamiltonian}. The free voltages
respond to the whole coalition boundary. Each cut term compares the
current at unit voltage across the boundary with the total degree of
the members. Section~\ref{sec:contours} reads the same cut as the share of
affinity that the members of each class give up, and
Section~\ref{sec:collusivecuts} treats relocation as a potential game.

\paragraph{Which scales and coordinates carry a cut.}
\label{sec:spectralrecipe}
Let $0=\lambda_1<\lambda_2\leq\cdots$ be the generalized
eigenvalues of $(\bar L,\bar D)$ and put $\mu_l=1-\lambda_l$.
The random-walk interpretation is that of \citet{meila2001random};
the multiway spectral start and Lloyd refinement use established
normalized-cut procedures \citep{ng2001spectral,dhillon2004kernel}.
The elbow uses ratios $\lambda_{k+1}/\lambda_k$ with
$\lambda_k>0$. Testing counts up to $k_{\max}$ needs the first
$k_{\max}+1$ eigenvalues; the corresponding $k$ slow modes give
the row embedding of the spectral start.

To examine coordinate $j$, multiply its box width by $\theta_j$
in the affinity, holding cuts and bandwidth fixed. Put
$V_j=-\partial_{\theta_j}\bar K$ at $\boldsymbol\theta=\one$;
its $(i,p)$ entry is the forest mean of
$R_j(B_v)\varphi(v)/h$ at their parting boxes and is zero within
a leaf. For selected modes $I$ separated from the remaining
spectrum, take $\bar D$-orthonormal eigenvectors
$\bar K u_l=\mu_l\bar D u_l$ with
$u_l^\top\bar D u_l=1$. Differentiating this equation gives
\begin{equation}
 I_j:=\partial_{\theta_j}\sum_{l\in I}\mu_l
 =\sum_{l\in I}\left[-u_l^\top V_j u_l
       +\mu_l\sum_i u_l(i)^2(V_j\one)_i\right].
 \label{eq:spectralimportance}
\end{equation}
The second term includes the changing degrees. A positive $I_j$
reduces the sum of the selected relaxed cuts. Importance per share
divides $I_j/\sum_{a=1}^d|I_a|$ by the coordinate's fraction of
affinity-weighted range,
$\one^\top V_j\one/\sum_{a=1}^d\one^\top V_a\one$.
The selection ranks positive ratios with positive range share and
keeps those above the largest consecutive ratio when it reaches
the gap threshold used for relaxed cuts. If no ratio is positive,
it keeps every coordinate. The experiment's final ablation
recomputes bandwidth after selection; the derivative itself has
fixed bandwidth.

\paragraph{Extending a fixed reference partition.}
Having grouped $P$, assign a new observation using that frozen
construction. The tree geometry, bandwidth, reference degrees,
and fitted class statistics are held fixed. The query supplies
only its routed row $k_q=\bar K(P,q)$ and
$d_q=\one^\top k_q$.

The fixed-class mean rule compares the query with the existing
class means in the normalized affinity geometry.
For $V_A=\operatorname{vol}(A)$ and
$B_A=\sum_{i,j\in A}\bar K_{ij}$, it chooses the class minimizing
\[
 -\frac{2\bar K(q,A)}{d_qV_A}+\frac{B_A}{V_A^2}.
\]
The Nystr\"om readout instead extends the spectral start by
\eqref{eq:nystromquery}, normalizes the resulting row, and assigns
it to an existing center of that spectral embedding
\citep{fowlkes2004spectral}.
The experiment uses the fixed-class mean rule after refinement and
compares it with the Nystr\"om readout of the spectral start.
These are two specified readouts of the frozen construction.
Joining the query to the graph and recomputing a partition would
ask a different question, treated in Section~\ref{sec:transduction}.

\begin{algorithm}[H]
\caption{Constructing a reference clustering and assigning an observation}
\label{alg:referenceclustering}
\small
\textbf{Input:} A forest snapshot $T_1,\ldots,T_N$, references $P$,
a common $h>0$, a normalized-cut procedure with declared
cluster-count and refinement choices, and a readout
(fixed-class means or Nystr\"om).\par\smallskip
\textbf{Construct reference model}
\begin{enumerate}\setlength{\itemsep}{2pt}\setlength{\parskip}{0pt}
\item Route $P$ through every tree. Build the induced hierarchies
and reference counts, inheriting the stored box ranges.
\item Compute degrees and supply products with
$\bar K=N^{-1}\sum_tK_t$ and
$\bar L=\operatorname{diag}(\bar K\one)-\bar K$
by the tree passes.
\item Run the chosen normalized-cut procedure. For fixed-class means,
retain the refined reference partition with its $V_A$ and $B_A$.
For Nystr\"om, retain the spectral start's partition, selected
eigenpairs, and centers.
\item Store subtree sums of class indicators or of
$z_l(j)/\sqrt{\bar d_j}$ for the chosen readout.
Return the model $M$ with its snapshot, $P$, $h$, and fitted statistics.
\end{enumerate}
\textbf{Assign$(M,x)$}
\begin{enumerate}\setlength{\itemsep}{2pt}\setlength{\parskip}{0pt}
\item Route $x$ through the same snapshot. Use the stored subtree
sums to obtain $d_q$ and the class-affinity sums or coordinates
\eqref{eq:nystromquery}.
\item Apply the declared readout and return the class with the
computed affinity summaries. The reference model remains fixed.
\end{enumerate}
\end{algorithm}

The passes give exact products and cut evaluations after routing;
finding the partition takes the iterations of the chosen solver.
The same fitted statistics serve repeated probes. Model updates
can reuse routes and initial solutions under
Section~\ref{sec:referenceupdates}'s conditions.

\subsection{What the Query Changes}
\label{sec:transduction}
The construction above assigns an observation using fixed
references. Other query intents change different parts of the
calculation. Write $x$ for the feature vector and $q$ for its
vertex in a field calculation.

A \emph{passive} query reads a prediction, class, or support against
fixed reference affinities and values. It may use routing, or the
randomized attachment of Section~\ref{sec:support} when its distance
outside the stored support matters. A \emph{self-referential} query
also prescribes the value or class to be tested. Its clamp lets
other free values respond; joining a class changes the class and
degree accounting. Sections~\ref{sec:privatecoalitions}
and~\ref{sec:collusivecuts} price those changes.

RCF scoring considers \emph{participation}: the separation law
averages over where $x$ would join the tree. Inverse depth reads
the inserted point's private escape, and displacement its effect
on the old records. These virtual insertions need not commit an
update. An actual insertion changes stored boxes and may reroute
external references, as in Section~\ref{sec:sampleinsertion}.

\section{Clamping, Games, and Cuts}
\label{sec:recursiveclamping}
Section~\ref{sec:exact} supplied a Hamiltonian, and
Section~\ref{sec:extension} put the trees on common references.
With the reference values fixed, minimizing the energy fills the
unknown values with the least disagreement. An intervention asks:
if we insist on a different value or class, how must the remaining
predictions adjust, and what additional energy is required?
A clamp imposes that requirement; releasing it measures the energy
saved when the requirement is removed. Moving a group between
classes also changes its boundary and the volumes used to normalize
its cost.

The group-clamp identity prices the first change. Stopping and
participation games then specify which members can act together.
CoDisp reads one release on a mass-resistance path. Finally, the
same descendant sets attribute the cost of a fixed partition and
supply joint moves that improve a cut even when smaller moves fail.

\subsection{Imposing and Releasing a Boundary}
\label{sec:privatecoalitions}
A private prediction or escape reads one coordinate of the harmonic
field under fixed evidence. To test an intervention, impose a new
boundary and let the other voltages adjust. On a fixed circuit let
\(f^\star\) be the equilibrium and \(G=(L_{FF})^{-1}\) the Green's
function on its free vertices. Prescribing \(z\) on a set \(C\subseteq F\)
increases the minimized energy by
\begin{equation}
 \mathcal H_C(z)-\mathcal H(f^\star)
   =\tfrac12(z-f_C^\star)^\top(G_{CC})^{-1}(z-f_C^\star).
 \label{eq:groupclamp}
\end{equation}
The required current on \(C\) is \((G_{CC})^{-1}(z-f_C^\star)\).
The old equilibrium cancels the linear term in the energy, leaving
half the current paired with the voltage change. A singleton clamp
gives the audit \eqref{eq:probeaudit}. Releasing a clamp measures the
same difference in the other direction.

For the range-affinity Hamiltonian on the common stored observations
of Theorem~\ref{thm:geometryenergy}, fix a nonempty clamp set $A$
and values $y_A$. Write
$\mathcal E^{(1)}_s(y_A)=\min_{f_A=y_A}\mathcal H^{(1)}_s(f)$.
Minimizing the pointwise energy comparison over the same free values
preserves its order:
\begin{equation}
 \mathcal E^{(1)}_{h/a}(y_A)
 \leq\min_{f_A=y_A}\E_T\mathcal H_h^T(f)
 \leq\mathcal E^{(1)}_h(y_A).
 \label{eq:affineboundaryenergy}
\end{equation}
The noisy version uses $a_W$ and multiplies the lower bound by
$e^{-b_W/h}$. These are total boundary costs for the pooled
range-affinity energy. The incremental price in
\eqref{eq:groupclamp} is read from its Green's function; the mass
path used for CoDisp below retains its own conductances.

% Figure: Proposition prop:laplacerealization on six observations, with the quantities of Sections 4 and 5.
% Edge widths in (b) encode log(1+c_uv); cross-void affinities in (a) are below 0.01; curves in (c) are exact.
\definecolor{labA}{HTML}{D55E00}
\definecolor{labB}{HTML}{0072B2}
\tikzset{
  obs/.style={circle,fill=black,inner sep=0pt,minimum size=3.2pt},
  lat/.style={circle,fill=black!40,inner sep=0pt,minimum size=3.2pt},
  term/.style={draw=black,line width=0.4pt,fill=white,inner sep=0pt,minimum size=8pt,
               font=\fontsize{6}{6}\selectfont},
  colA/.style={term,fill=labA,text=white},
  colB/.style={term,fill=labB,text=white},
  probe/.style={diamond,draw=black,line width=0.5pt,fill=white,inner sep=0pt,minimum size=6.5pt},
  ptitle/.style={font=\footnotesize,anchor=south,inner sep=0pt},
  lab/.style={font=\scriptsize,inner sep=1pt},
}
\begin{figure}[!htb]
\centering
% (a) complete Laplace-affinity graph on the six terminals
\begin{tikzpicture}[baseline=(base),scale=0.92]
  \coordinate (base) at (0,-0.1);
  \coordinate (n1) at (0.60, 0.80); \coordinate (n2) at (0.10,-0.40);
  \coordinate (n3) at (0.95,-0.75); \coordinate (n4) at (2.85, 0.75);
  \coordinate (n5) at (3.70, 0.40); \coordinate (n6) at (3.20,-0.80);
  \foreach \a in {n1,n2,n3}{\foreach \b in {n4,n5,n6}{\draw[black!25,line width=0.3pt] (\a)--(\b);}}
  \draw[line width=1.1pt] (n1)--(n2) (n1)--(n3) (n4)--(n6) (n5)--(n6);
  \draw[line width=1.4pt] (n2)--(n3) (n4)--(n5);
  \node[colA] at (n1) {1}; \node[term] at (n2) {2}; \node[term] at (n3) {3};
  \node[term] at (n4) {4}; \node[colB] at (n5) {5}; \node[term] at (n6) {6};
  \draw[densely dashed,line width=0.5pt] (1.9,-1.05)--(1.9,1.05) node[above,lab] {cut};
  \node[ptitle] at (1.9,1.55) {(a) complete graph};
\end{tikzpicture}%
\hspace{4pt}%
\begin{tikzpicture}[baseline=(base)]
  \coordinate (base) at (0,-0.1);
  \node at (0,0.05) {$\equiv$};
  \node[lab] at (0,-0.35) {Thm.~\ref{prop:laplacerealization}};
\end{tikzpicture}%
\hspace{4pt}%
% (b) sparse network on the tree: pendant terminals (squares) and probes (diamonds)
\begin{tikzpicture}[baseline=(base),scale=0.92]
  \coordinate (base) at (0,-0.1);
  \coordinate (u) at (-1.15, 0.00);  \coordinate (v) at (1.15, 0.00);
  \coordinate (o1) at (-1.65, 0.55); \coordinate (o6) at (1.65,-0.55);
  \coordinate (w) at (-1.75,-0.42);  \coordinate (x) at (1.75, 0.42);
  \coordinate (o2) at (-2.45,-0.20); \coordinate (o5) at (2.45, 0.20);
  \coordinate (o3) at (-2.05,-0.88); \coordinate (o4) at (2.05, 0.88);
  \draw[line width=1.05pt] (u)--(o1) (v)--(o6);
  \draw[line width=1.50pt] (u)--(w)  (v)--(x);
  \draw[line width=1.20pt] (w)--(o2) (w)--(o3) (x)--(o4) (x)--(o5);
  \draw[line width=0.30pt] (u)--(v);
  \draw[line width=0.75pt] (o1)--++(-0.42, 0.32) node[colA] {1};
  \draw[line width=0.75pt] (o2)--++(-0.50, 0.05) node[term] {2};
  \draw[line width=0.75pt] (o3)--++(-0.45,-0.27) node[term] {3};
  \draw[line width=0.75pt] (o4)--++( 0.45, 0.27) node[term] {4};
  \draw[line width=0.75pt] (o5)--++( 0.50,-0.05) node[colB] {5};
  \draw[line width=0.75pt] (o6)--++( 0.42,-0.32) node[term] {6};
  \foreach \p in {o1,o2,o3,o4,o5,o6} \node[obs] at (\p) {};
  \foreach \p in {u,v,w,x} \node[lat] at (\p) {};
  \node[probe] (q1) at ($(u)!0.5!(w)$) {};
  \node[probe] (q2) at (0,0) {};
  \node[lab,anchor=north west] at (q1.south east) {$q_1$};
  \node[lab,anchor=north] at (q2.south) {$q_2$};
  \node[lab,anchor=south] at (0,0.10) {void edge $e$};
  \node[ptitle] at (0,1.55) {(b) sparse network on the tree};
\end{tikzpicture}%
\hspace{12pt}%
% (c) audit at each probe: H_q(z) - H_q(f(q)) = (a_q/2) (z - f(q))^2
\begin{tikzpicture}[baseline=(base),xscale=2.95,yscale=2.94]
  \coordinate (base) at (0,0.265);
  \draw[line width=0.4pt] (0,0)--(1.07,0) node[right,lab] {$z$};
  \draw[line width=0.4pt] (0,0)--(0,0.6);
  \foreach \t in {0,1} \draw[line width=0.4pt] (\t,0)--(\t,-0.02) node[below,lab] {$\t$};
  \draw[line width=0.4pt] (0.501,0)--(0.501,-0.02) node[below,lab] {$f(q_2)$};
  \draw[line width=0.4pt] (0.955,0)--(0.955,0.02) node[above,lab,xshift=-6pt] {$f(q_1)$};
  \draw[line width=0.9pt,domain=0:1,samples=60] plot (\x,{0.63785*(\x-0.955)^2});
  \draw[line width=0.9pt,densely dashed,domain=0:1,samples=60] plot (\x,{0.11925*(\x-0.501)^2});
  \node[lab,anchor=west] at (0.05,0.55) {$q_1$};
  \node[lab,anchor=south] at (0.17,0.035) {$q_2$};
  \node[lab,anchor=west] at (0.33,0.36) {$\tfrac12 a_q\,(z-f(q))^2$};
  \node[ptitle] at (0.5,0.7775) {(c) probe audits};
\end{tikzpicture}
\caption{Six observations in two clusters, their complete affinity graph
(a), and its exact sparse realization (b). Edge widths in (b) show
conductance, squares are observation terminals, and diamonds are
probes. With terminals 1 and 5 held at one and zero, (c) shows the
energy needed to prescribe a value $z$ at each probe. The probe in
the void has a prediction near one half and a much flatter audit.}
\label{fig:realization}
\end{figure}
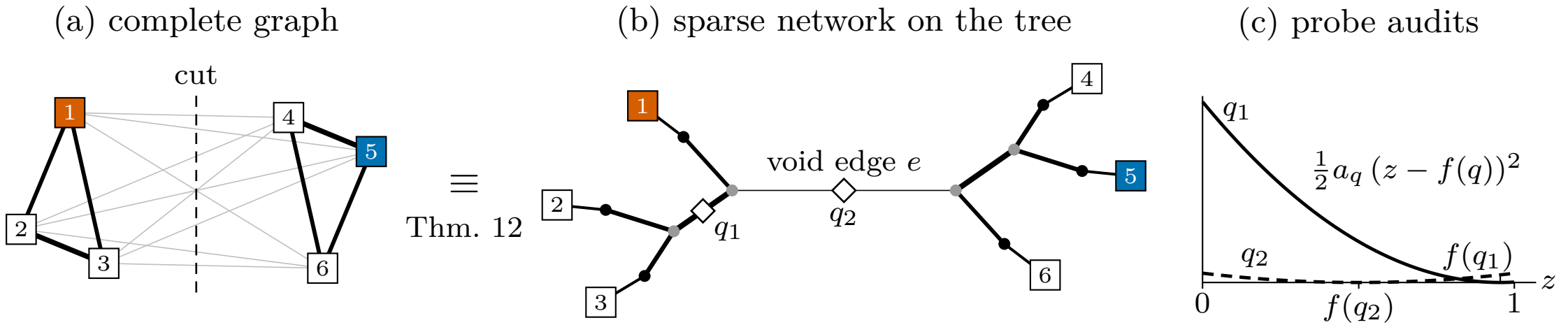

\paragraph{Two clusters and a void.}
\label{app:example}
Figure~\ref{fig:realization} applies the realization of
Theorem~\ref{prop:laplacerealization} to six observations.
Use dimensionless leaf-edge lengths $0.35$ at observations 1 and 6
and $0.30$ at the other four, inner-edge lengths $0.25$, and void
length $4$; each probe subdivides its indicated edge at the midpoint.
The probe inside the left cluster reads $0.955$; the probe in the
void reads $0.501$. Their reference supports are $1.96$ and $0.50$,
and their audit stiffnesses $1.28$ and $0.24$. Thus the same
deviation from the predicted value costs about a fifth as much
energy in the void. The prediction is available in both places;
the clamp calculation records how strongly the evidence holds it.

\subsection{Passage Times and Stopping Coalitions}
\label{sec:passagetimes}
Changing an absorbing boundary changes both the first place reached
and the cost collected before stopping. On the fixed finite network
of Section~\ref{sec:game}, let $A$ be absorbing and $F$ its complement,
with every component meeting $A$. At each visit to $v$, spend mean
time $\eta_v>0$, then choose the next vertex with probability
$c_{vw}/d_v$, where $d_v=\sum_wc_{vw}$.
Fix a vector of nonnegative cost weights $\omega_v$. Its value is
\[
 T_v^\omega=\E_v\int_0^{\tau_A}\omega_{X_t}\,dt.
\]
The walk is fixed while these weights specify what is counted.
Unit weights give mean passage time. The first-step equation is
\begin{equation}
 T_A^\omega=0,\qquad T_v^\omega=\eta_v\omega_v
           +\sum_w\frac{c_{vw}}{d_v}T_w^\omega,
 \qquad (L_{FF}T_F^\omega)_v=d_v\eta_v\omega_v.
 \label{eq:passagetimepoisson}
\end{equation}
Choosing stopping vertices to reduce passage time is studied as
absorbing random-walk centrality \citep{mavroforakis2015absorbing}.
Its diminishing-returns property also holds for the nonnegative
weighted costs used here.

\begin{proposition}[Weighted passage costs and stopping]
\label{prop:stoppinggame}
The values $T^\omega$ are the unique minimizer, with $u_A=0$, of
\begin{equation}
 \Phi_\omega(u)=\tfrac12u^\top Lu
               -\sum_{v\in F}d_v\eta_v\omega_vu_v.
 \label{eq:passagetimepotential}
\end{equation}
Equivalently, each free vertex chooses $u_v$ at cost
$\tfrac12\sum_wc_{vw}(u_v-u_w)^2-d_v\eta_v\omega_vu_v$.
Its best response is \eqref{eq:passagetimepoisson}, and
$\Phi_\omega$ is an exact potential.
For $G=L_{FF}^{-1}$ and an additional absorbing vertex $z\in F$,
\begin{equation}
 T_v^\omega-T_v^{\omega,A\cup\{z\}}
 =\frac{G(v,z)}{G(z,z)}T_z^\omega
 =\Prob_v(\tau_z<\tau_A)T_z^\omega,\qquad v\in F.
 \label{eq:stoppingclamp}
\end{equation}
For a fixed starting query $q$, the worth
$v_q^\omega(C)=T_q^\omega-T_q^{\omega,A\cup C}$ is normalized,
monotone, and submodular in the additional absorbing members
$C\subseteq F$.
\end{proposition}
\begin{proof}
Differentiating \eqref{eq:passagetimepotential} gives the first-step
equation, and $L_{FF}$ is positive definite. A unilateral change has
the same cost in the player's objective and in $\Phi_\omega$.
Adding $z$ is the clamp $u_z\mapsto0$; the Green response and the
strong Markov property give \eqref{eq:stoppingclamp}.
For the last claim, couple all stopped walks to the original walk
up to its hit of $A$. On a trajectory, let
$Y_z=\int_{\tau_z}^{\tau_A}\omega_{X_t}\,dt$ if $z$ is visited
before $A$, and zero otherwise. Since $\omega\geq0$, the cost saved
by stopping at the first member of $C$ is $\max_{z\in C}Y_z$.
A maximum of nonnegative single-member rewards is monotone and
submodular. Taking expectations preserves both properties.
\end{proof}

The response is linear in the weights:
$T_F^\omega=G\operatorname{diag}((d_v\eta_v)_{v\in F})\omega_F$.
On a tree the same upward elimination and downward response compute
it. Taking the stopping coalition to be a descendant set supplies
concrete worths for the restricted-coalition construction below.

\subsection{Which Coalitions Can Act}
\label{sec:participation}

Let \(P\) be the players at the leaves of a finite rooted tree,
and \(S_u\subseteq P\) the descendants of vertex \(u\). The
children of each internal vertex partition its players into at
least two nonempty sets. A family of nested intervals
with disjoint record sets at each fork is an example. Give each
\(S_u\) a transferable worth \(\nu(S_u)\in\R\) and set
\(\nu(\emptyset)=0\). At \(u\), the increment for acting together
is
\[
 \delta_u=\nu(S_u)-\sum_{w\text{ child of }u}\nu(S_w).
\]
It can have either sign. A coalition \(A\) formed from a subset
of the leaves can act through the maximal descendant sets
contained in \(A\). With this convention its worth is
\begin{equation}
 \nu_{\mathcal L}(A)
 =\sum_{u:S_u\subseteq A}\delta_u
 =\sum_{u:S_u\text{ maximal among descendant sets in }A}\nu(S_u).
 \label{eq:laminarworth}
\end{equation}
The equality follows by telescoping inside each maximal block.
In particular \(\nu_{\mathcal L}(S_u)=\nu(S_u)\). If a game
specifies a separate worth for a set crossing branches, the
extension agrees there exactly when that worth is the sum over
its maximal blocks.

This is a restriction of communication. Take every nonsingleton descendant set as a conference in the sense of \citet{myerson1980conference}, which can act only when all of its members participate. Inside any coalition the conferences that can meet join the players into its maximal descendant sets, so \eqref{eq:laminarworth} is the laminar case of the hypergraph communication restriction \citep{vandennouweland1992allocation}. Its first sum writes $\nu_{\mathcal L}$ as a combination of unanimity games on the descendant sets, with the increments $\delta_u$ as the Harsanyi dividends \citep{harsanyi1963simplified}. Paying each participant its marginal contribution to a common worth makes that worth an exact potential, a standard construction. The following construction puts the worths into preferred edge increments of one positive quadratic energy. The hierarchy supplies the admissible participation states and the recursion under partial clamps.

For a participation profile $A$, put
$z_u(A)=\one\{S_u\subseteq A\}$. At a leaf this is its participation
choice; at an internal vertex it is the product of its children's
activation values. These logical constraints specify the admissible
states of the following quadratic energy.

\begin{theorem}[A quadratic Hamiltonian for laminar participation]
\label{prop:laminarparticipation}
Give the edge entering each vertex $u$ conductance $c_u>0$.
The parent of the root is an additional ground $\partial$, held
at $z_\partial=0$. Let $Q$ be this tree's grounded Laplacian, and put
\begin{equation}
 \beta_u=\tfrac12+\frac{\nu(S_u)}{c_u},\qquad
 \mathcal H_\nu(z)=\tfrac12\sum_{u\in V(T)}c_u
                (z_u-z_{p(u)}-\beta_u)^2.
 \label{eq:laminarhamiltonian}
\end{equation}
The worths $\nu(S_u)$ may have either sign. Then:
\begin{enumerate}\setlength{\itemsep}{1pt}
\item[(i)] For $\vartheta>0$, the continuous density proportional to
$e^{-\mathcal H_\nu/\vartheta}$ is Gaussian, with covariance
$\vartheta Q^{-1}$ and mean given by
$\mu_\partial=0$, $\mu_u-\mu_{p(u)}=\beta_u$.
Its quadratic part $Q$ is independent of the worths.
\item[(ii)] For every coalition $A\subseteq P$,
\begin{equation}
 \mathcal H_\nu(0)-\mathcal H_\nu(z(A))=\nu_{\mathcal L}(A).
 \label{eq:laminarenergyworth}
\end{equation}
\item[(iii)] Pay a participating leaf $i$ its marginal
$\nu_{\mathcal L}(A)-\nu_{\mathcal L}(A\setminus\{i\})$ and a
nonparticipant zero. This is an exact potential game with potential
$\nu_{\mathcal L}=\mathcal H_\nu(0)-\mathcal H_\nu\circ z$.
Under any partial assignment of participation, one upward and one
downward pass find a pure Nash equilibrium maximizing this potential
among profiles satisfying the clamps.
\end{enumerate}
\end{theorem}
\begin{proof}
The change from vertex values to edge increments is invertible with
the ground fixed. The increments are independent Gaussians of means
$\beta_u$ and variances $\vartheta/c_u$, proving (i).
For an activation profile, $z_{p(u)}\leq z_u$ and both are Boolean.
Thus $z_u-z_{p(u)}$ is one exactly at the roots of the maximal
participating descendant sets. On such an edge, changing the
increment from zero to one reduces energy by
$c_u(\beta_u-1/2)=\nu(S_u)$. Summing these reductions gives
\eqref{eq:laminarworth}, proving (ii).

When \(i\) joins, its payoff increases by precisely the increase
in \(\nu_{\mathcal L}\). The marginal is the sum of the increments
at the fully participating ancestors of \(i\), including its
singleton. Leaving reverses that change. This is
the exact potential property \citep{monderer1996potential}, also
after any of the other choices have been fixed. A constrained
global maximizer is therefore a Nash equilibrium among the free
players.

To find it, let \(F_u\) be the best worth with all players of
\(S_u\) participating, and \(Q_u\) the best with a proper subset
participating. Each state respects the clamps below \(u\), with
value \(-\infty\) when infeasible. At a leaf, the two candidate
values are its singleton worth and zero. At an internal node,
\(F_u=\nu(S_u)\) if all its children can be full. For a proper
selection at least one child must be proper. The worth then
adds across the children by \eqref{eq:laminarworth}, so
\begin{equation}
 Q_u=\max_{w\text{ child of }u}
 \left[Q_w+\sum_{z\ne w\text{ child of }u}\max(F_z,Q_z)\right].
 \label{eq:clampfold}
\end{equation}
Sum the best child values once and use the loss of requiring
each child to be proper to evaluate the maximum. The root
takes the better of its two states. Following the stored
choices downward recovers a maximizing participation profile.
The work is linear in the number of vertices given the worth
table.

\end{proof}

The continuous field allows real vertex values. Participation uses
the finite set $\mathcal Z_T=\{z(A):A\subseteq P\}$ with its Boolean
activation constraints; the two-pass recursion minimizes the same
Hamiltonian on that set. With counting measure on participation
profiles, its Gibbs weights are proportional to
$e^{\nu_{\mathcal L}(A)/\vartheta}$. At each step choose a free
player uniformly; its logit update chooses participation or departure
with probabilities proportional
to the exponentials of their payoffs divided by $\vartheta$
\citep{blume1993statistical}. For two profiles differing only in
that player's choice, the ratio of transition probabilities is
$e^{[\nu_{\mathcal L}(A')-\nu_{\mathcal L}(A)]/\vartheta}$.
Detailed balance therefore gives precisely these Gibbs weights,
also after participation choices have been clamped. The finite
participation law and the continuous Gaussian law use the same
energy on different state spaces. Expanding the squares in
\eqref{eq:laminarhamiltonian} gives a linear forcing
\[
 b_u=c_u\beta_u-\sum_{w:\,p(w)=u}c_w\beta_w
     =\tfrac12\left(c_u-\sum_{w:\,p(w)=u}c_w\right)+\delta_u.
\]
The Gaussian mean solves $Q\mu=b$, the Poisson equation of the
weighted passage calculation.

\paragraph{Dividends as sources.}
Let $\mu^0$ be the mean for zero worths, with forcing
$b^0_u=\tfrac12(c_u-\sum_{w:p(w)=u}c_w)$.
The worth-dependent response is exactly
\begin{equation}
 Q(\mu-\mu^0)=\boldsymbol\delta,\qquad
 \mu-\mu^0=Q^{-1}\boldsymbol\delta.
 \label{eq:dividendsources}
\end{equation}
Thus the Harsanyi dividends are sources and the worth-free Green's
function propagates them. Taking conductances that add over
children, for example $c_u=|S_u|$, makes $b^0_u=0$ at every
internal vertex. The leaves retain the baseline source $c_u/2$;
subtracting $\mu^0$ removes that baseline at all vertices.

\paragraph{Vector-valued fields.}
The same construction allows vector values: replace a squared edge
increment by its quadratic form under a fixed $W\succ0$.
The precision becomes $Q\otimes W$, so
$\operatorname{Cov}(F_u,F_v)=\vartheta(Q^{-1})_{uv}W^{-1}$.
In particular a fixed linear measurement $a^\top F$ has covariance
$\vartheta(a^\top W^{-1}a)Q^{-1}$. These formulas follow from the
independent Gaussian edge increments and persist under whole-vector
clamps with the conditioned Green's function.

To compare \(A\subset B\), hold the members of \(A\) in and
everyone outside \(B\) out. Pinning \(B\setminus A\) out or in
reads the two prescribed coalition worths. Freeing those
individual choices instead asks for the best participation
profile between \(A\) and \(B\), which the fold supplies. On a
query path the change from one descendant set to its parent
adds exactly the sibling's records.

The worths and unilateral gains are the quantities preserved
by this construction. Its maximizing equilibrium is selected by
the recursion. Other equilibria can occur. For example, if two
singletons are worth \(-1\) each and their union is worth \(1\),
both staying out is an equilibrium, while both joining maximizes
the potential. The marginal payoffs describe participation
incentives. If a budget-balanced allocation is desired, the
Shapley rule for \(\nu_{\mathcal L}\) gives each \(i\in S_u\)
the share \(\delta_u/|S_u|\) of that node's increment, as for any game
with these dividends. In a random
arrival order each member is equally likely to be the last to
complete \(S_u\), so summing these shares down each leaf path
gives the allocation.

\subsection{CoDisp as a Relative Change of the Field}
\label{sec:codispstability}

The mass escape of Corollary~\ref{cor:displacement} gives the expected
normalized displacement. The same corollary writes the CoDisp cut ratio
\(b/a\) as odds, where \(a\) is the query-side mass and \(b\) the
sibling mass. There is also a variational reading on the mass-resistance
path. Let \(v_0,\ldots,v_\ell\) be the path to the root and \(m_i\)
the cumulative resistance from the query-side ground to \(v_i\),
including the query mass. Set all voltages at and above \(v_i\) to one
and let \(u_i\) minimize the same quadratic Hamiltonian. Releasing the
constraint at \(v_i\) extends the active prefix from \(a=m_i\) to
\(a+b=m_{i+1}\), so the minimum falls from \(1/(2a)\) to \(1/(2(a+b))\).

Since \(u_i-u_{i+1}\) is a feasible variation at the relaxed minimum,
it is energy-orthogonal to \(u_{i+1}\). Writing
\(\|u\|_L^2=2\mathcal H(u)\), we obtain
\begin{equation}
 \frac{\|u_i-u_{i+1}\|_L^2}{\|u_{i+1}\|_L^2}
   =\frac{\mathcal H(u_i)-\mathcal H(u_{i+1})}
          {\mathcal H(u_{i+1})}=\frac{b}{a}.
 \label{eq:codisprelaxation}
\end{equation}
Thus \(\sqrt{C_T}\), with \(C_T\) the tree's CoDisp, is the largest
relative change of the minimizing field over one release. Equivalently
\(\log(1+C_T)\) is the largest one-step decrease of the log minimum
energy. These log decreases sum to \(\log(m_\ell/m_0)\), so the maximum
measures how much of that change can occur at a single cut. The mass
chains \(1,2,4,8\) and \(1,6,7,8\) have the same total, and their largest
one-step ratios are \(1\) and \(5\). On a fixed chain, changing each
\(\log m_i\) by at most \(\eta\) moves \(\log(1+C_T)\) by at most \(2\eta\);
the bound concerns the value of the maximum and lets its location switch.

The sequence of clamps follows the exact descendant sets on the
query's path. Section~\ref{sec:collusivecuts} compares this release
with the reclamping of a class in a normalized cut.

\paragraph{Smoothed CoDisp from relative field responses.}
The response identity suggests an alternative ensemble reading while
leaving the per-tree coalition move and Hamiltonian unchanged.
Here $T$ includes the sampled records, the cuts conditional on them,
and any randomized query attachment, under the sampling law of
Section~\ref{sec:sampling}. The calibration is applied before this
ensemble average.
For an increasing invertible calibration \(F\), the score
\(F(\E_T[F^{-1}(Q_T)])\) averages in the coordinate \(F^{-1}\) and
reports the result on the scale of \(Q_T\). With \(Q_T=C_T\) and
\(F(u)=u^2\) on nonnegative \(u\), the field interpretation gives the
score
\begin{equation}
 S=\bigl(\E_T\sqrt{C_T}\bigr)^2
   =\E_T C_T-\operatorname{Var}_T(\sqrt{C_T}).
 \label{eq:codispmeanfield}
\end{equation}
Ordinary forest CoDisp averages squared relative field-change
magnitudes. Equation~\eqref{eq:codispmeanfield} squares the mean
relative magnitude, so its discount is exactly the between-realization
variance of that magnitude. Each tree has already selected its
maximizing release; the mean is over nonnegative magnitudes, not
over aligned field vectors. For instance
two trees with scores \(1\) and \(81\) give ordinary CoDisp \(41\)
and \(S=25\). A single large value has less effect in this example. The CoDisp
comparisons in Section~\ref{sec:fan} use the ordinary arithmetic
mean; \eqref{eq:codispmeanfield} specifies the alternative reporting
rule.

Write $R_T=\sqrt{C_T}$. The variance removed in
\eqref{eq:codispmeanfield} has the exact decomposition
\[
 \operatorname{Var}(R_T)
 =\operatorname{Var}_{S\sim\mathsf Q}\!\bigl(\E[R_T\mid S]\bigr)
  +\E_{S\sim\mathsf Q}\!\bigl[\operatorname{Var}(R_T\mid S)\bigr].
\]
The first term comes from sampling the records; the second includes
cut construction and query attachment with those records fixed.
Quadratic energy identifies $R_T$ as the response magnitude.
Sampling then explains part of the difference between averaging
squared magnitudes and squaring their average. If a common amplitude
$\rho$ is the target and $R_T=\rho+\varepsilon_T$ has centred
fluctuations, mean CoDisp is $\rho^2+\operatorname{Var}(\varepsilon_T)$.
For $N$ independent realizations, the expected square of the empirical
mean amplitude is instead
$\rho^2+\operatorname{Var}(\varepsilon_T)/N$.
This gives a sampling interpretation of the smoothing, conditional
on that choice of target. Variation between samples can also reflect
meaningful geometry, and the two ensemble readings retain different
amounts of it.

\paragraph{Departure or a declared mass.}
CoDisp can test the departure of a whole near-duplicate cloud when
it is the query side of a split. The declared-mass escape instead
retains the records and reads the query at an ancestor with enough
support (Section~\ref{sec:games}). The former changes a boundary;
the latter chooses where a private payoff is read. Their common
resilience to near duplicates does not identify their payoffs.

\subsection{Boundary Costs of a Fixed Partition}
\label{sec:contours}
At a fixed partition the normalized cut can be attributed exactly: to the faces of the trees, to their merges, and to the references themselves. Each attribution answers the question of who pays for a boundary in its own way, and all of them add up to the same cut. Fix a partition $A_1,\ldots,A_k$ of the routed reference set $P$, with pooled volumes $V_c=\operatorname{vol}(A_c)$, and an induced tree $t$ with affinity $K_t$. Here $S_u\subseteq P$ is the set of reference descendants of $u$, so masses in this calculation count the references. For a vertex $u$ with children $a$ and $b$ write $m_a=|S_a|$ and $n_a^c=|S_a\cap A_c|$.

\paragraph{The boundary game.}
For a descendant set $S$ let
\begin{equation}
 \nu_t(S)=\sum_{c=1}^k\frac{1}{V_c}\sum_{i\in S\cap A_c,\;j\in S\setminus A_c}K_t(i,j),
 \label{eq:boundaryworth}
\end{equation}
the normalized cut that the partition draws inside $S$ when only the tree's affinities among the members of $S$ count. It is a cost: the worth of a coalition is what its members pay for the boundaries among them.

\begin{theorem}[Boundary prices]
\label{prop:boundaryprices}
\leavevmode
\begin{enumerate}\setlength{\itemsep}{1pt}
\item[(i)] Two siblings that join at $u$ add
\begin{equation}
 \delta_u=\nu_t(S_u)-\nu_t(S_a)-\nu_t(S_b)
 =\varphi_t(u)\sum_{c=1}^k\frac{n_a^c(m_b-n_b^c)+n_b^c(m_a-n_a^c)}{V_c}\;\geq0 .
 \label{eq:boundaryprice}
\end{equation}
A leaf $\ell$ holding several references has worth $\nu_t(S_\ell)=\sum_cn_\ell^c(m_\ell-n_\ell^c)/V_c$, and
\[
 \operatorname{Ncut}(A)=\frac1N\sum_{t=1}^N\nu_t(P)
 =\frac1N\sum_{t=1}^N\Bigl[\sum_{u\text{ internal}}\delta_u^t+\sum_{\ell\text{ leaf}}\nu_t(S_\ell)\Bigr].
\]
\item[(ii)] For any $z$ on $P$, the sibling sums
\[
 \delta_u(z)=\varphi_t(u)\sum_{i\in S_a,\,j\in S_b}(z_i-z_j)^2
 =\varphi_t(u)\Bigl[m_b\sum_{S_a}z^2+m_a\sum_{S_b}z^2-2\sum_{S_a}z\sum_{S_b}z\Bigr],
\]
with the leaves' within variations $m_\ell\sum_{S_\ell}(z-\bar z_\ell)^2$, add up to $z^\top L_tz$. For a slow eigenvector $z_l$ of the pencil $(\bar L,\bar D)$ with $z_l^\top\bar Dz_l=1$, their forest average is its relaxed cut $\lambda_l$.
\end{enumerate}
Each takes one upward pass of class counts, or of sums of $z$ and $z^2$, per tree.
\end{theorem}
\begin{proof}
Two references in different children of $u$ part at $u$, with affinity $\varphi_t(u)$; pairs inside a child are counted in that child's worth. A pair crosses the boundary of $A_c$ exactly when one of its endpoints lies in $A_c$, which gives \eqref{eq:boundaryprice}, and references that share a leaf have affinity one. Telescoping from the leaves gives $\nu_t(P)=\sum_c\operatorname{cut}_t(A_c)/V_c$, and since a cut is linear in the affinity its forest average is $\operatorname{Ncut}(A)$. For (ii) expand the square; the class indicators are the case $z=\one_{A_c}$, divided by $V_c$ and summed over $c$.
\end{proof}

Equation~\eqref{eq:wardenergy} distributes the same energy over the merges with the coefficients $\deg(v)$ of a variance decomposition; the boundary game charges each pair to the split that separates it. The two differ already at a single merge. When the children of $u$ hold the classes in the same proportions $p_c$, the contrast of their means vanishes and the merge contributes nothing to \eqref{eq:wardenergy}, yet its price is $\varphi_t(u)\,m_am_b\sum_c2p_c(1-p_c)/V_c$, positive unless $S_u$ lies in one class: the split still separates pairs with different labels. By \eqref{eq:publicclamp} on the tree induced on $S$, each worth is a clamped energy of a sparse network, with the class indicators as clamps. The price $\delta_u$ is the interaction added when two siblings are joined at $u$: the cross affinities $\varphi_t(u)$ between them enter the complete affinity graph, while the class indicators and pooled volumes stay fixed. In the sparse realization the join changes the degrees, and so the conductances, throughout both subtrees. It is not a release, which enlarges the feasible set on a fixed network and can only lower a clamped minimum (Section~\ref{sec:privatecoalitions}).

\paragraph{The value members give up at a cut.}
Consider the induced-subgraph game with worth \(B_A\) for coalition
\(A\) \citep{deng1994complexity}. Each self-affinity belongs to its
own record. Each unordered pair contributes \(2\bar K_{ij}\), which
the Shapley allocation divides equally between its endpoints. Indeed,
in the random ordering that defines the allocation, either endpoint
is equally likely to arrive second and complete the pair. Thus the
share of \(i\in A\) is \(\phi_i(A)=\bar K(i,A)\), compared with
\(\phi_i(P)=\bar d_i\) in the grand coalition. In particular,
\begin{equation}
 \frac{\operatorname{cut}(A)}{V_A}
   =\frac{\sum_{i\in A}[\phi_i(P)-\phi_i(A)]}
          {\sum_{i\in A}\phi_i(P)} .
 \label{eq:cutshares}
\end{equation}
The cut term is the fraction of the members' original allocation
that they lose by forming their own coalition. Each loss is already
available from the degree and the class-affinity product. The share
identity uses the fixed pooled affinity and holds for any partition.

\paragraph{Laminar shares of the boundary.}
The boundary game extends from the descendant sets to every coalition as in \eqref{eq:laminarworth}: a coalition acts through the maximal descendant sets it contains. In the language of \citet{myerson1980conference} the descendant sets are the conferences that can meet, and the extension is the laminar case of the hypergraph communication restriction \citep{vandennouweland1992allocation}. The prices \eqref{eq:boundaryprice} are its Harsanyi dividends, so its Shapley allocation (Section~\ref{sec:participation}) charges $i$ the share $\sigma(i)=N^{-1}\sum_t\sum_{u}\delta_u^t/m_u$, over the vertices $u$ above $i$ and its leaf, with the leaf's worth in place of $\delta$ there. The shares sum to $\operatorname{Ncut}(A)$. The two allocations answer who pays differently. The pairwise rule charges each crossing pair to its two endpoints, as the induced-subgraph game above charges the affinity of a pair. The laminar rule shares the cost of the pair over the smallest descendant block that contains both, so a reference deep inside a group shares the price of the group's boundary: collusion, restricted to siblings.

\paragraph{Contours.}
Each split $u$ of each tree is an oriented face $F_u^t$: the cut on its coordinate at its threshold, with the two siblings on its sides. For routed references the faithful extent of the face is the box of the references routed to $u$, since the tree's box comes from its own sample. The boundary prices make the faces a measure,
\begin{equation}
 \Gamma_A=\frac1N\sum_{t=1}^N\sum_{u\text{ internal}}\delta_u^t\,\mathsf u_{F_u^t},
 \label{eq:contourmeasure}
\end{equation}
with $\mathsf u_F$ the uniform probability on the face $F$. Its mass is $\operatorname{Ncut}(A)$ less the price of references that share a leaf, which no face separates. Its density in space locates where the partition pays for its boundaries; each face records which coordinate separates the two sides, and the depth of $u$ records the scale. With $z_l$ in place of the class indicators and weight $1/\lambda_l$, each slow mode contributes a contour of unit mass, less its leaf terms. This is the forest's counterpart of the spectral boundary of \citet{arbelaez2011contour}, which differentiates the slow eigenvectors of a normalized cut, weights them by $1/\sqrt{\lambda_l}$, and orders the resulting regions into an ultrametric contour map. Here each tree is already an ultrametric hierarchy, with box ranges as its heights. The boundary game puts the partition's cut, or a slow mode's relaxed cut, on its faces, and the forest averages $N$ such hierarchies instead of selecting one.

\paragraph{Exact attribution on approximately nested trees.}
Each tree is exactly nested, so on each tree the prices are defined, nonnegative, and add up to that tree's reading of the cut. The forest is only approximately nested: its trees hold different samples, and the pooled affinity $\bar K$ belongs to no single hierarchy. The attribution is exact nonetheless. A cut is linear in the affinity and the worths share the pooled volumes, so the forest average of the per-tree attributions, to the faces and leaves in \eqref{eq:contourmeasure} or to the members in $\sigma$, is exactly $\operatorname{Ncut}(A)$. The approximation shows only in where the trees place the price, on which faces and at what depth, never as an error in the total. The spread of priced faces across trees along one stretch of a boundary is the boundary's counterpart of the swing of a query (Section~\ref{sec:faces}): small where the trees agree on where the partition pays, and large where each tree draws the boundary through its own boxes.

\subsection{Collusive Cuts}
\label{collusivecuts}\label{sec:collusivecuts}
In this subsection we use the collusion and coalition dynamics discussed in the context of subtree restricted games and apply them in the context of normalized cuts. Unlike the attributions of Section~\ref{sec:contours}, which hold the partition fixed, a move changes the partition and with it the class volumes.

\paragraph{Evaluating a relocation.}
Write \(W(C,A)=\sum_{i\in C,j\in A}\bar K_{ij}\),
\(B_A=W(A,A)\), and \(V_A=\operatorname{vol}(A)\).

\begin{proposition}[Exact evaluation and relocation]
\label{prop:ncutdescent}
For a partition into \(k\) classes, the cuts, volumes, and
class-affinity products \(\bar K(i,A_c)\) take \(O(Nk|P|)\) work
by tree passes on the induced trees. Moving a nonempty group
\(C\subsetneq A\) from class \(A\) to class \(B\) changes the
normalized cut by
\begin{equation}
 \frac{B_A}{V_A}+\frac{B_B}{V_B}
 -\frac{B_A-2W(C,A)+B_C}{V_A-V_C}
 -\frac{B_B+2W(C,B)+B_C}{V_B+V_C}.
 \label{eq:cutmove}
\end{equation}
\end{proposition}
\begin{proof}
Apply Proposition~\ref{prop:nested} to the class counts and
indicator vectors on each tree. Removing \(C\) subtracts the rows
and columns of \(C\) from \(B_A\), adding back \(B_C\) once.
Adding \(C\) to \(B\) contributes both cross terms and \(B_C\).
The degrees of \(C\) transfer between the two volumes. Substitute
these changes into \(k-\sum_c B_{A_c}/V_{A_c}\).
\end{proof}

For a singleton $C=\{i\}$, $B_C=1$ and $V_C=\bar d_i$.

\paragraph{A potential game of relocation.}
Potential games for normalized partition objectives have been studied
in \citet{avrachenkov2018network}. For the full-degree normalization
of \eqref{eq:ncut}, let \(\Phi(A)=B_A/V_A\) on nonempty classes and
\(\Phi(\emptyset)=0\). Give record \(i\in A\) the payoff
\(u_i(A)=\Phi(A)-\Phi(A\setminus\{i\})\). A unilateral change between
classes, keeping all classes nonempty, changes this payoff by exactly
the change in the partition potential \(\sum_A\Phi(A)=k-\operatorname{Ncut}\).
For a joint move the analogous gain is the group's marginal value
in the destination less its marginal value in the source. This is
the payoff for changing membership. Equation~\eqref{eq:cutshares}
instead allocates affinity at a given partition.

\paragraph{A self-referential cut.}
To test a trial class for the query, let its row join the graph and prescribe that class. With $k_c=\bar K(q,A_c)$ the total affinity to class $c$, and $d_q=1+\sum_c k_c$, the row adds $k_{c'}$ to each old volume $\operatorname{vol}(A_{c'})$. Joining class $c$ adds $d_q$ more to $\operatorname{vol}(A_c)$ and $2k_c+1$ to its within-class affinity. The normalized cut for each trial follows from the $k$ sums $k_c$, one path per tree; the smallest chooses the class. The passive probe had support $\sum_c k_c$; the self-referential test gives the query a degree including its unit diagonal, and changes the volumes it was previously read against. The stored tree is fixed. Actual sample insertion can also change the old affinities or the routing (Section~\ref{sec:sampleinsertion}).

\paragraph{A joint move when every singleton move fails.}
Multilevel graph clustering uses block moves on coarsened graphs
\citep{dhillon2005multilevel}. Here the range trees supply the candidate
blocks: on one tree, every descendant set contained in a single current
class can move together to any other class, and
Theorem~\ref{prop:collusivecut} below scans all of them in one pass.
The example and the proposition that follow show why the blocks matter.

For an example, use the points \(0,\ldots,5\) on the line with
successive pairs \(P_1=\{0,1\}\), \(P_2=\{2,3\}\), and
\(P_3=\{4,5\}\). The root splits off \(P_3\) and its left child
separates \(P_1\) from \(P_2\). With \(h=1/\log 2\), the
affinities within a pair are \(1/2\) off the diagonal, between
\(P_1,P_2\) they are \(1/8\), and to \(P_3\) they are \(1/32\).
The partition \((P_1\cup P_3,P_2)\) has normalized cut
\(84/319\approx 0.263\). The best single-record move raises it
to \(280/783\approx 0.358\). Moving the whole branch \(P_1\)
to \(P_2\) lowers it to \(42/377\approx 0.111\).

\begin{theorem}[An unbounded gap for small moves]
\label{prop:ncutcoalitiongap}
For every \(s\ge 2\), there are one-dimensional range trees on
\(3s\) distinct points with a two-class partition whose normalized
cut can exceed optimal by an arbitrarily large factor. Every
nonempty relabeling of fewer than \(s\) points increases its cut.
Moving one descendant
set of size \(s\) attains the optimum and strictly improves
\(u_i\) for every record that moves.
\end{theorem}
\begin{proof}
Take three translated copies \(P_1,P_2,P_3\) of \(s\) points in
\([0,1]\) including both endpoints, with shifts \(0,R-1,2R-2\)
for \(R\ge 3\). The root splits off \(P_3\), its other child
separates \(P_1\) and \(P_2\), and all three copies use the same
internal splitting topology. With \(h=1/\log 2\) and \(\epsilon=2^{-R}\),
within-copy affinities are at least \(1/2\), those between \(P_1\)
and \(P_2\) are \(\epsilon\), and those to \(P_3\) are
\(2\epsilon^2\). The six-point example uses \(s=2,R=3\).

Any bipartition that splits a copy has crossing affinity at least
\((s-1)/2\). Its normalized cut is at least \(2(s-1)/(9s^2)\)
since \(V_A+V_B\le 9s^2\) and \(1/V_A+1/V_B\ge 4/(V_A+V_B)\).
The partition \((P_1\cup P_3,P_2)\) and the root partition
\((P_3,P_1\cup P_2)\) both have costs tending to zero as
\(R\to\infty\). Their crossing affinities are
\(s^2(\epsilon+2\epsilon^2)\) and \(4s^2\epsilon^2\), respectively.
If \(M\) is the common within-copy affinity, their reciprocal-volume
sums both tend to \(3/(2M)\). Their cost ratio is therefore
asymptotic to \(1/(4\epsilon)\). The third whole-copy partition has
the same cost as the starting one by symmetry of \(P_1,P_2\).
Thus for large \(R\) the root partition is optimal. Every nonempty
relabeling of fewer than \(s\) points splits a copy, while moving
\(P_1\) to \(P_2\) gives the root partition.

For the payoff comparison, let \(A=P_1\cup P_3\),
\(A'=P_1\cup P_2\), \(i\in P_1\), \(V=V_A\), \(d=\bar d_i\),
and \(\eta=s^2(\epsilon-2\epsilon^2)\). The volume and within-class
affinity increase by \(\eta\) and \(2\eta\) from \(A\) to \(A'\).
With \(i\) excluded the increases are \(\eta\) and
\(2(1-1/s)\eta\). Taking the difference of the two marginal
payoffs gives
\[
 \frac{u_i(A')-u_i(A)}{\eta}
 =\frac{2-\Phi(A)}{V+\eta}
 -\frac{2(1-1/s)-\Phi(A\setminus\{i\})}{V-d+\eta}.
\]
As \(R\to\infty\), write \(d_0=\sum_{j\in P_1}K_{ij}\le s\).
Then \(V\to 2M\), \(d\to d_0\), \(\Phi(A)\to 1\), and
\(\Phi(A\setminus\{i\})\to 1-(d_0-1)/(2M-d_0)\).
Multiplying the limiting expression by \(2M-d_0\) gives
\[
 \frac{2}{s}-\frac{d_0}{2M}-\frac{d_0-1}{2M-d_0}>0
\]
because \(2M\ge s(s+1)\) and each subtracted term is less
than \(1/s\). Hence for large \(R\) every moving record gains.
\end{proof}

\paragraph{Moves as reclamping.}
For a class \(A\) on the affinity graph, ground \(A^c\) and then
clamp all of \(A\) at one. The initial field is zero, and
\eqref{eq:groupclamp} gives twice the energy increase as
\(\one^\top(G_{AA})^{-1}\one=\operatorname{cut}(A)\).
Dividing by \(V_A\) gives the class's normalized boundary cost.
A move of a block \(C\) from \(A\) to \(B\) reclamps it: in the field of
\(A\) the block changes from one to zero, in the field of \(B\) from zero
to one, and both class volumes are recomputed. The renormalization is
why relocation is not a release and need not lower any one energy.
The move in Theorem~\ref{prop:ncutcoalitiongap} preserves the
branch's internal affinity while exchanging a weak connection for
a stronger one. It lowers the cut and benefits each mover even
though each would lose by moving alone. The tree specifies which
records can act together, and the two energies measure the
consequences of that action.

Fix the partition, the induced tree, and the notation of Section~\ref{sec:contours}.

\paragraph{The deviation game and the collusive cut.}
A descendant set $C$ is \emph{pure} if it lies inside one class $A$ and is not all of it. Tree $t$ reads the forest's partition with its own affinities and the pooled volumes,
\[
 \operatorname{Ncut}_t(A_1,\ldots,A_k)=\nu_t(P)=\sum_{c=1}^k\frac{\operatorname{cut}_t(A_c)}{V_c},
\]
the worth \eqref{eq:boundaryworth} of the whole reference set. A cut is linear in the affinity, so for every partition the forest average of $\operatorname{Ncut}_t$ is $\operatorname{Ncut}$. The gain of $C$ from changing class together is
\[
 g_t(C)=\max_{B\neq A}\bigl[\operatorname{Ncut}_t(A_1,\ldots,A_k)-\operatorname{Ncut}_t(\text{the partition with $C$ moved from $A$ to $B$})\bigr],
\]
in which the moved partition has the pooled volumes $V_A-V_C$ and $V_B+V_C$ and the cuts
\[
 \operatorname{cut}_t(A\setminus C)=\operatorname{cut}_t(A)-\operatorname{cut}_t(C)+2W_t(C,A\setminus C),\qquad
 \operatorname{cut}_t(B\cup C)=\operatorname{cut}_t(B)+\operatorname{cut}_t(C)-2W_t(C,B),
\]
with $W_t$ the cross affinity under $K_t$. Collusive displacement takes, over the descendant sets on a query's path, the largest displacement per member \citep{guha2016rrcf}. In the same way put
\begin{equation}
 \operatorname{CC}_t(i)=\max\Bigl\{0,\ \max_{C}\frac{g_t(C)}{|C|}\Bigr\},\qquad
 \operatorname{CC}(i)=\frac1N\sum_{t=1}^N\operatorname{CC}_t(i),
 \label{eq:collusivecut}
\end{equation}
the inner maximum over the pure descendant sets on the path of $i$ in tree $t$, the singleton $\{i\}$ included when it is pure. The zero is the option to stay put. It is also the value for the only member of a singleton class, which cannot move without emptying its class.

\begin{theorem}[The collusive cut]
\label{prop:collusivecut}
\leavevmode
\begin{enumerate}\setlength{\itemsep}{1pt}
\item[(i)] On an induced tree, the class products $K_t(i,A_c)$ summed upward give $W_t(S_u,A_c)$; the within-coalition affinities satisfy $B_t(S_u)=B_t(S_a)+B_t(S_b)+2\varphi_t(u)m_am_b$, with $B_t(S_\ell)=m_\ell^2$ at a leaf; and the tree's degrees and the pooled degrees, summed upward, give $\operatorname{vol}_t(S_u)$ and $V(S_u)$, hence $\operatorname{cut}_t(S_u)=\operatorname{vol}_t(S_u)-B_t(S_u)$. Hence $\operatorname{CC}_t$ of every reference takes $O(k|P|)$ work.
\item[(ii)] $\operatorname{CC}_t(i)$ is at least the gain per member of every pure descendant set on the path of $i$, and in particular of the best single move of $i$. If $\operatorname{CC}_t(i)=0$ for every reference, no descendant set of tree $t$ gains by changing class together. Stability under single moves does not imply this, and the gap in the objective can be unbounded (Theorem~\ref{prop:ncutcoalitiongap}).
\item[(iii)] For one block $C$ and one destination $B$, the forest average of the per-tree changes of $\operatorname{Ncut}_t$ is the pooled change \eqref{eq:cutmove}. A block proposed by one tree is approved by testing it with \eqref{eq:cutmove}, which takes one more pooled product, for $B_C=\one_C^\top\bar K\one_C$.
\end{enumerate}
\end{theorem}
\begin{proof}
(i) Proposition~\ref{prop:nested} gives the class products in one pass, and cross pairs part at $u$, which gives the recursion; $W_t(C,A\setminus C)=W_t(C,A)-B_t(C)$. The cuts of the moved partition then cost $O(1)$ per vertex and destination, hence $O(k)$ per vertex. (ii) Every pure descendant set on the path of $i$ is a candidate, and every pure descendant set is on the path of each of its members. Theorem~\ref{prop:ncutcoalitiongap} gives partitions in which every move of fewer than $s$ records loses while a descendant set of size $s$ reaches the optimum. (iii) For any set $S$ the forest average of $\operatorname{cut}_t(S)$ is the pooled cut, and the volumes are common to the trees, so the forest average of $\operatorname{Ncut}_t$ is $\operatorname{Ncut}$ for the partitions before and after the move.
\end{proof}

As with collusive displacement, the maximum over nested coalitions answers masking: a group of near duplicates, none of which gains by moving alone, is tested as the descendant set that holds them. The collusive cut is quenched: each tree chooses its own block and destination and reads them with its own affinities, against the pooled volumes. By (iii) the per-tree readings of one common move average exactly to its pooled gain, so the per-tree diagnostic and the pooled approval differ only in which move they test, not in how they normalize it. A positive pooled gain for a candidate means that the forest would redraw that boundary too. A positive collusive cut without one marks references where single trees would redraw the boundary and the forest would not, much as the Kriging variance separates missing support from the incompatibility of the trees' reconstructions \eqref{eq:krigingsupport}. The gains of pure descendant sets also define a laminar participation game (Theorem~\ref{prop:laminarparticipation}), whose fold \eqref{eq:clampfold} proposes disjoint blocks to move. Its extension adds the gains of those blocks and so ignores their interaction through shared volumes and cross affinities. The blocks are therefore only proposals: \eqref{eq:cutmove} tests one at a time, and after each accepted move the others are rescored against the updated partition.

\section{Directional Audits and Perturbations}
\label{sec:audits}
An answer can draw on different geometric directions, and a weak
connection can make it sensitive to small changes. We want to know
which directions supply its evidence and how strongly they hold
the answer. Signed route lengths record where the evidence lies.
Changing a directional scale tests its effect on the prediction
after minimizing the updated Hamiltonian over the unknown values;
prescribing another answer instead measures its energy cost.

We specify the directional units, then compute predictions together
with their route evidence and audit stiffness. Differentiating the
sparse realization connects geometric moments to the response of
boundary energy. The stability theorem bounds finite changes in
the resulting harmonic predictions.

\subsection{Directional Lengths and Gauges}
\label{sec:gaugemass}
A gauge specifies the units in which the retained directional lengths
are compared. It consists of positive scales on the signed half
dimensions of Section~\ref{sec:halfdimensions}. Dividing each length
by its scale gives a dimensionless path length and hence an affinity.
The same reference set $P$ and the same pooling operation of
Section~\ref{sec:routed} are used after these scales are specified.

\paragraph{Lengths in local units.}
For directional interpretation, retain nonnegative signed length witnesses $\ell_{e,j}^{\pm}(x)$ with sum $\ell_e$, recording the extent or gap assigned to each half dimension. Parent-box half extents, with declared positive floors, supply local scales $s_{e,j}^{\pm}$ that can change with resolution along the traversal. With $2d$ independent positive multipliers, set
\begin{equation}
 h_{e,j}^{\pm}=\lambda_j^{\pm}(x)s_{e,j}^{\pm},\qquad \gamma_{e,j}^{\pm}=\ell_{e,j}^{\pm}(x)/h_{e,j}^{\pm},\qquad \Gamma_e=\sum_{j,\pm}\gamma_{e,j}^{\pm}.
 \label{eq:passagegauge}
\end{equation}
Fix these annotations for the query before folding. The dimensionless lengths $\Gamma_e$ define an additive tree metric $d_\Gamma$; replacing $\ell_e/h$ by $\Gamma_e$ gives $\rho_e=e^{-\Gamma_e}$ and $K_{ij}=e^{-d_\Gamma(i,j)}$. Theorem~\ref{prop:laplacerealization} applies to this symmetric metric with all $2d$ multipliers retained. Equal scales recover $\Gamma_e=\ell_e/h$. For edge-dependent scales the general path-product construction of Theorem~\ref{prop:laplacerealization} and the two passes \eqref{eq:kernelmassfold} apply directly.

The affine comparison also quantifies a bounded change of these
units. On the stored observations of
Theorem~\ref{thm:geometryenergy}, suppose
$0<h_-\leq h_{e,j}^{\pm}\leq h_+$ uniformly over trees, edges,
and faces. Nonnegativity of the witnesses and their sum $\ell_e$
give $d_T/h_+\leq d_\Gamma\leq d_T/h_-$.
Let $\mathcal H_\Gamma^T$ use affinities $e^{-d_\Gamma}$.
Theorem~\ref{thm:geometryenergy}, applied at $h_-$ and $h_+$,
then gives
\begin{equation}
 \mathcal H^{(1)}_{h_-/a}(f)
 \leq\E_T\mathcal H_\Gamma^T(f)
 \leq\mathcal H^{(1)}_{h_+}(f).
 \label{eq:affinegaugeenergy}
\end{equation}
The noisy version replaces $a$ by $a_W$ and multiplies the lower
bound by $e^{-b_W/h_-}$. The comparison requires the stated uniform
bounds on the gauges; the exact folds and directional derivatives
below use the declared positive scales directly.

\paragraph{An edge decomposition from nested boxes.}
For a complete query $x$, resolve the occupied projection of $B_v$ on each side of $x$:
\[
 o_{vj}^-=[\min(u_{vj},x_j)-\ell_{vj}]_+,\qquad o_{vj}^+=[u_{vj}-\max(\ell_{vj},x_j)]_+.
\]
Their sum is $u_{vj}-\ell_{vj}$. If $B_v\subseteq B_p$, each occupied half width is no larger than its parent's. Set
\[
 \ell_{\{p,v\},j}^{\pm}(x)
   =\tfrac12(o_{pj}^{\pm}-o_{vj}^{\pm}).
\]
These entries are nonnegative and sum to $H_p-H_v$, where $H_v=R(B_v)/2$. In a tree of nonempty axis-aligned splits, the cut-coordinate width strictly decreases in each child, so $H_p-H_v>0$. With singleton-location leaves, $H_v=0$ at the leaves and the induced tree distance between observations is $R(B_{\operatorname{lca}(i,j)})$, the range-based tree distance used in \citet{guha2016rrcf}. Thus a maintained box tree supplies a concrete $2d$ decomposition without changing its cuts. Directional empty intervals between sibling boxes can be recorded separately within these length witnesses. The decomposition changes with $x$, while its unscaled total edge length does not.

\paragraph{Directional scales and operator response.}
With the tree, signed witnesses, and base scales fixed, let $\boldsymbol r_{ij}$ collect normalized path lengths in the $2d$ half dimensions and let $\theta_\alpha=1/\lambda_\alpha$ be their inverse scale multipliers. The gauge \eqref{eq:passagegauge} gives $K_{ij}=e^{-\boldsymbol\theta^\top\boldsymbol r_{ij}}$ and $L_K=\operatorname{diag}(K\one)-K$. Writing $\operatorname{Lap}(W)=\operatorname{diag}(W\one)-W$, its directional response is
\begin{equation}
 -\partial_{\theta_\alpha}L_K
   =\operatorname{Lap}\bigl((r_{ij,\alpha}K_{ij})_{ij}\bigr)\succeq0.
 \label{eq:scalelaplacian}
\end{equation}
Each pair contributes a positive exponential coefficient to $L_K$, making it convex in the matrix order as a function of the inverse scales; its energy at fixed potentials is convex as well. Section~\ref{app:boundaryaudit} differentiates the conductances of the sparse network instead, edge by edge.

\paragraph{Reference support and direction.}
\label{app:laplacecounts}
Write \(\boldsymbol r_{qj}\) for the sum of the signed witness-to-scale
vectors along the route from \(q\) to \(j\) before multiplication by
the inverse scales \(\boldsymbol\theta\). The reference support and its weighted route moments are
\begin{equation}
 b_q(\boldsymbol\theta)
 =\sum_{j\in P}e^{-\boldsymbol\theta^\top\boldsymbol r_{qj}}.
 \label{eq:masslaplace}
\end{equation}
The negative gradient of \(\log b_q\) is the corresponding
affinity-weighted mean route vector. The mass passes of
\eqref{eq:kernelmassfold} carry these first moments by adding the edge
vector times the mass being passed.

\subsection{Predictions and Their Directional Evidence}
\label{sec:colors}\label{sec:colorfold}\label{sec:colorprobes}
\label{app:directionalpassage}
\paragraph{Harmonic prediction.}
On the sparse affinity network of Theorem~\ref{prop:laplacerealization},
the next observation terminal visited from \(i'\), including a return
to \(i'\), is \(j'\) with probability \(K_{ij}/b_i\), where
\(b_i=\sum_jK_{ij}\). Indeed the pendant conductance is \(b_i\), the
trace conductance to \(j'\ne i'\) is \(K_{ij}\), and the remaining
probability is \(1/b_i\) \citep{lyons2016probability}. Thus a harmonic
prediction with colored observations as boundary values can be evaluated
on the sparse tree \citep{zhu2003semisupervised}. For a passive probe it is
\(f(q)=\sum_{j\in P}K_{qj}f_j/b_q\), with \(b_q=\sum_{j\in P}K_{qj}\)
and \(f_j\) the solved reference values. The support \(b_q\) distinguishes
a strong reference connection from a well-defined voltage in a void.

\paragraph{Route evidence of a prediction.}
Let \(\omega_q(a')\) be the first-hit probability at colored terminal
\(a'\), let \(y_{a,z}\) be its weight for color \(z\), and let
\(\boldsymbol B(q,a')\) be the sum of the signed dimensionless lengths along the unique
route. The value and its route evidence are
\begin{equation}
 f_z(q)=\sum_a\omega_q(a')y_{a,z},\qquad
 \boldsymbol J_z(q)=\sum_a\omega_q(a')y_{a,z}\boldsymbol B(q,a').
 \label{eq:colorlengths}
\end{equation}
For \(f_z(q)>0\), \(\boldsymbol J_z/f_z\) is the mean route length
conditional on that color. A branch response \(a u-\boldsymbol b\),
with unnormalized route moment \(\boldsymbol M_z\), extends through an
edge of conductance \(c\) and length vector \(\boldsymbol\gamma_e\) as
\[
 (a',\boldsymbol b')=\frac{c}{c+a}(a,\boldsymbol b),\qquad
 \boldsymbol M'_z=\frac{c}{c+a}
       (\boldsymbol M_z+\boldsymbol\gamma_e b_z).
\]
The series factor multiplies every first-hit coefficient in that
branch, and the new edge adds its length vector to every route. Parallel
branches add their responses and moments. Dividing the resulting
\((b_z,\boldsymbol M_z)\) by \(a\) gives \eqref{eq:colorlengths}.
This evaluates the predictions and their route evidence together.
If the forest prediction averages passive voltages with support
weights \(b_q\), its route length vector for color \(z\) uses the same weights,
\(\E[b_q\boldsymbol J_z(q)]/\E[b_q f_z(q)]\), when the denominator
is positive.
Changing the scale of a route moves a prediction through its voltage
drops and the Green's response to that change. For example, constant
boundary values give zero response to a scale change even when the
route lengths are large. Section~\ref{app:boundaryaudit} gives the
directional derivatives.

\paragraph{Auditing a value.}
\label{sec:probeaudit}
Prescribe \(z\) at a free probe \(q\) and let the other free voltages
minimize. If \(F\) is the free set and \(G=(L_{FF})^{-1}\), the cost is
\begin{equation}
 \mathcal H_q(z)-\mathcal H_q(f(q))
  =\tfrac12 a_q\|z-f(q)\|_2^2,\qquad a_q=1/G(q,q).
 \label{eq:probeaudit}
\end{equation}
The residual current is \(a_q(z-f(q))\). This follows by eliminating
the other free voltages, leaving one scalar stiffness per component.
For a regression audit the boundary values and the probe value are
scalars. The support and the audit stiffness report separate aspects
of the local geometry.

\subsection{Response to a Change of Scale}
\label{app:boundaryaudit}\label{sec:boundaryaudit}
The route moments describe the geometry used by a prediction.
To ask how changing that geometry changes the answer, we also need
the voltage drops of the field. Fix the hierarchy, references,
signed witnesses, and base scales throughout the differentiation.

\paragraph{Boundary-energy response.}
Partition the sparse network into retained boundary $B$ and
interior $I$, with invertible $L_{II}$. Its response is
$\Lambda_B=L_{BB}-L_{BI}L_{II}^{-1}L_{IB}$.
For boundary potentials $y$ and harmonic extension $v_y$,
differentiating the minimized energy gives
\begin{equation}
 \partial_{\theta_\alpha}(y^\top\Lambda_B y)
 =\sum_{e=\{u,v\}}\partial_{\theta_\alpha}c_e
                    (v_y(u)-v_y(v))^2.
 \label{eq:gaugeresponse}
\end{equation}
The interior derivative vanishes by harmonicity. Finite single-edge
updates follow from the standard rank-one Kron sensitivity formula
\citep{dorfler2013kron}. A geometric scale change can alter many
conductances, including the pendant edges, so all their terms enter
\eqref{eq:gaugeresponse}.

\paragraph{From directional mass to conductance.}
Write $u_{e,\alpha}=\ell_{e,\alpha}/s_{e,\alpha}$ and
$\rho_e=e^{-\boldsymbol\theta^\top\boldsymbol u_e}$.
For a tree vertex $v$, its normalized directional mass moment is
\[
 \bar r_{v,\alpha}
 =\frac{\sum_{j\in P}K_{vj}r_{vj,\alpha}}{b_v}
 =-\partial_{\theta_\alpha}\log b_v.
\]

\begin{proposition}[Directional response of the sparse realization]
\label{prop:directionalaudit}
For the conductances of Theorem~\ref{prop:laplacerealization},
\begin{equation}
 -\partial_{\theta_\alpha}\log c_{uv}
 =\bar r_{u,\alpha}+\bar r_{v,\alpha}
  +\frac{1+\rho_{uv}^2}{1-\rho_{uv}^2}u_{\{u,v\},\alpha}.
 \label{eq:massconductancesensitivity}
\end{equation}
A pendant has $c_{j,j'}=b_j$ and
$\partial_{\theta_\alpha}\log c_{j,j'}=-\bar r_{j,\alpha}$.
For fixed boundary-value dimension, the $2d$ directional derivatives
of boundary energy in \eqref{eq:gaugeresponse} take $O(d|W|)$ work
on the sparse tree.
\end{proposition}
\begin{proof}
Differentiate
$\log c_{uv}=\log b_u+\log b_v+\log\rho_{uv}
-\log(1-\rho_{uv}^2)$ to obtain
\eqref{eq:massconductancesensitivity}. The mass passes of
\eqref{eq:kernelmassfold} accumulate their directional moments by
adding an edge vector times the mass passed across it. Tree
elimination supplies the voltages; summing the edge contributions
in \eqref{eq:gaugeresponse} proves the work bound.
\end{proof}

\paragraph{A void on one edge.}
\label{app:gaprecovery}
For an edge $e=\{u,v\}$ separating $V_L,V_R$, let $\Gamma_e$ be its dimensionless length. Every cross-terminal path passes through that edge. Its affinity factors as
\[
 K_{xy}=e^{-d_\Gamma(x,u)}\,\rho\,e^{-d_\Gamma(v,y)},
 \qquad \rho=e^{-\Gamma_e},
\]
so summing the cross affinities with
\[
 Z_L=\sum_{x\in V_L}e^{-d_\Gamma(x,u)},\qquad
 Z_R=\sum_{y\in V_R}e^{-d_\Gamma(v,y)}
\]
gives the total cross-cut affinity
\begin{equation}
 C_K(V_L)=e^{-\Gamma_e}Z_LZ_R.
 \label{eq:recoveredgap}
\end{equation} A geometric gap of length $G$ at common scale $h$ is the special case $\Gamma_e=G/h$.

The cross-cut affinity is also the current when the groups are clamped at zero and one, preserved by the sparse realization. Holding $Z_L,Z_R$ fixed, an increase $\Delta\Gamma_e$ reduces this coupling by $e^{-\Delta\Gamma_e}$. Because $\Gamma_e$ retains all $2d$ signed length-to-scale contributions, the weak connection can be traced to a gap and the local scale at which it matters. These same contributions appear in the color-conditioned routes of \eqref{eq:colorlengths}.

The exact sparse realization makes this barrier visible on one electrical edge. On the two sides of the gap, \eqref{eq:treelaplace} gives $b_u=Z_L+\rho Z_R$ and $b_v=Z_R+\rho Z_L$. Therefore
\begin{equation}
 c_{uv}=\frac{\rho(Z_L+\rho Z_R)(Z_R+\rho Z_L)}{1-\rho^2},
 \qquad
 \frac{c_{uv}}{\rho Z_LZ_R}\longrightarrow1
 \label{eq:sparsegap}
\end{equation}
as $\Gamma_e\to\infty$ with fixed positive $Z_L,Z_R$. This edge retains the leading exponential attenuation of the complete graph's cross-cut capacity. Exact effective capacity between terminal sets also includes any resistance within the sides; it is preserved by the network elimination of Theorem~\ref{prop:laplacerealization}.

\subsection{Stability under Conductance Perturbations}
\label{sec:stability}\label{app:recoverynoise}
\label{sec:noiserecovery}\label{app:metricresponse}
Keep the edge support and supplied boundary values fixed. We first
bound the change in the whole harmonic field, then in its hitting
and directional evidence. These are perturbations of a specified
Hamiltonian. The coordinate-noise result of
Corollary~\ref{cor:noisygeometry} controls the geometry from which
the trees are built, and Theorem~\ref{thm:geometryenergy} brackets
their averaged energy. The following bounds compare two specified
operators through relative errors in their conductances.

\begin{theorem}[Stability under relative conductance errors]
\label{prop:energystability}
Let a finite connected network have positive edge conductances and a nonempty fixed boundary. Compare conductances $c,c'>0$ on the same edge support, with the same prescribed boundary values. Write $\kappa_-=\min_ec'_e/c_e>0$, $\kappa_+=\max_ec'_e/c_e$ and $\eta=\tfrac12\log(\kappa_+/\kappa_-)$, which is zero for a common rescaling of all conductances. If $u$ and $u'$ minimize the old and the new energies, then at every interior vertex $q$
\[
 \|u'-u\|_L\leq\sinh(\eta)\,\|u\|_L,\qquad
 |u'_q-u_q|\leq\sinh(\eta)\sqrt{(L_{II}^{-1})_{qq}}\,\|u\|_L,
\]
where $\|u\|_L^2=u^\top Lu$ for the old Laplacian $L$, with interior block $L_{II}$. The relative bound is attained on a path of two edges.
\end{theorem}
\begin{proof}
Put $w=u'-u$, $k=(\kappa_++\kappa_-)/2$ and $B_0=L'/k-L$. Edgewise, the weights of $B_0$ are at most $\tanh(\eta)$ times those of $L$ in absolute value, so Cauchy--Schwarz gives $|w^\top B_0u'|\leq\tanh(\eta)\|w\|_L\|u'\|_L$. Harmonicity implies $w^\top Lu=0$ and $w^\top L'u'=0$, so $\|w\|_L^2=-w^\top B_0u'$ and $\|u'\|_L^2=\|u\|_L^2+\|w\|_L^2$. Solving the resulting inequality gives the first bound. The second follows from Cauchy--Schwarz on the grounded block applied to $w_q=e_q^\top w_I$. For the path, take resistances $e^{2\eta}$ and $1$ in series between boundary values zero and one, and multiply the two conductances by $e^{\eta}$ and $e^{-\eta}$. The divider voltage becomes $1/2$, and substitution gives $\|w\|_L/\|u\|_L=\sinh(\eta)$.
\end{proof}
The factor $\sqrt{(L_{II}^{-1})_{qq}}$ is the square root of the effective resistance from $q$ to the boundary, the Kriging standard deviation of Section~\ref{sec:green} when the boundary holds the references: the sensitivity bound is larger for a weakly supported probe.

\begin{corollary}[Stability of hitting and directional evidence]
\label{cor:evidencestability}
On a fixed connected network, suppose
$|\log(c'_e/c_e)|\leq\eta$ and let $m$ be the number of free
vertices. The first-hit laws on the same nonempty absorbing
boundary satisfy
\begin{equation}
 \TV(p'_q,p_q)\leq\tanh(m\eta/2).
 \label{eq:recoverynoise}
\end{equation}
On a tree, use the same nonnegative boundary color weights in both
models. If reference route lengths in half dimension $\alpha$
lie in $[0,B_\alpha^{\max}]$ and their changes are at most
$\delta_\alpha$, the color-conditioned route moments satisfy
\begin{equation}
 \left|\frac{J'_{z,\alpha}(q)}{f'_z(q)}
       -\frac{J_{z,\alpha}(q)}{f_z(q)}\right|
 \leq\delta_\alpha+B_\alpha^{\max}\tanh(m\eta/2),
 \label{eq:directionalrecoverynoise}
\end{equation}
whenever the conditioning color has positive weight.
\end{corollary}
\begin{proof}
The harmonic forest formula represents the first-hit law by
absorbing-root spanning forests, each with $m$ edges
\citep{pitman2018tree}. Their weight ratios lie between
$e^{-m\eta}$ and $e^{m\eta}$. A probability law tilted by a weight
in $[a,b]$ changes in total variation by at most
$(\sqrt b-\sqrt a)/(\sqrt b+\sqrt a)$, which proves
\eqref{eq:recoverynoise}. Multiplying each forest weight by the
same nonnegative endpoint color preserves these ratio bounds.
Changing the route lengths costs at most $\delta_\alpha$;
changing their conditional law costs at most
$B_\alpha^{\max}\tanh(m\eta/2)$.
\end{proof}

On the complete affinity graph, a uniform distance error $\delta$
at fixed $h$ gives log-conductance error at most $\delta/h$.
For weighted $\ell_1$ distances, moving each endpoint by at most
$\sigma_j$ in coordinate $j$ gives
$\delta\leq2\sum_j\beta_j\sigma_j$.
There is a corresponding bound for the range tree when descendant
memberships remain fixed. Each box width changes by at most
$2\sigma_j$, so its range, and hence every stored-pair tree distance,
changes by at most $2W$, where $W=\sum_j\sigma_j$.
At fixed $h$ this gives
$|\log(K'_{ij}/K_{ij})|\leq2W/h$.
Theorem~\ref{prop:energystability} therefore applies to the complete
observation-affinity graph with $\eta\leq2W/h$.
Under log-conductance error $\eta$, every fixed cut capacity and
volume changes by a factor in $[e^{-\eta},e^\eta]$, so its normalized
cut changes by a factor in $[e^{-2\eta},e^{2\eta}]$.

These bounds condition on common support and boundaries. A stream
can also change routes or topology. Section~\ref{sec:referenceupdates}
then tests a reused field by its residual against the current
operator; the fixed-support bounds do not supply that correspondence.

\section{Measurements and Illustration}
\label{sec:measurements}
The preceding sections compute fields, their boundary costs, and
their directional response. We now specify the query measurements
used in the illustration. Local outlier factors compare a query's
density with its neighbors. We introduce a few more measures that capture anisotropy. 
The box calibration ALOFT compares the extent needed to include it at a declared reference mass.
Face diffusion
reads attachment to spatial boundaries. We then define uncertainty and support readings from Kriging variance and Nystr\"om residuals. 
The rotating star uses the
same maintained trees to compare these readings and to assign new
observations to a reference clustering.

\subsection{Local Outlier Factors and Calibration}
\label{sec:classicaldefinitions}
\paragraph{Local outlier factors.}
For a reference set $S$, let $N_{k,S}(x)$ contain up to $k$ nearest points of $S$ in Euclidean distance, using all available points if fewer than $k$ are available and a fixed ordering to break ties. Exclude $x$ itself when it belongs to $S$. Let $d_{k,S}(o)$ be the last distance in the neighborhood of $o$. With this capped-neighborhood convention, the local reachability density and local outlier factor of \citet{breunig2000lof}, and their average over the trees, are
\begin{equation}
 \operatorname{lrd}_{k,S}(x)
 =\left[\frac{1}{|N_{k,S}(x)|}\sum_{o\in N_{k,S}(x)}
          \max\{d_{k,S}(o),\|x-o\|_2\}\right]^{-1},
 \label{eq:lofreachdensity}
\end{equation}
\begin{equation}
 \operatorname{LOF}_{k,S}(x)
 =\frac{\sum_{o\in N_{k,S}(x)}\operatorname{lrd}_{k,S}(o)}
        {|N_{k,S}(x)|\operatorname{lrd}_{k,S}(x)},\qquad
 \operatorname{LOFT}_{k}(x)=\frac1{|\mathcal T_*|}
                   \sum_{t\in\mathcal T_*}\operatorname{LOF}_{k,S_t}(x).
 \label{eq:lofloft}
\end{equation}
where $S_t$ is the set of distinct stored points of tree $t$ and
$\mathcal T_*$ contains the trees with at least two such points.
Pooled LOF uses the distinct reference points across the forest;
LOFT computes the ratio separately in each tree's sample and averages
those ratios \citep{zimek2013subsampling}. The tie and cap conventions
above are the ones used in the comparison.

\paragraph{Calibration at a declared mass.}
\label{sec:ratios}
The reference neighborhood must be chosen before its expansion by
the query is measured. Use the calibration box at a declared mass
$\mu$ in Definition~\ref{def:queryboxes}. The default is
$\min\{\psi,2\sqrt\psi\}$; \eqref{eq:calibrationmass} parameterizes other
mass choices. Tree ancestry can overshoot that requested mass.
This controls the population being compared, just as $k$ controls
the neighborhood in LOF, SOD, and COF.

\paragraph{Ratios of extents.} We introduce a new measure ALOFT, which asks how much a neighborhood of the declared mass must
stretch to admit the query. It reads the stretch along each coordinate
after averaging the widths across trees.
Let $E_{tj}$ be the coordinate extents of tree $t$'s calibration box, and let $A_{tj}\geq E_{tj}$ be its extents enlarged to include $x$. Average both over the trees first, to $\bar E_j$ and $\bar A_j$; then
\begin{equation}
 \operatorname{ALOFT}_\mu(x)=\frac1d\sum_j\log\frac{\bar A_j}{\bar E_j}.
 \label{eq:aloft}
\end{equation}
summarizes the per-coordinate ratios, each coordinate in its own units, with no centre, covariance or inverse. A ratio of averages weights each tree by its extent. It is zero when $x$ lies in the box in every tree, and it is the log of the geometric mean of the ratios, which reads a stretch on its own scale rather than as a small excess over one. A coordinate with $\bar E_j=0$ contributes nothing when $\bar A_j=0$; otherwise every tree's box is flat there and $x$ is off it, and $\operatorname{ALOFT}_\mu$ is infinite. Per tree, the log volume ratio is the Kullback--Leibler divergence between the uniform laws on the box without and with the query; we note that other divergences can also be used \citep{amari2016information}.
At the passage box itself the ratio loses most of its contrast, since the passage vertex is chosen by the query's own cut probability, the expansion it measures (Section~\ref{sec:fan}).

\paragraph{Directional expansion of a reference box.}
Let $P_t$ be the uniform law on a tree's calibration box and $Q_t$
the uniform law on that box enlarged to include $x$. For positive
widths, their density ratio on the smaller box is the volume ratio.
Consequently
\begin{equation}
 \operatorname{KL}(P_t\,\|\,Q_t)
 =\sum_j\log\frac{A_{tj}}{E_{tj}}
 =\sum_j\log\left(1+\frac{g^-_{tj}+g^+_{tj}}{E_{tj}}\right).
 \label{eq:boxkl}
\end{equation}
The coordinate term belongs to the one face beyond which $x$ lies.
It therefore gives a signed direction as well as a scalar expansion.
A coordinate on which both boxes are flat contributes zero; leaving
a flat coordinate gives an infinite term unless a positive
regularization scale has been declared.

\paragraph{The order of averaging.}
Equation~\eqref{eq:aloft} averages widths before taking ratios.
The per-coordinate mean $(dN)^{-1}\sum_t\operatorname{KL}(P_t\|Q_t)$
instead measures expansion in each tree before averaging.
These are the two orders already distinguished in
Section~\ref{sec:ensemblegreen}. They need not have the same ranking
and neither uniformly bounds the other. For example, the width pairs
$(E,A)=(1,2),(3,3)$ give $\log(5/4)$ for the ratio of means and
$\tfrac12\log2$ for the mean log ratio; $(1,1),(1,3)$ give
$\log2$ and $\tfrac12\log3$. Pooling widths reduces the influence
of a nearly flat tree, while the mean per-tree divergence retains
that tree's large relative change. The experiment reports both.

\paragraph{Face Diffusion and Boundary Payoffs.} face diffusion asks whether the query can wander to the outer
face of the tree's box before reaching the local data face. This measures escape separately on each face.
The half dimensions of Definition~\ref{def:halfdimensions}
keep the two directions of each coordinate separate. A diffusion
in each direction gives both the probability of reaching the
outer boundary before the data box and the mean time to reach
either boundary. We combine these probabilities in two ways:
using the cut law to choose a face, or giving more weight to
faces with shorter mean passage times.
For face $f=(j,\pm)$ at level $v$, orient the coordinate outward.
Place the query at zero, the data face at $-g_{v,f}<0$, and the
outer face of the root box with $x$ at $\rho_f>0$. A drift-free
diffusion with variance $\sigma_f^2>0$ per unit time has outward
exit probability and mean exit time
\begin{equation}
 p^{\rm out}_{v,f}=\frac{g_{v,f}}{g_{v,f}+\rho_f},\qquad
 m_{v,f}=\E_0\tau_{v,f}=\frac{g_{v,f}\rho_f}{\sigma_f^2}.
 \label{eq:facediffusion}
\end{equation}
These follow from the generator
$\mathcal A_f=(\sigma_f^2/2)\,d^2/dz^2$:
the exit probability solves $\mathcal A_fp^{\rm out}=0$ with endpoint
values $0,1$, and the mean duration solves $\mathcal A_fm=-1$
with zero endpoint values \citep{pitman2018brownian}.
A symmetric walk with steps $\pm\epsilon\sigma_f$ and step
duration $\epsilon^2$ gives the same small-step limit.
The probability depends on the two distances; the duration
also depends on the directional scale $\sigma_f$.

First choose the face using the cut law. Conditional on
separation at $v$, face $f$ has probability $g_{v,f}/G_v$.
Running its diffusion and recording outward exit gives,
after averaging over separation events,
\begin{equation}
 D(x)=\sum_v\frac{s_v}{R(B_v\cup\{x\})}
              \sum_f\frac{g_{v,f}^2}{g_{v,f}+\rho_f}.
 \label{eq:cutdiffusion}
\end{equation}
The inner sum uses positive-gap faces. A level with zero separation probability contributes zero. The directional scales affect duration while the combined exit probability stays fixed.

The reported directional statistic weights each face by the reciprocal of its mean duration, $\nu_{v,f}=1/m_{v,f}$. Its outward probability is
\begin{equation}
 D_\sigma(x)=\sum_vs_v\frac{G_v}{R(B_v\cup\{x\})}
       \frac{\sum_f\nu_{v,f}p^{\rm out}_{v,f}}{\sum_f\nu_{v,f}},\qquad
 \nu_{v,f}=\frac{\sigma_f^2}{g_{v,f}\rho_f}.
 \label{eq:gaugeddiffusion}
\end{equation}
Each direction now influences the answer according to how
quickly its two distances can be traversed at its own scale.
Increasing one $\sigma_f$ increases that face's weight;
multiplying all the $\sigma_f$ by the same factor changes
duration but leaves the weights unchanged.
Section~\ref{sec:fan} uses the forest means of the cut-box
face gaps as these scales.
These are specified inverse-duration weights, rather than the joint exit law of simultaneous face diffusions.
Eligible faces have positive gap and diffusivity. If an eligible face has $\rho_f=0$, use its limiting outward probability one at that level. Otherwise use the finite rates in \eqref{eq:gaugeddiffusion}. With no eligible face, the level contributes zero.

\paragraph{Weighted boundary payoffs.}
Keep the face diffusions and the cut-law face selection in
\eqref{eq:cutdiffusion} fixed. Give outward exit on face $f$ the
declared payoff $\alpha_f\in\R$, with zero payoff at the data face.
Collect the directional exit contributions as
\begin{equation}
 d_f(x)=\sum_{v:g_{v,f}>0}\frac{s_v}{R(B_v\cup\{x\})}
                    \frac{g_{v,f}^2}{g_{v,f}+\rho_f},\qquad
 D_\alpha(x)=\sum_f\alpha_fd_f(x).
 \label{eq:weightedfacepayoff}
\end{equation}
Thus a vector of face contributions supports every fixed linear
payoff; $\alpha_f=1$ recovers \eqref{eq:cutdiffusion}.
The same payoff weighting applies to any fixed face-choice law.
On each interval the harmonic field has boundary values zero and
$\alpha_f$. Its discretizations are the local quadratic potential
game of Section~\ref{sec:game}; their interval limit reads
$\alpha_fp^{\rm out}_{v,f}$ at the query. The interior field
values are the players' actions, while $\alpha$ specifies the
boundary rewards. Changing $\alpha$ changes those rewards while
keeping the process fixed.

The face statistic reads one query against fixed boundaries.
The stopping worth in Proposition~\ref{prop:stoppinggame} measures
the cost saved by adding absorbing members.

\paragraph{The Nystr\"om residual.} This asks how much of the query's kernel
feature the few modes of the reference spectral start leave
unexplained. It evaluates the feature residual of
\eqref{eq:nystromsupport} with the passive row $k_q$ and the modes retained
for the chosen reference model.

\paragraph{Kriging variance.} This asks how much uncertainty about the query
remains when the whole reference set is available. With unit prior
query variance and nugget $\epsilon>0$, it is
$1-k_q^\top(\bar K+\epsilon I)^{-1}k_q$.
Theorem~\ref{prop:krigingsupport} decomposes this variance using
the placement's first-contact scale. The illustration uses the
capped placement below; identification with the original range
escape requires uncapped placement and the stored sample as reference.

\subsection{Rotating Star and Noise Coordinates}
\label{sec:fan}
The rotating-star example compares two uses of the same forest:
clustering a frame after insertion, and reading its queries before
insertion. The former has generating blade identities against
which assignments can be checked. The anomaly comparison measures
agreement with the forest's own flags.

\paragraph{Clustering protocol.}
The library example streams $2700$ points per frame from five
elongated blades, rotated one degree per frame, into $25$ trees of
$256$ points with time decay $1/800$ \citep{rcfcode}.
After insertion, each frame draws $300$ distinct stored points
without replacement in proportion to their copies and routes them
through $N$ trees drawn without replacement. The shared bandwidth
is five percent of their median root range. Cluster count uses the
largest ratio of successive positive relaxed cuts, up to twelve
classes. Thus changing $N$ can change both the ensemble and its
bandwidth. Table~\ref{tab:fan} reports a full turn of $360$ frames.

The normalized-cut procedure uses the spectral start and Lloyd
refinement of Section~\ref{sec:spectralrecipe}. Its representatives
come from recursive two-way normalized cuts of each piece
\citep{shi2000normalized}, bisecting the cheapest piece while its
two-way cut is at most $0.5$, and selecting each piece's point of
largest affinity. CURE uses the library's multi-centroid summarizer
with at most twelve clusters, five representatives per cluster,
and shrinkage $0.05$ \citep{guha1998cure}. The reference partitions
are scored by their exact cut under the same $\bar K$.
The frame's points use the fixed-class mean rule after refinement,
or the Nystr\"om readout of the spectral start; CURE assigns by the
nearest representative. All these assignments use routed rows.

\begin{table}[!htb]
\centering
\small
\begin{tabular}{lcccc}
\toprule
 & forest cut, $N=25$ & $N=10$ & $N=5$ & CURE \\
\midrule
frames with five clusters & $100\%$ & $99.4\%$ & $97.2\%$ & $98.1\%$\\
agreement with the blades & $0.998$ & $0.995$ & $0.981$ & $0.996$\\
\quad by the Nystr\"om extension & $0.999$ & $0.996$ & $0.982$ & \\
$\operatorname{Ncut}$, cut~/~CURE & $0.042/0.044$ & $0.042/0.044$ & $0.039/0.043$ & \\
time per frame (ms) & $18.5+7.7$ & $10.7+3.3$ & $7.8+1.8$ & $20.5$\\
\bottomrule
\end{tabular}
\caption{Clustering on $360$ frames with $300$ reference points.
Agreement matches the $2700$ observations to their generating blades.
Both partitions are scored under the same affinity on frames where
both found five classes. Times are clustering plus assignment for
the forest cut, and their combined time for CURE, on one core.
The Nystr\"om row extends the spectral start.}
\label{tab:fan}
\end{table}

With $N=25$ the forest cut found five classes on every frame.
Where both methods found five, its cut was lower on $84$ frames,
equal on $254$, and higher on $15$. Nystr\"om and fixed-class means
assigned $99.8$--$99.9\%$ of points alike; Nystr\"om took about half
the assignment time in these runs. Refinement moved references on
only five frames. The reported Lloyd pass tests old class means;
it does not certify the exact relocation test
\eqref{eq:cutmove}. The collusive-cut theorems describe additional
tests, not changes made in this experiment.

\paragraph{Coordinate noise.}
Pad every point with eight coordinates drawn independently on
$[-a,a]$ each frame, against blade standard deviations $2$ along
and $0.2$ across. This has the form of
Corollary~\ref{cor:noisygeometry}, with a two-dimensional affine
part and $W=8a$. That corollary controls localization and stretch
above $16W$; the measured behavior at smaller scales is empirical.
Table~\ref{tab:noise} gives the clustering results and the
coordinate selection of \eqref{eq:spectralimportance}.
Up to $a=1$ the rule selected exactly the star's two coordinates
on $344$--$356$ of $360$ frames. When it missed, it retained extra
noise coordinates and kept both star coordinates. Recomputing the
cut on the selected coordinates restored five classes on every
frame and raised agreement at $a=1$ from $0.883$ to $0.996$.
At $a=11$, the trees spent most cuts on noise, neither method
recovered the blades, and the importance ratios followed the range
shares.

\begin{table}[!htb]
\centering
\small
\begin{tabular}{lcccc}
\toprule
noise amplitude $a$ & $0$ & $0.25$ & $0.5$ & $1$\\
\midrule
cut: frames with five clusters & $100\%$ & $99.7\%$ & $97.5\%$ & $85.8\%$\\
\quad agreement with the blades & $0.998$ & $0.996$ & $0.983$ & $0.883$\\
\quad by the Nystr\"om extension & $0.999$ & $0.997$ & $0.984$ & $0.884$\\
CURE: frames with five clusters & $98.1\%$ & $99.4\%$ & $96.4\%$ & $44.4\%$\\
\quad agreement with the blades & $0.996$ & $0.997$ & $0.991$ & $0.856$\\
importance per share: star & $1.05,\,0.96$ & $1.87,\,1.65$ & $2.11,\,2.16$ & $2.44,\,2.38$\\
\quad each noise coordinate & & $0.27$ & $0.37$ & $0.53$\\
frames selecting exactly the star's two & $360$ & $356$ & $347$ & $344$\\
cut on the selected coordinates: five clusters & $100\%$ & $100\%$ & $100\%$ & $100\%$\\
\quad agreement with the blades & $0.998$ & $0.998$ & $0.997$ & $0.996$\\
\bottomrule
\end{tabular}
\caption{Eight independent noise coordinates uniform on $[-a,a]$,
$N=25$ trees, and $360$ frames. Importance per share
\eqref{eq:spectralimportance} is averaged over frames and, in the
noise row, over the eight coordinates. The last two rows recompute
the cut and bandwidth on the same trees using the selected coordinates.}
\label{tab:noise}
\end{table}

\paragraph{SOD: deviation in locally tight coordinates.}
SOD \citep{kriegel2009outlier} chooses a reference set $R(x)$ by
shared-neighbor similarity. Let $\mu_R$ and $v_j$ be its mean and
coordinate variances, and set
\[
 I_R=\left\{j:v_j<\frac{\alpha}{d}\sum_l v_l\right\},\qquad
 \operatorname{SOD}(x)=\frac{\sqrt{\sum_{j\in I_R}(x_j-\mu_{R,j})^2}}
                            {|I_R|}.
\]
Thus it measures deviation from the reference mean along coordinates
on which the references are comparatively tight. The retained coordinate
set records which directions are relevant. Use zero when no coordinate
is selected.

\paragraph{COF: the cost of connecting to neighbors.}
COF \citep{tang2002enhancing} grows a connected set from $x$ through
its $k$ neighbors, repeatedly adding the shortest link from the reached
set to an unreached point. If those links have lengths
$\ell_1,\ldots,\ell_k$, its average chaining distance is
\[
 a_k(x)=\sum_{i=1}^k\frac{2(k-i+1)}{k(k+1)}\ell_i,\qquad
 \operatorname{COF}_k(x)=
 \frac{a_k(x)}{k^{-1}\sum_{o\in N_k(x)}a_k(o)}.
\]
The reference points' chaining distances are computed in their own
neighborhoods. The statistic compares connection cost, rather than
the single-radius reachability density used by LOF.

\paragraph{Measurements before each frame.}
Each query is evaluated on the forest before its frame is inserted.
The comparison reports displacement as a count of old records.
For CoDisp, each possible virtual attachment supplies the largest
sibling ratio on its augmented query path, including the new leaf.
The reported score averages those maxima over attachments and trees.
Range escape and gauged diffusion are the forest means of
\eqref{eq:rangeescape} and \eqref{eq:gaugeddiffusion}. A tree with
zero augmented root range contributes zero to the range mean. The latter
uses one set of directional scales per query, with each $\sigma_f$
the mean cut-box gap on that face over trees that supply geometry
(Section~\ref{sec:density}). Pooled LOF and LOFT are
\eqref{eq:lofloft}, $\operatorname{ALOFT}_\mu$ is \eqref{eq:aloft},
and the mean per-tree comparison is $(dN)^{-1}\sum_t\operatorname{KL}(P_t\|Q_t)$ from \eqref{eq:boxkl}.

The $\mathrm{TF}_0$ and $\mathrm{TF}_2$ readings simulate the power
allocations \eqref{eq:powerallocation} on the maintained length-weighted
trees. They reweight surviving paths by \eqref{eq:cutlikelihood} and
use the target separation probabilities \eqref{eq:hyperhazard}, with
the same inverse-depth payoff \eqref{eq:inverse} for an unseen query.
\emph{They ask how readily the query would escape if cuts gave equal
weight to nonflat coordinates ($\mathrm{TF}_0$), or weight proportional
to squared width ($\mathrm{TF}_2$).}
The first uses the coordinate allocation of isolation forests
\citep{liu2008isolation}. Section~\ref{sec:cutgeometry} discusses
simulation under these cut laws.

\paragraph{Pre-frame reference diagnostics.}
The Nystr\"om residual and the Kriging variance use the preceding
frame's reference set and its fixed, routed affinity matrix $\bar K$.
Both use the same passive query row $k_q$, averaged over trees and
randomized placements as in Section~\ref{sec:support}. In the example,
the placement height is capped at the parent of the separation
vertex $v$,
\begin{equation}
 2H_{\rm run}=\min\{R(B_v\cup\{x\}),R(B_{\operatorname{par}(v)})\}\quad(v\ne{\rm root}),
 \label{eq:fanplacement}
\end{equation}
and at the root the full augmented range is used. Each placement
registers the query's gap from its separation box and uses routed
affinities for references that parted higher up.
The Kriging calculation uses unit probe variance and
nugget $\epsilon=10^{-6}$; both residuals are read before insertion.

\paragraph{Comparing the query readings.}
On frames with a preceding reference snapshot, every ninth point
was read before insertion, giving $2346$ flagged and $105354$ other
points. Table~\ref{tab:readings} collects agreement with those
flags on the unpadded star and at $a=0.5$. These are agreement
measurements, since the forest's score supplies the flags.

\begin{table}[!htb]
\centering
\small
\begin{tabular}{lcc}
\toprule
Reading & Unpadded star & Eight noise coordinates, $a=0.5$\\
\midrule
CoDisp & $0.997$ & $0.998$\\
Displacement & $0.981$ & $0.993$\\
Range escape & $0.982$ & $0.976$\\
Gauged diffusion & $0.995$ & $0.828$\\
Pooled LOF, $k=32$ & $0.966$ & $0.970$\\
LOFT, $k=32$ & $0.989$ & $0.989$\\
ALOFT, default calibration mass & $0.973$ & $0.970$\\
COF & $0.929$ & $0.873$\\
SOD & $0.790$ & $0.602$\\
$\mathrm{TF}_0$ & $0.963$ & $0.746$\\
$\mathrm{TF}_2$ & $0.998$ &  0.908\\
Nystr\"om residual & $0.951$ & $0.899$\\
Kriging variance & $0.956$ & $0.867$\\
\bottomrule
\end{tabular}
\caption{Area under the curve against the forest's pre-insertion
flags. Ordinary arithmetic-mean CoDisp is used. COF and SOD use
$k=32$ on the pooled references; SOD uses $16$ shared neighbors and
variance threshold $0.8$. A dash denotes a value not reported here.
The ALOFT row uses the reported default-mass run.}
\label{tab:readings}
\end{table}

The neighborhood being compared matters. Pooled LOF and LOFT at
the same $k=32$ had rank correlation $0.38$: $32$ of a tree's $256$
records is a coarser neighborhood than $32$ of roughly $2200$
pooled points. LOFT and ALOFT at mass $32$ had correlation $0.59$.
Using the passage box for ALOFT instead gave area $0.419$, against
$0.973$ at the calibration box: the passage vertex is selected by
the query's own expansion, whereas calibration fixes the reference
mass first.

The geometry also locates the contribution. In the first $120$
frames, $99.6\%$ of flagged points lay at blade tips beyond $1.5$
standard deviations along the blade. Their signed expansion vectors
had mean $|\cos|$ $0.942$ with its long axis, against $0.733$ inside
a blade. At $a=0.5$, these vectors kept $91\%$ of their mass for
flagged tip points on the two star coordinates, against $26\%$
inside a blade. The retained faces distinguish extension along a
blade from deviation across its narrow side.

The two support residuals lose contrast under noise. At $a=0.5$
the Kriging variance averaged $0.991$ for flagged and $0.979$ for
other points: both were weakly supported under the chosen geometry.
The noise coordinates widen the box at which the probe first meets
the references. This is the missing-support term of
Theorem~\ref{prop:krigingsupport}; its initial-rate identification
with range escape uses uncapped placement and the stored sample
as reference, rather than the capped placement used here.

\subsubsection*{AI Use Statement}
We used multiple generative AI tools for this work. We have reviewed all AI-assisted work.

\bibliography{anisotropic_full}

\begin{thebibliography}{76}
\providecommand{\natexlab}[1]{#1}
\providecommand{\url}[1]{\texttt{#1}}
\expandafter\ifx\csname urlstyle\endcsname\relax
  \providecommand{\doi}[1]{doi: #1}\else
  \providecommand{\doi}{doi: \begingroup \urlstyle{rm}\Url}\fi

\bibitem[Alzate and Suykens(2010)]{alzate2010multiway}
Carlos Alzate and Johan A.~K. Suykens.
\newblock Multiway spectral clustering with out-of-sample extensions through
  weighted kernel {PCA}.
\newblock \emph{IEEE Transactions on Pattern Analysis and Machine
  Intelligence}, 32\penalty0 (2):\penalty0 335--347, 2010.

\bibitem[Amari(2016)]{amari2016information}
Shun-ichi Amari.
\newblock \emph{Information Geometry and Its Applications}, volume 194 of
  \emph{Applied Mathematical Sciences}.
\newblock Springer, Tokyo, 2016.
\newblock \doi{10.1007/978-4-431-55978-8}.

\bibitem[Amari and Nagaoka(2000)]{amari2000methods}
Shun-ichi Amari and Hiroshi Nagaoka.
\newblock \emph{Methods of Information Geometry}, volume 191 of
  \emph{Translations of Mathematical Monographs}.
\newblock American Mathematical Society and Oxford University Press, 2000.
\newblock \doi{10.1090/mmono/191}.

\bibitem[{Amazon Web Services}(2026)]{ampanomaly}
{Amazon Web Services}.
\newblock Amazon managed service for prometheus user guide: Anomaly detection.
\newblock
  \url{https://docs.aws.amazon.com/prometheus/latest/userguide/prometheus-anomaly-detection.html},
  2026.

\bibitem[Arbel{\'a}ez et~al.(2011)Arbel{\'a}ez, Maire, Fowlkes, and
  Malik]{arbelaez2011contour}
Pablo Arbel{\'a}ez, Michael Maire, Charless Fowlkes, and Jitendra Malik.
\newblock Contour detection and hierarchical image segmentation.
\newblock \emph{IEEE Transactions on Pattern Analysis and Machine
  Intelligence}, 33\penalty0 (5):\penalty0 898--916, 2011.
\newblock \doi{10.1109/TPAMI.2010.161}.

\bibitem[Avrachenkov et~al.(2018)Avrachenkov, Kondratev, Mazalov, and
  Rubanov]{avrachenkov2018network}
Konstantin Avrachenkov, Aleksei~Yu. Kondratev, Vladimir~V. Mazalov, and Dmytro
  Rubanov.
\newblock Network partitioning algorithms as cooperative games.
\newblock \emph{Computational Social Networks}, 5:\penalty0 11, 2018.
\newblock \doi{10.1186/s40649-018-0059-5}.

\bibitem[Belkin and Niyogi(2003)]{belkin2003laplacian}
Mikhail Belkin and Partha Niyogi.
\newblock Laplacian eigenmaps for dimensionality reduction and data
  representation.
\newblock \emph{Neural Computation}, 15\penalty0 (6):\penalty0 1373--1396,
  2003.
\newblock \doi{10.1162/089976603321780317}.
\newblock URL \url{https://misha.belkin-wang.org/papers/LEM_NC_03.pdf}.

\bibitem[Bendikov et~al.(2014)Bendikov, Grigor'yan, Pittet, and
  Woess]{bendikov2014isotropic}
A.~D. Bendikov, A.~A. Grigor'yan, Ch. Pittet, and W.~Woess.
\newblock Isotropic {M}arkov semigroups on ultra-metric spaces.
\newblock \emph{Russian Mathematical Surveys}, 69\penalty0 (4):\penalty0
  589--680, 2014.

\bibitem[Bengio et~al.(2004)Bengio, Paiement, Vincent, Delalleau, Le~Roux, and
  Ouimet]{bengio2004outofsample}
Yoshua Bengio, Jean-Fran{\c{c}}ois Paiement, Pascal Vincent, Olivier Delalleau,
  Nicolas Le~Roux, and Marie Ouimet.
\newblock Out-of-sample extensions for {LLE}, {Isomap}, {MDS}, {Eigenmaps}, and
  spectral clustering.
\newblock In \emph{Advances in Neural Information Processing Systems 16}, 2004.
\newblock URL
  \url{https://proceedings.neurips.cc/paper/2003/hash/cf05968255451bdefe3c5bc64d550517-Abstract.html}.

\bibitem[Bhattacharjee et~al.(2025)Bhattacharjee, Frohnapfel, and von
  Luxburg]{bhattacharjee2025discard}
Robi Bhattacharjee, Karolin Frohnapfel, and Ulrike von Luxburg.
\newblock How to safely discard features based on aggregate {SHAP} values.
\newblock In \emph{Proceedings of the 38th Conference on Learning Theory},
  volume 291 of \emph{Proceedings of Machine Learning Research}, pages
  280--314, 2025.
\newblock URL \url{https://proceedings.mlr.press/v291/bhattacharjee25a.html}.

\bibitem[Blume(1993)]{blume1993statistical}
Lawrence~E. Blume.
\newblock The statistical mechanics of strategic interaction.
\newblock \emph{Games and Economic Behavior}, 5\penalty0 (3):\penalty0
  387--424, 1993.
\newblock \doi{10.1006/game.1993.1023}.

\bibitem[Breiman(1996)]{breiman1996bagging}
Leo Breiman.
\newblock Bagging predictors.
\newblock \emph{Machine Learning}, 24\penalty0 (2):\penalty0 123--140, 1996.
\newblock \doi{10.1007/BF00058655}.

\bibitem[Breiman(2001)]{breiman2001random}
Leo Breiman.
\newblock Random forests.
\newblock \emph{Machine Learning}, 45\penalty0 (1):\penalty0 5--32, 2001.
\newblock \doi{10.1023/A:1010933404324}.

\bibitem[Breiman et~al.(1984{\natexlab{a}})Breiman, Friedman, Olshen, and
  Stone]{Breiman1984}
Leo Breiman, Jerome~H. Friedman, Richard~A. Olshen, and Charles~J. Stone.
\newblock \emph{Classification and Regression Trees}.
\newblock Wadsworth, 1984{\natexlab{a}}.
\newblock ISBN 0-534-98053-8.

\bibitem[Breiman et~al.(1984{\natexlab{b}})Breiman, Friedman, Olshen, and
  Stone]{breiman1984classification}
Leo Breiman, Jerome~H. Friedman, Richard~A. Olshen, and Charles~J. Stone.
\newblock \emph{Classification and Regression Trees}.
\newblock Wadsworth, Belmont, CA, 1984{\natexlab{b}}.

\bibitem[Breunig et~al.(2000)Breunig, Kriegel, Ng, and Sander]{breunig2000lof}
Markus~M. Breunig, Hans-Peter Kriegel, Raymond~T. Ng, and J{\"o}rg Sander.
\newblock {LOF}: Identifying density-based local outliers.
\newblock In \emph{Proceedings of the ACM SIGMOD International Conference on
  Management of Data}, pages 93--104, 2000.
\newblock \doi{10.1145/342009.335388}.

\bibitem[Chou et~al.(1994)Chou, Willsky, and Benveniste]{chou1994multiscale}
Kenneth~C. Chou, Alan~S. Willsky, and Albert Benveniste.
\newblock Multiscale recursive estimation, data fusion, and regularization.
\newblock \emph{IEEE Transactions on Automatic Control}, 39\penalty0
  (3):\penalty0 464--478, 1994.

\bibitem[Cohen and Peng(2015)]{cohen2015lewis}
Michael~B. Cohen and Richard Peng.
\newblock {$\ell_p$} row sampling by lewis weights.
\newblock In \emph{Proceedings of the 47th Annual ACM Symposium on Theory of
  Computing}, pages 183--192, 2015.
\newblock \doi{10.1145/2746539.2746567}.
\newblock URL \url{https://arxiv.org/abs/1412.0588}.

\bibitem[Coifman and Lafon(2006)]{coifman2006diffusion}
Ronald~R. Coifman and St{\'e}phane Lafon.
\newblock Diffusion maps.
\newblock \emph{Applied and Computational Harmonic Analysis}, 21\penalty0
  (1):\penalty0 5--30, 2006.

\bibitem[Dasgupta and Freund(2008)]{dasgupta2008random}
Sanjoy Dasgupta and Yoav Freund.
\newblock Random projection trees and low dimensional manifolds.
\newblock In \emph{Proceedings of the 40th Annual ACM Symposium on Theory of
  Computing}, pages 537--546, 2008.
\newblock \doi{10.1145/1374376.1374452}.
\newblock URL \url{https://cseweb.ucsd.edu/~yfreund/papers/rptree-tr.pdf}.

\bibitem[Delalleau et~al.(2005)Delalleau, Bengio, and
  Le~Roux]{delalleau2005efficient}
Olivier Delalleau, Yoshua Bengio, and Nicolas Le~Roux.
\newblock Efficient non-parametric function induction in semi-supervised
  learning.
\newblock In \emph{Proceedings of the Tenth International Workshop on
  Artificial Intelligence and Statistics}, volume~R5 of \emph{Proceedings of
  Machine Learning Research}, pages 96--103, 2005.
\newblock URL \url{https://proceedings.mlr.press/r5/delalleau05a.html}.

\bibitem[Deng and Papadimitriou(1994)]{deng1994complexity}
Xiaotie Deng and Christos~H. Papadimitriou.
\newblock On the complexity of cooperative solution concepts.
\newblock \emph{Mathematics of Operations Research}, 19\penalty0 (2):\penalty0
  257--266, 1994.
\newblock \doi{10.1287/moor.19.2.257}.

\bibitem[Dhillon et~al.(2004)Dhillon, Guan, and Kulis]{dhillon2004kernel}
Inderjit~S. Dhillon, Yuqiang Guan, and Brian Kulis.
\newblock Kernel $k$-means, spectral clustering and normalized cuts.
\newblock In \emph{Proceedings of the Tenth ACM SIGKDD International Conference
  on Knowledge Discovery and Data Mining}, pages 551--556, 2004.
\newblock URL \url{https://people.bu.edu/bkulis/pubs/spectral_kdd.pdf}.

\bibitem[Dhillon et~al.(2005)Dhillon, Guan, and Kulis]{dhillon2005multilevel}
Inderjit~S. Dhillon, Yuqiang Guan, and Brian Kulis.
\newblock A fast kernel-based multilevel algorithm for graph clustering.
\newblock In \emph{Proceedings of the Eleventh ACM SIGKDD International
  Conference on Knowledge Discovery and Data Mining}, 2005.
\newblock \doi{10.1145/1081870.1081948}.

\bibitem[Domingos and Hulten(2000)]{domingos2000mining}
Pedro Domingos and Geoff Hulten.
\newblock Mining high-speed data streams.
\newblock In \emph{Proceedings of the Sixth ACM SIGKDD International Conference
  on Knowledge Discovery and Data Mining}, pages 71--80, 2000.
\newblock \doi{10.1145/347090.347107}.
\newblock URL
  \url{https://alchemy.cs.washington.edu/papers/pdfs/domingos-hulten00.pdf}.

\bibitem[D{\"o}rfler and Bullo(2013)]{dorfler2013kron}
Florian D{\"o}rfler and Francesco Bullo.
\newblock {Kron} reduction of graphs with applications to electrical networks.
\newblock \emph{IEEE Transactions on Circuits and Systems I: Regular Papers},
  60\penalty0 (1):\penalty0 150--163, 2013.
\newblock \doi{10.1109/TCSI.2012.2215780}.
\newblock URL \url{https://motion.me.ucsb.edu/pdf/2011d-db.pdf}.

\bibitem[Doyle and Snell(1984)]{doyle1984random}
Peter~G. Doyle and J.~Laurie Snell.
\newblock \emph{Random Walks and Electric Networks}, volume~22 of \emph{Carus
  Mathematical Monographs}.
\newblock Mathematical Association of America, 1984.
\newblock URL \url{https://arxiv.org/abs/math/0001057}.

\bibitem[Fowlkes et~al.(2004)Fowlkes, Belongie, Chung, and
  Malik]{fowlkes2004spectral}
Charless Fowlkes, Serge Belongie, Fan Chung, and Jitendra Malik.
\newblock Spectral grouping using the {N}ystr{\"o}m method.
\newblock \emph{IEEE Transactions on Pattern Analysis and Machine
  Intelligence}, 26\penalty0 (2):\penalty0 214--225, 2004.

\bibitem[Freund and Schapire(1996)]{freund1996experiments}
Yoav Freund and Robert~E. Schapire.
\newblock Experiments with a new boosting algorithm.
\newblock In \emph{Machine Learning: Proceedings of the Thirteenth
  International Conference (ICML '96)}, pages 148--156, Bari, Italy, 1996.
  Morgan Kaufmann.

\bibitem[Friedman(2001)]{friedman2001greedy}
Jerome~H. Friedman.
\newblock Greedy function approximation: A gradient boosting machine.
\newblock \emph{The Annals of Statistics}, 29\penalty0 (5):\penalty0
  1189--1232, 2001.

\bibitem[Geurts et~al.(2006)Geurts, Ernst, and Wehenkel]{geurts2006extremely}
Pierre Geurts, Damien Ernst, and Louis Wehenkel.
\newblock Extremely randomized trees.
\newblock \emph{Machine Learning}, 63\penalty0 (1):\penalty0 3--42, 2006.
\newblock \doi{10.1007/s10994-006-6226-1}.

\bibitem[Gomes et~al.(2018)Gomes, Barddal, Boiko, and
  Bifet]{gomes2018regression}
Heitor~Murilo Gomes, Jean~Paul Barddal, Luis~Eduardo Boiko, and Albert Bifet.
\newblock Adaptive random forests for data stream regression.
\newblock In \emph{Proceedings of the European Symposium on Artificial Neural
  Networks, Computational Intelligence and Machine Learning}, pages 267--272,
  2018.
\newblock URL
  \url{https://www.esann.org/sites/default/files/proceedings/legacy/es2018-183.pdf}.

\bibitem[Guha and Tokle(2019)]{guha2019rcfblog}
Sudipto Guha and Joshua Tokle.
\newblock Random cut forests.
\newblock OpenSearch Blog; first posted on the Open Distro for Elasticsearch
  blog, 26 November 2019.
  \url{https://opensearch.org/blog/random-cut-forests/}, 2019.

\bibitem[Guha et~al.(1998)Guha, Rastogi, and Shim]{guha1998cure}
Sudipto Guha, Rajeev Rastogi, and Kyuseok Shim.
\newblock {CURE}: An efficient clustering algorithm for large databases.
\newblock In \emph{Proceedings of the ACM SIGMOD International Conference on
  Management of Data}, pages 73--84, 1998.
\newblock \doi{10.1145/276304.276312}.

\bibitem[Guha et~al.(2003)Guha, Meyerson, Mishra, Motwani, and
  O'Callaghan]{guha2003streams}
Sudipto Guha, Adam Meyerson, Nina Mishra, Rajeev Motwani, and Liadan
  O'Callaghan.
\newblock Clustering data streams: Theory and practice.
\newblock \emph{IEEE Transactions on Knowledge and Data Engineering},
  15\penalty0 (3):\penalty0 515--528, 2003.
\newblock URL
  \url{https://www.cs.emory.edu/~cheung/Courses/584/Syllabus/papers/Clustering/2003-Guha-clustering.pdf}.

\bibitem[Guha et~al.(2016)Guha, Mishra, Roy, and Schrijvers]{guha2016rrcf}
Sudipto Guha, Nina Mishra, Gourav Roy, and Okke Schrijvers.
\newblock Robust random cut forest based anomaly detection on streams.
\newblock In \emph{Proceedings of the 33rd International Conference on Machine
  Learning}, volume~48 of \emph{Proceedings of Machine Learning Research},
  pages 2712--2721, 2016.
\newblock URL \url{https://proceedings.mlr.press/v48/guha16.html}.

\bibitem[Harsanyi(1963)]{harsanyi1963simplified}
John~C. Harsanyi.
\newblock A simplified bargaining model for the $n$-person cooperative game.
\newblock \emph{International Economic Review}, 4\penalty0 (2):\penalty0
  194--220, 1963.
\newblock \doi{10.2307/2525487}.

\bibitem[Hein et~al.(2007)Hein, Audibert, and von Luxburg]{hein2007graph}
Matthias Hein, Jean-Yves Audibert, and Ulrike von Luxburg.
\newblock Graph laplacians and their convergence on random neighborhood graphs.
\newblock \emph{Journal of Machine Learning Research}, 8\penalty0
  (48):\penalty0 1325--1368, 2007.
\newblock URL \url{https://jmlr.org/papers/v8/hein07a.html}.

\bibitem[Hoffmann and Jones(2015)]{hoffmann2015unified}
Till Hoffmann and Nick~S. Jones.
\newblock Unified treatment of the asymptotics of asymmetric kernel density
  estimators.
\newblock arXiv:1512.03188, 2015.
\newblock URL \url{https://arxiv.org/abs/1512.03188}.

\bibitem[Kriegel et~al.(2009)Kriegel, Kr{\"o}ger, Schubert, and
  Zimek]{kriegel2009outlier}
Hans-Peter Kriegel, Peer Kr{\"o}ger, Erich Schubert, and Arthur Zimek.
\newblock Outlier detection in axis-parallel subspaces of high dimensional
  data.
\newblock In \emph{Advances in Knowledge Discovery and Data Mining, 13th
  Pacific-Asia Conference, {PAKDD} 2009}, volume 5476 of \emph{Lecture Notes in
  Computer Science}, pages 831--838. Springer, 2009.
\newblock \doi{10.1007/978-3-642-01307-2_86}.

\bibitem[Langevin(1908)]{langevin1908}
Paul Langevin.
\newblock Sur la th{\'e}orie du mouvement brownien.
\newblock \emph{Comptes Rendus de l'Acad{\'e}mie des Sciences}, 146:\penalty0
  530--533, 1908.

\bibitem[Last and Penrose(2017)]{last2017poisson}
G{\"u}nter Last and Mathew Penrose.
\newblock \emph{Lectures on the {Poisson} Process}.
\newblock Cambridge University Press, 2017.
\newblock \doi{10.1017/9781316104477}.
\newblock URL \url{https://stoch.math.kit.edu/img/Last/lastpenrose2017.pdf}.

\bibitem[LeCun et~al.(2006)LeCun, Chopra, Hadsell, Ranzato, and
  Huang]{lecun2006tutorial}
Yann LeCun, Sumit Chopra, Raia Hadsell, Marc'Aurelio Ranzato, and Fu~Jie Huang.
\newblock A tutorial on energy-based learning.
\newblock In G{\"o}khan Bak{\i}r, Thomas Hofmann, Bernhard Sch{\"o}lkopf,
  Alexander~J. Smola, Ben Taskar, and S.~V.~N. Vishwanathan, editors,
  \emph{Predicting Structured Data}. MIT Press, 2006.

\bibitem[Lemons and Gythiel(1997)]{lemons1997}
Don~S. Lemons and Anthony Gythiel.
\newblock Paul langevin's 1908 paper ``on the theory of brownian motion''.
\newblock \emph{American Journal of Physics}, 65\penalty0 (11):\penalty0
  1079--1081, 1997.

\bibitem[Lewis(1978)]{lewis1978finite}
D.~R. Lewis.
\newblock Finite dimensional subspaces of {$L_p$}.
\newblock \emph{Studia Mathematica}, 63\penalty0 (2):\penalty0 207--212, 1978.
\newblock \doi{10.4064/sm-63-2-207-212}.

\bibitem[Li and Blomberg(2025)]{li2025missing}
Kaituo Li and Patrik Blomberg.
\newblock A customer impact journey: When no data is still important data.
\newblock OpenSearch Blog, December 2025.
\newblock URL
  \url{https://opensearch.org/blog/a-customer-impact-journey-when-no-data-is-still-important-data/}.

\bibitem[Li and Kolchina(2025)]{li2025falsepositives}
Kaituo Li and Fanit Kolchina.
\newblock Enhancing {OpenSearch} anomaly detection: Reducing false positives
  through algorithmic improvements.
\newblock OpenSearch Blog, February 2025.
\newblock URL
  \url{https://opensearch.org/blog/reducing-false-positives-through-algorithmic-improvements/}.

\bibitem[Liu et~al.(2008)Liu, Ting, and Zhou]{liu2008isolation}
Fei~Tony Liu, Kai~Ming Ting, and Zhi-Hua Zhou.
\newblock Isolation forest.
\newblock In \emph{Proceedings of the Eighth IEEE International Conference on
  Data Mining}, pages 413--422, 2008.
\newblock \doi{10.1109/ICDM.2008.17}.

\bibitem[Loftsgaarden and Quesenberry(1965)]{loftsgaarden1965nonparametric}
D.~O. Loftsgaarden and C.~P. Quesenberry.
\newblock A nonparametric estimate of a multivariate density function.
\newblock \emph{The Annals of Mathematical Statistics}, 36\penalty0
  (3):\penalty0 1049--1051, 1965.
\newblock \doi{10.1214/aoms/1177700079}.

\bibitem[Lundberg et~al.(2020)Lundberg, Erion, Chen, DeGrave, Prutkin, Nair,
  Katz, Himmelfarb, Bansal, and Lee]{lundberg2020local}
Scott~M. Lundberg, Gabriel Erion, Hugh Chen, Alex DeGrave, Jordan~M. Prutkin,
  Bala Nair, Ronit Katz, Jonathan Himmelfarb, Nisha Bansal, and Su-In Lee.
\newblock From local explanations to global understanding with explainable {AI}
  for trees.
\newblock \emph{Nature Machine Intelligence}, 2:\penalty0 56--67, 2020.
\newblock \doi{10.1038/s42256-019-0138-9}.
\newblock URL \url{https://pmc.ncbi.nlm.nih.gov/articles/PMC7326367/}.

\bibitem[Lyons and Peres(2016)]{lyons2016probability}
Russell Lyons and Yuval Peres.
\newblock \emph{Probability on Trees and Networks}, volume~42 of
  \emph{Cambridge Series in Statistical and Probabilistic Mathematics}.
\newblock Cambridge University Press, 2016.
\newblock \doi{10.1017/9781316672815}.
\newblock URL \url{https://rdlyons.pages.iu.edu/prbtree/}.

\bibitem[Mavroforakis et~al.(2015)Mavroforakis, Mathioudakis, and
  Gionis]{mavroforakis2015absorbing}
Charalampos Mavroforakis, Michael Mathioudakis, and Aristides Gionis.
\newblock Absorbing random-walk centrality: Theory and algorithms.
\newblock In \emph{2015 IEEE International Conference on Data Mining}, pages
  901--906, 2015.
\newblock \doi{10.1109/ICDM.2015.103}.
\newblock URL \url{https://arxiv.org/abs/1509.02533}.

\bibitem[Meil{\u{a}} and Shi(2001)]{meila2001random}
Marina Meil{\u{a}} and Jianbo Shi.
\newblock A random walks view of spectral segmentation.
\newblock In \emph{Proceedings of the Eighth International Workshop on
  Artificial Intelligence and Statistics}, volume~R3 of \emph{Proceedings of
  Machine Learning Research}, pages 203--208, 2001.
\newblock URL \url{https://proceedings.mlr.press/r3/meila01a.html}.

\bibitem[Monderer and Shapley(1996)]{monderer1996potential}
Dov Monderer and Lloyd~S. Shapley.
\newblock Potential games.
\newblock \emph{Games and Economic Behavior}, 14\penalty0 (1):\penalty0
  124--143, 1996.

\bibitem[Myerson(1980)]{myerson1980conference}
Roger~B. Myerson.
\newblock Conference structures and fair allocation rules.
\newblock \emph{International Journal of Game Theory}, 9\penalty0 (3):\penalty0
  169--182, 1980.
\newblock \doi{10.1007/BF01781371}.

\bibitem[Nash(1951)]{nash1951noncooperative}
John Nash.
\newblock Non-cooperative games.
\newblock \emph{Annals of Mathematics}, 54\penalty0 (2):\penalty0 286--295,
  1951.

\bibitem[Ng et~al.(2001)Ng, Jordan, and Weiss]{ng2001spectral}
Andrew~Y. Ng, Michael~I. Jordan, and Yair Weiss.
\newblock On spectral clustering: Analysis and an algorithm.
\newblock In \emph{Advances in Neural Information Processing Systems},
  volume~14, pages 849--856, 2001.
\newblock URL
  \url{https://papers.nips.cc/paper/2092-on-spectral-clustering-analysis-and-an-algorithm}.

\bibitem[{OpenSearch Project}(2026{\natexlab{a}})]{opensearchad}
{OpenSearch Project}.
\newblock {OpenSearch} documentation: Anomaly detection.
\newblock
  \url{https://docs.opensearch.org/latest/observing-your-data/ad/index/},
  2026{\natexlab{a}}.

\bibitem[{OpenSearch Project}(2026{\natexlab{b}})]{opensearchforecast}
{OpenSearch Project}.
\newblock {OpenSearch} documentation: Forecasting.
\newblock
  \url{https://docs.opensearch.org/latest/observing-your-data/forecast/index/},
  2026{\natexlab{b}}.

\bibitem[Pavliotis(2014)]{pavliotis2014stochastic}
Grigorios~A. Pavliotis.
\newblock \emph{Stochastic Processes and Applications: Diffusion Processes, the
  Fokker--Planck and Langevin Equations}, volume~60 of \emph{Texts in Applied
  Mathematics}.
\newblock Springer, New York, 2014.
\newblock \doi{10.1007/978-1-4939-1323-7}.

\bibitem[Pitman and Tang(2018)]{pitman2018tree}
Jim Pitman and Wenpin Tang.
\newblock Tree formulas, mean first passage times and {Kemeny's} constant of a
  {Markov} chain.
\newblock \emph{Bernoulli}, 24\penalty0 (3):\penalty0 1942--1972, 2018.
\newblock \doi{10.3150/16-BEJ916}.
\newblock URL \url{https://arxiv.org/abs/1603.09017}.

\bibitem[Pitman and Yor(2018)]{pitman2018brownian}
Jim Pitman and Marc Yor.
\newblock A guide to {Brownian} motion and related stochastic processes.
\newblock arXiv:1802.09679, 2018.
\newblock URL \url{https://arxiv.org/abs/1802.09679}.

\bibitem[{Random Cut Forest contributors}(2026)]{opensearchrcf}
{Random Cut Forest contributors}.
\newblock Random cut forest: {Java} library used by {OpenSearch}.
\newblock
  \url{https://github.com/aws/random-cut-forest-by-aws/tree/4.4.0-java}, 2026.

\bibitem[Rasmussen and Williams(2006)]{rasmussen2006gaussian}
Carl~Edward Rasmussen and Christopher K.~I. Williams.
\newblock \emph{Gaussian Processes for Machine Learning}.
\newblock MIT Press, Cambridge, MA, 2006.
\newblock \doi{10.7551/mitpress/3206.001.0001}.
\newblock URL \url{https://gaussianprocess.org/gpml/}.

\bibitem[Shi and Malik(2000)]{shi2000normalized}
Jianbo Shi and Jitendra Malik.
\newblock Normalized cuts and image segmentation.
\newblock \emph{IEEE Transactions on Pattern Analysis and Machine
  Intelligence}, 22\penalty0 (8):\penalty0 888--905, 2000.
\newblock \doi{10.1109/34.868688}.

\bibitem[Stein(1999)]{stein1999interpolation}
Michael~L. Stein.
\newblock \emph{Interpolation of Spatial Data: Some Theory for {K}riging}.
\newblock Springer, New York, 1999.

\bibitem[{Streaming Algorithms}(2026)]{rcfcode}
{Streaming Algorithms}.
\newblock Random cut forest: {Java} implementation.
\newblock \url{https://github.com/streamingalgorithms/randomcutforest}, 2026.

\bibitem[Tang et~al.(2002)Tang, Chen, Fu, and Cheung]{tang2002enhancing}
Jian Tang, Zhixiang Chen, Ada Wai-Chee Fu, and David~W. Cheung.
\newblock Enhancing effectiveness of outlier detections for low density
  patterns.
\newblock In \emph{Advances in Knowledge Discovery and Data Mining, 6th
  Pacific-Asia Conference, {PAKDD} 2002}, volume 2336 of \emph{Lecture Notes in
  Computer Science}, pages 535--548. Springer, 2002.
\newblock \doi{10.1007/3-540-47887-6_53}.

\bibitem[Terrell and Scott(1992)]{terrell1992variable}
George~R. Terrell and David~W. Scott.
\newblock Variable kernel density estimation.
\newblock \emph{The Annals of Statistics}, 20\penalty0 (3):\penalty0
  1236--1265, 1992.
\newblock \doi{10.1214/aos/1176348768}.
\newblock URL \url{https://doi.org/10.1214/aos/1176348768}.

\bibitem[van~den Nouweland et~al.(1992)van~den Nouweland, Borm, and
  Tijs]{vandennouweland1992allocation}
Anne van~den Nouweland, Peter Borm, and Stef Tijs.
\newblock Allocation rules for hypergraph communication situations.
\newblock \emph{International Journal of Game Theory}, 20\penalty0
  (3):\penalty0 255--268, 1992.
\newblock \doi{10.1007/BF01253780}.

\bibitem[Verma et~al.(2009)Verma, Kpotufe, and Dasgupta]{verma2009spatial}
Nakul Verma, Samory Kpotufe, and Sanjoy Dasgupta.
\newblock Which spatial partition trees are adaptive to intrinsic dimension?
\newblock In \emph{Proceedings of the 25th Conference on Uncertainty in
  Artificial Intelligence}, pages 565--574, 2009.

\bibitem[Vitter(1985)]{vitter1985reservoir}
Jeffrey~S. Vitter.
\newblock Random sampling with a reservoir.
\newblock \emph{ACM Transactions on Mathematical Software}, 11\penalty0
  (1):\penalty0 37--57, 1985.
\newblock \doi{10.1145/3147.3165}.

\bibitem[Wagner et~al.(2018)Wagner, Guha, Kasiviswanathan, and
  Mishra]{wagner2018temporal}
Tal Wagner, Sudipto Guha, Shiva Kasiviswanathan, and Nina Mishra.
\newblock Semi-supervised learning on data streams via temporal label
  propagation.
\newblock In \emph{Proceedings of the 35th International Conference on Machine
  Learning}, volume~80 of \emph{Proceedings of Machine Learning Research},
  pages 5095--5104. PMLR, 2018.
\newblock URL \url{https://proceedings.mlr.press/v80/wagner18a.html}.

\bibitem[Ward(1963)]{ward1963hierarchical}
Joe~H. Ward, Jr.
\newblock Hierarchical grouping to optimize an objective function.
\newblock \emph{Journal of the American Statistical Association}, 58\penalty0
  (301):\penalty0 236--244, 1963.

\bibitem[Zhu et~al.(2003)Zhu, Ghahramani, and Lafferty]{zhu2003semisupervised}
Xiaojin Zhu, Zoubin Ghahramani, and John Lafferty.
\newblock Semi-supervised learning using {Gaussian} fields and harmonic
  functions.
\newblock In \emph{Proceedings of the Twentieth International Conference on
  Machine Learning}, pages 912--919, 2003.
\newblock URL \url{https://pages.cs.wisc.edu/~jerryzhu/pub/zgl.pdf}.

\bibitem[Zimek et~al.(2013)Zimek, Gaudet, Campello, and
  Sander]{zimek2013subsampling}
Arthur Zimek, Matthew Gaudet, Ricardo J. G.~B. Campello, and J{\"o}rg Sander.
\newblock Subsampling for efficient and effective unsupervised outlier
  detection ensembles.
\newblock In \emph{Proceedings of the 19th ACM SIGKDD International Conference
  on Knowledge Discovery and Data Mining}, pages 428--436, 2013.

\end{thebibliography}

\end{document}